\documentclass[letterpaper]{article}
\usepackage[margin=1in,dvips]{geometry}
\usepackage{graphicx,psfrag,amsmath,amsthm,amssymb}
\usepackage[numbers,sort&compress]{natbib}
\usepackage{url,color,booktabs}
\usepackage[bookmarksnumbered=true,bookmarksopen=true]{hyperref}

\definecolor{DarkRed}{rgb}{0.545,0,0}
\hypersetup{
  breaklinks=true, colorlinks=true, linkcolor=black,
  citecolor=black, urlcolor=DarkRed,
  pdftitle={Very fast, approximate counterfactual explanations for decision forests},
  pdfauthor={Miguel {\'A}.\ Carreira-Perpi{\~n}{\'a}n and Suryabhan Singh Hada}
}

\def \ifempty#1{\def\temp{#1} \ifx\temp\empty }

\newcommand{\A}{\ensuremath{\mathbf{A}}}
\newcommand{\B}{\ensuremath{\mathbf{B}}}

\newcommand{\RR}{\ensuremath{\mathbf{R}}}

\newcommand{\X}{\ensuremath{\mathbf{X}}}

\renewcommand{\aa}{\ensuremath{\mathbf{a}}}
\renewcommand{\b}{\ensuremath{\mathbf{b}}}

\newcommand{\w}{\ensuremath{\mathbf{w}}}
\newcommand{\x}{\ensuremath{\mathbf{x}}}

\newcommand{\0}{\ensuremath{\mathbf{0}}}
\newcommand{\1}{\ensuremath{\mathbf{1}}}

\newcommand{\bbR}{\ensuremath{\mathbb{R}}}

\newcommand{\calO}{\ensuremath{\mathcal{O}}}

\newcommand{\calS}{\ensuremath{\mathcal{S}}}

\newcommand{\abs}[2][]{%
  \ifempty{#1} {\left\lvert#2\right\rvert} \else {#1\lvert#2#1\rvert} \fi}
\newcommand{\norm}[2][]{%
  \ifempty{#1} {\left\lVert#2\right\rVert} \else {#1\lVert#2#1\rVert} \fi}

\newcommand{\caja}[4][1]{{%
    \renewcommand{\arraystretch}{#1}%
    \begin{tabular}[#2]{@{}#3@{}}%
      #4%
    \end{tabular}%
    }}

{%
\begin{list}{#1}{
\vspace{-\topsep}
\vspace{-\partopsep}
\setlength{\itemindent}{0cm}
\setlength{\rightmargin}{0cm}
\setlength{\listparindent}{0cm}
\settowidth{\labelwidth}{#1}
\setlength{\leftmargin}{\labelwidth}
\addtolength{\leftmargin}{\labelsep}
\setlength{\itemsep}{0cm}
}%
}%
{%
\end{list}
\vspace{-\topsep}
\vspace{-\partopsep}
}

{\begin{enumerate}%
}%
{\end{enumerate}}

\DeclareMathOperator*{\argmin}{arg\,min}

\graphicspath{{grf/},{grf-miguel/}}

\AtBeginDvi{}

\title{Very fast, approximate counterfactual explanations \\ for decision forests\thanks{A shorter version of this paper appears as \cite{CarreirHada23a}.}}
\author{
  Miguel {\'A}.\ Carreira-Perpi{\~n}{\'a}n \hspace{5ex} Suryabhan Singh Hada \\
  Dept.\ of Computer Science \& Engineering, University of California, Merced \\
  {\url{http://eecs.ucmerced.edu}}
}

\date{March 5, 2023}

\begin{document}

\maketitle

\begin{abstract}

  We consider finding a counterfactual explanation for a classification or regression forest, such as a random forest. This requires solving an optimization problem to find the closest input instance to a given instance for which the forest outputs a desired value. Finding an exact solution has a cost that is exponential on the number of leaves in the forest. We propose a simple but very effective approach: we constrain the optimization to only those input space regions defined by the forest that are populated by actual data points. The problem reduces to a form of nearest-neighbor search using a certain distance on a certain dataset. This has two advantages: first, the solution can be found very quickly, scaling to large forests and high-dimensional data, and enabling interactive use. Second, the solution found is more likely to be realistic in that it is guided towards high-density areas of input space.

\end{abstract}

\section{Introduction}
\label{s:intro}

A counterfactual explanation (CE) seeks the minimum change to an input instance that will result in a desired outcome under a given predictive model. For example, ``reducing your weight by 10 kg will reduce your risk of stroke by 80\%'' (regression) or ``you will be eligible for the loan if you increase your annual salary by \$10k'' (classification). CEs extend the use of a machine learning model beyond just prediction to querying about potential scenarios. This is especially relevant in applications where interpretability or explainability is important, such as in financial, legal, human resources, government or health models. It can also make it possible to audit a model to find errors or bias, and to have an objective measure of the importance of the input features. CEs are also formally equivalent to adversarial attacks, but the latter have a different motivation: they seek to trick a model into making the wrong prediction by making imperceptible changes to the input.

CEs can be naturally formulated as an optimization problem over the input instance of the form ``minimize the distance to a source instance subject to the model predicting a desired outcome''. Here, we consider as model an ensemble of decision trees (a decision forest). We consider both axis-aligned trees, which are widely used in Random Forest \cite{Breiman01a}, AdaBoost \cite{FreundSchapir97a} and Gradient Boosting \cite{Friedm01a}, but also oblique trees, which achieve state-of-the-art accuracy using fewer and shallower trees \cite{ZharmagCarreir20a,GabidolCarreir22a,Carreir_23a}. The optimization problem is difficult because forests define a piecewise constant predictive function, so gradients are not applicable. The number of constant-value regions is exponential on the size of the forest (number of leaves and number of trees), so exhaustive search approaches (even making use of clever pruning and engineering heuristics) will not be able to scale to real-world forests, for which the number of leaves per tree and the number of trees each run into hundreds or thousands.

We propose a simple but effective approach: to limit the search to the set of regions containing actual (training) data points. This makes the search extremely fast, producing a good, feasible CE estimate in less than a second (for axis-aligned forests) or a few seconds (for oblique forests) even for the largest problems we experimented with. A secondary advantage is that it tends to produce realistic CEs, since the live regions can be seen as a nonparametric density estimate built in the forest. In section~\ref{s:geom} we study the geometry of the forest predictive function and the number of nonempty and live regions. In sections~\ref{s:LIRE}--\ref{s:LIRE-effic} we give our algorithm (LIRE) and evaluate it in section~\ref{s:expts}.

\section{Related work}
\label{s:related}

Much of the work about counterfactual explanations, particularly in the guise of adversarial attacks, has focused on deep neural nets \cite{Szeged_14a,Goodfel_15a}. The optimization here is relatively easy because gradients of the model are available (at least approximately, with ReLU activations). That said, with a heavily nonlinear function such as a neural net, the optimization problem may have multiple local minima, some of them not even feasible (i.e., failing to produce the desired prediction). This makes finding a good solution difficult. Some work is agnostic to the model \cite{Karimi_20a,Guidot_19a,WhiteDavila20a}, requiring only function evaluations (and possibly constructing a mimic of the model), but this restricts its performance severely in terms of computational efficiency and quality or feasibility of the result.

For decision trees, whether axis-aligned or oblique, the problem can be solved exactly and efficiently \cite{CarreirHada21a,HadaCarreir21b} by finding an optimal CE within each leaf's region and picking the closest one. This scales nicely because an individual tree, particularly an oblique tree, will rarely have more than several hundred leaves. For a decision forest, the problem is NP-hard. This was shown for a special case (a binary classification axis-aligned tree with discrete features) by reduction from a maximum coverage problem \cite{Yang_06c}, and for a more general case by reduction from a DNF-MAXSAT problem \cite{Cui_15a}.

Some works use heuristic approaches to solve the CE for forests. \citet{Lucic_22a} use a gradient-based algorithm that approximates the splits of the decision trees with sigmoid functions. \citet{Tolomei_17a} propose an approximate algorithm based on propagating the source instance down each tree towards a leaf. As seen in our experiments, these approaches are slow for large forests and often fail to find a feasible solution as the number of features or trees grows.

Several approaches \cite{Cui_15a,Kanamor_20a,ParmenVidal21a} are based on encoding, more or less efficiently, the CE problem as a mixed-integer program (where binary variables encode the path that the instance takes through each tree) and then solving this using an existing solver (typically based on branch-and-bound). This capitalizes on the significant advances that highly-optimized commercial solvers have incorporated in the last decades (which, unfortunately, are not yet available to free or open-source solvers). This is guaranteed to find the global optimum but only if the solver terminates its essentially brute-force search. Even with highly efficient solvers, the optimistic claims in some of these papers about scalability to large datasets and forests just do not hold up, as seen in our experiments.

Finally, several approaches (not necessarily for forests) seek to generate CEs that are more plausible or realistic \cite{Russel19a,Ustun_19a,Karimi_20a,Kanamor_20a,Mothil_20a,VanloovKlaise21a,ParmenVidal21a}. They do this by adding distances to a set of training instances as a penalty to the cost function to encourage the solution to be close to the training data.

\section{Counterfactual explanation problem: definition, geometry and complexity}
\label{s:geom}

We define the counterfactual explanation (CE) problem as the following optimization:
\begin{equation}
  \label{e:CE}
  \min_{\x \in \bbR^D}{ d(\x,\overline{\x}) } \quad \text{s.t.} \quad F(\x) \in \calS.
\end{equation}
Here, $\overline{\x} \in \bbR^D$ is the \emph{source instance} (whose prediction under $F$ we wish to change), $\x \in \bbR^D$ is the \emph{solution instance}, and $d(\cdot,\cdot)$ is a \emph{distance} in input (feature) space, which measures the cost of changing each feature. We will focus primarily on the $\ell^2_2$ or $\ell_1$ distances; weighted distances can be handled by appropriately rescaling the features. $F$ is the \emph{predictive function} of the model (a decision forest in our case), which maps an input instance \x\ to either a value in $\{1,\dots,K\}$ for multiclass classification, or to a real value for regression (we can also include classification here if the forest output is a class probability). Finally, \calS\ is a set of target predictions, i.e., we want \x's prediction to be a value in \calS. For example, for classification \calS\ can be a specific class (or a subset of classes) in $\{1,\dots,K\}$; for regression, \calS\ can be an interval (e.g.\ $F(\x) \ge 7$) or a set of intervals. We may also have constraints on \x.

A decision forest is an ensemble of decision trees. We consider two types of trees: axis-aligned, where each decision node $i$ has the form ``$\smash{x_{d(i)}} \ge \theta_i$'' for some feature $d(i) \in \{1,\dots,D\}$ and bias $\theta_i \in \bbR$; and oblique trees, where each decision node has the form ``$\w^T_i \x \ge w_{i0}$'' for some weight vector $\w_i \in \smash{\bbR^D}$ and bias $w_{i0} \in \bbR$. In both cases, each leaf of a tree outputs a constant value (class label in $\{1,\dots,K\}$ or value in \bbR). A forest of $T$ trees computes its prediction $F(\x)$ by finding the leaf $l_t$ that \x\ reaches in each tree $t \in \{1,\dots,T\}$ and applying a function to the leaf outputs (usually the majority vote for discrete labels or the average for real values). The forest is trained on a dataset using an algorithm to learn individual trees (such as CART, C4.5 or any of its variations) and an ensemble mechanism (bagging and random feature subsets in Random Forests, reweighted training set in AdaBoost, residual error fitting in Gradient Boosting, etc.\@) \cite{Hastie_09a}. Although the vast majority of work on forests uses axis-aligned trees, here we also consider forests of oblique trees. These can be learned with any ensemble mechanism using as base learner the Tree Alternating Optimization (TAO) algorithm \cite{CarreirTavall18a,Carreir22a,Zharmag_21b}, and have been recently shown to outperform axis-aligned forests in accuracy while resulting in forests having fewer and shallower trees \cite{CarreirZharmag20a,ZharmagCarreir20a,GabidolCarreir22a,Gabidol_22a,Carreir_23a}. This is important here because, as shown later, oblique forests need to search far fewer regions. That said, the details of how a forest was constructed are irrelevant here. All we need is to be able to apply the forest to an input to compute two things: which leaf it reaches in each tree, and the forest output.

\subsection{Geometry of the forest predictive function $F$}

A single tree with $L$ leaves partitions $\bbR^D$ into $L$ regions, since an input \x\ reaches exactly one leaf, and each region outputs a constant value. For an axis-aligned tree, each region is a box and can be put in the form $\aa \le \x \le \b$ elementwise, where $\aa,\b \in \bbR^D$ contain the lower and upper bounds (including $\pm \infty$), respectively; this can be obtained from the (feature,bias) pairs in the decision nodes in the path from the root to the leaf. For an oblique tree, each region is a convex polytope bounded by the hyperplanes at the decision nodes in the root-leaf path.

A forest with $T$ trees (where tree $t$ has $L_t$ leaves) partitions $\bbR^D$ into at most $L_1 L_2 \cdots L_T$ regions ($L^T$ if each tree has $L$ leaves), since an input \x\ reaches exactly one leaf in each tree. We can encode each region as a tuple $(l_1,\dots,l_T)$ indicating the leaf reached in each tree. Hence, each region is the intersection of exactly $T$ leaf regions, and it is a box for axis-aligned trees and a convex polytope for oblique trees. In each region, the forest output is constant, so the forest predictive function $F$ is piecewise constant. Although many tuples $(l_1,\dots,l_T)$ result in empty intersections, the number of (nonempty) regions is still exponential in general.

\subsection{Number of regions in $F$ in practice}

The fact that $F$ is piecewise constant means that problem~\eqref{e:CE} can be solved exactly by enumerating all regions that satisfy the constraint $F(\x) \in \calS$, finding the CE in each region%
\footnote{This requires minimizing $d(\x,\overline{\x})$ over the region. As shown in \cite{CarreirHada21a}, for axis-aligned trees, this is a box, and the exact solution is given, separately along each dimension, by the median of 3 numbers: the lower and upper bound of the box, and $\overline{x}_d$. For oblique trees, the region is a polytope, and the exact solution results from solving a quadratic program (QP) for the $\ell_2$ distance and a linear program (LP) for the $\ell_1$ distance.},
and returning the one with lowest distance to $\overline{\x}$. This was the approach in \cite{CarreirHada21a,HadaCarreir21b} for single tree models, where it works very well because the number of leaves is always relatively small. But, how many nonempty regions can we expect with a forest, and how far is that from the upper bound $L^T$? It is difficult to answer this in general for practical forests, which are the result of a complex optimization algorithm, so we estimate this empirically. We can enumerate all nonempty regions with the constructive algorithm of fig.~\ref{f:pseudocode-LIRE-set} (left). This proceeds sequentially to construct a sparse $t$-way tensor $I_t$, where $I_t(l_1,\dots,l_t) = 1$ if tuple $(l_1,\dots,l_t)$ defines a nonempty intersection and 0 otherwise. $I_t$ is constructed by intersecting every nonempty region in $I_{t-1}$ with every leaf in tree $t$. Its correctness relies on the fact that if $I_{t-1}(l_1,\dots,l_{t-1}) = 0$ then $I_t(l_1,\dots,l_{t-1},l_t) = 0$ for any leaf $l_t$. The final tensor $I_T$ has $L_1\cdots L_T$ entries, but only those for which $I_T = 1$ are nonempty. The number of regions in $I_t$ grows monotonically with $t$ because, if $I_{t-1}(l_1,\dots,l_{t-1}) = 1$ then $I_t(l_1,\dots,l_t) = 1$ for at least one leaf $l_t$ in tree $t$ (since the leaves of each tree form a partition of the input space).

\begin{figure}[t]
  \centering
  \psfrag{estimators}{}
  \psfrag{leaves}[b][b]{Upper bound}
  \psfrag{brest cancer}[l][l]{Breast cancer}
  \psfrag{spambase}[l][l]{Spambase}
  \psfrag{letter}[l][l]{Letter}
  \psfrag{mnist}[l][l]{MNIST}
  \psfrag{adult}[l][l]{Adult}
  \begin{tabular}{@{\hspace{0.03\linewidth}}c@{\hspace{0.01\linewidth}}c@{}}
    \psfrag{regions}[cb][b]{\caja{b}{c}{axis-aligned trees \\[1ex] \# regions}}
    \includegraphics*[width=.48\linewidth]{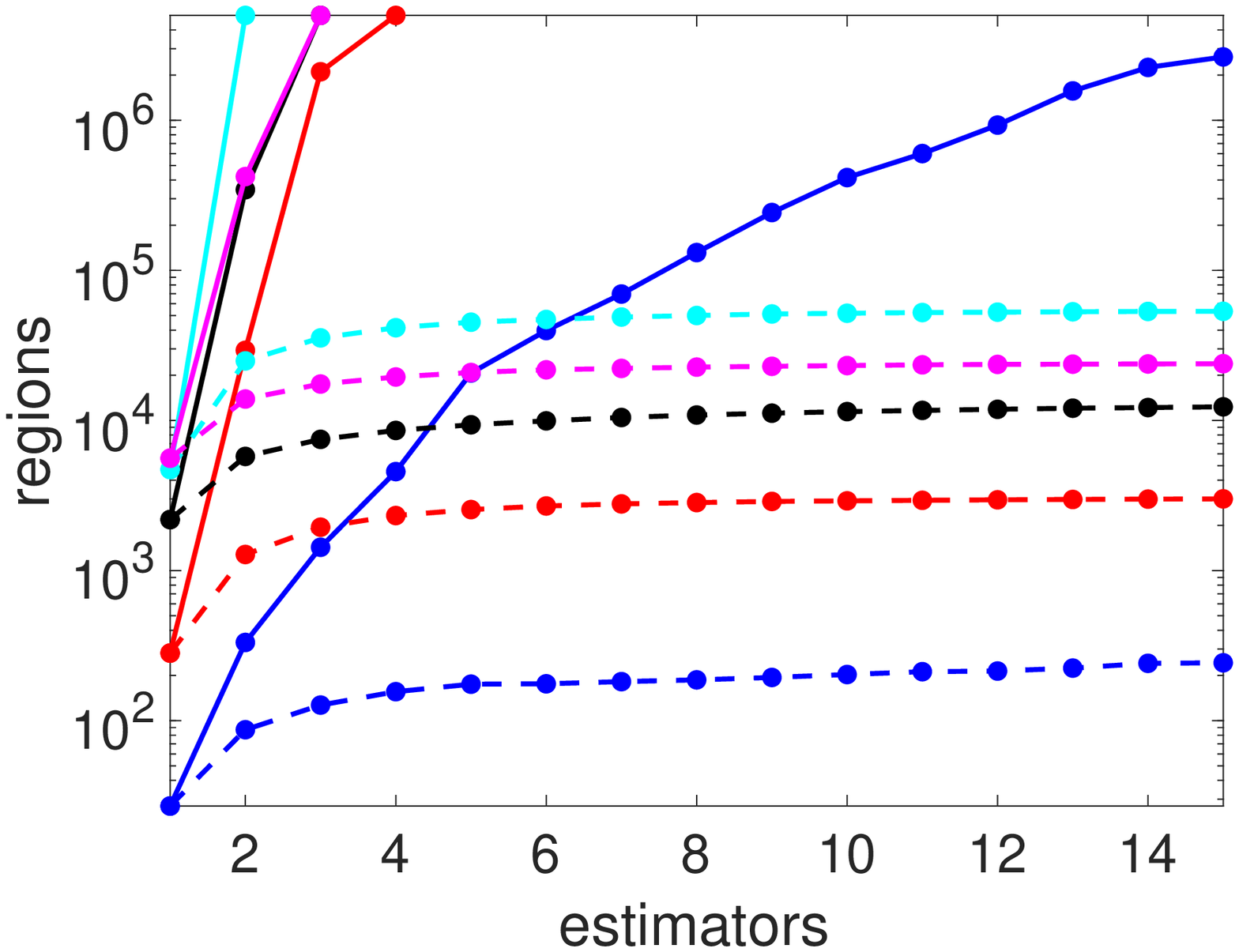}&
    \includegraphics*[width=.48\linewidth]{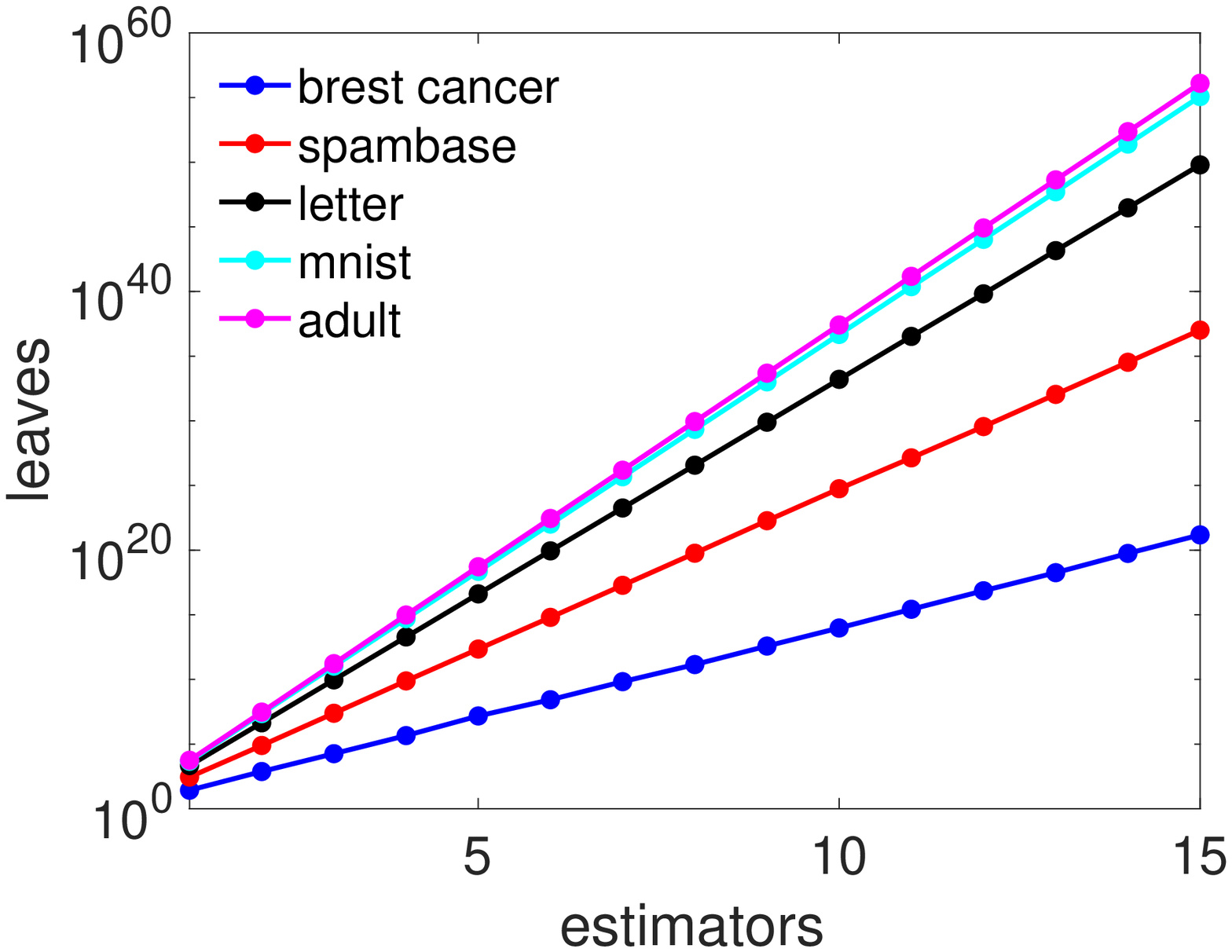} \\
    \psfrag{estimators}[c][][1]{\# trees}
    \psfrag{regions}[cb][b]{\caja{b}{c}{oblique trees \\[1ex] \# regions}}
    \includegraphics*[width=.48\linewidth]{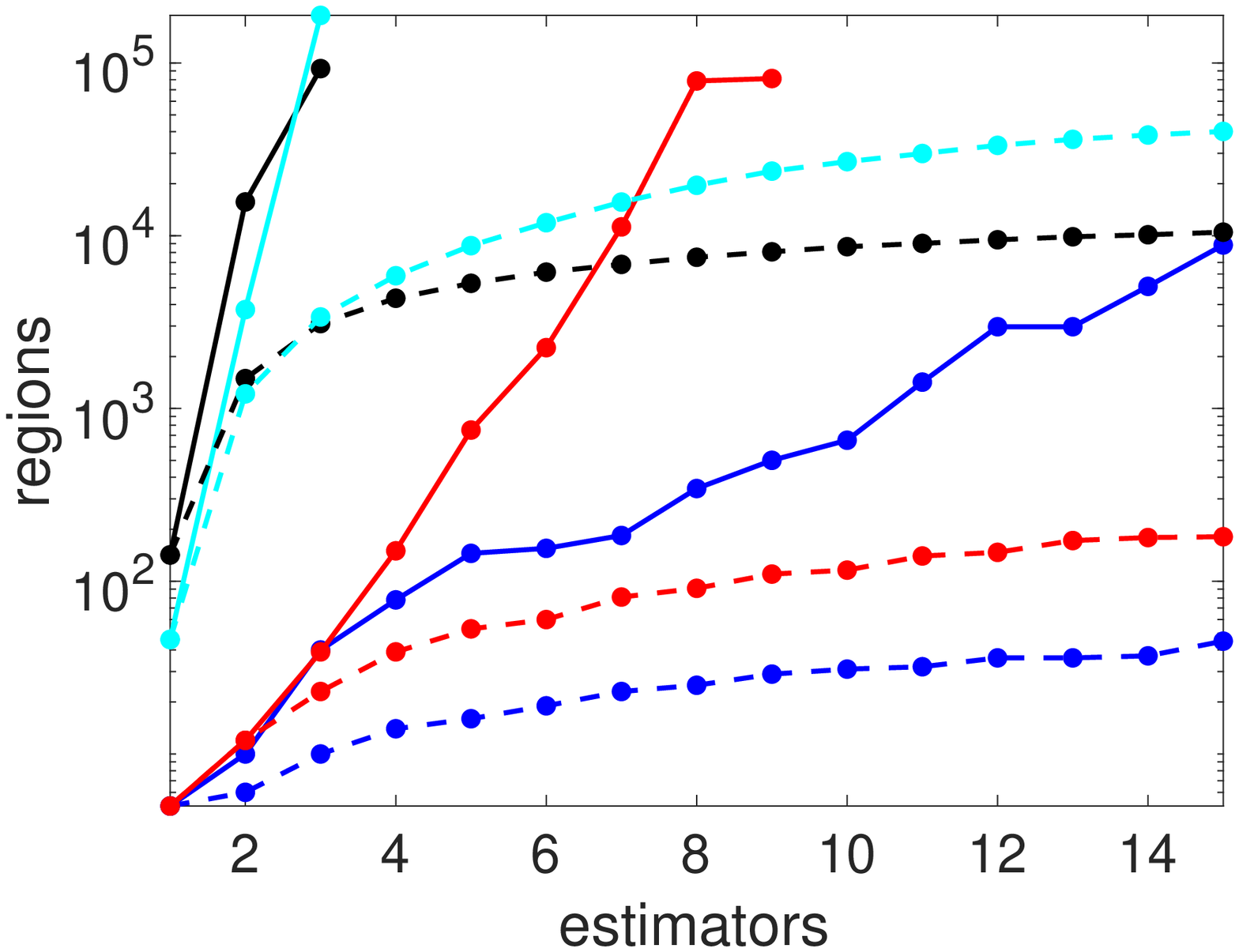}&
    \psfrag{estimators}[c][][1]{\# trees}
    \includegraphics*[width=.48\linewidth]{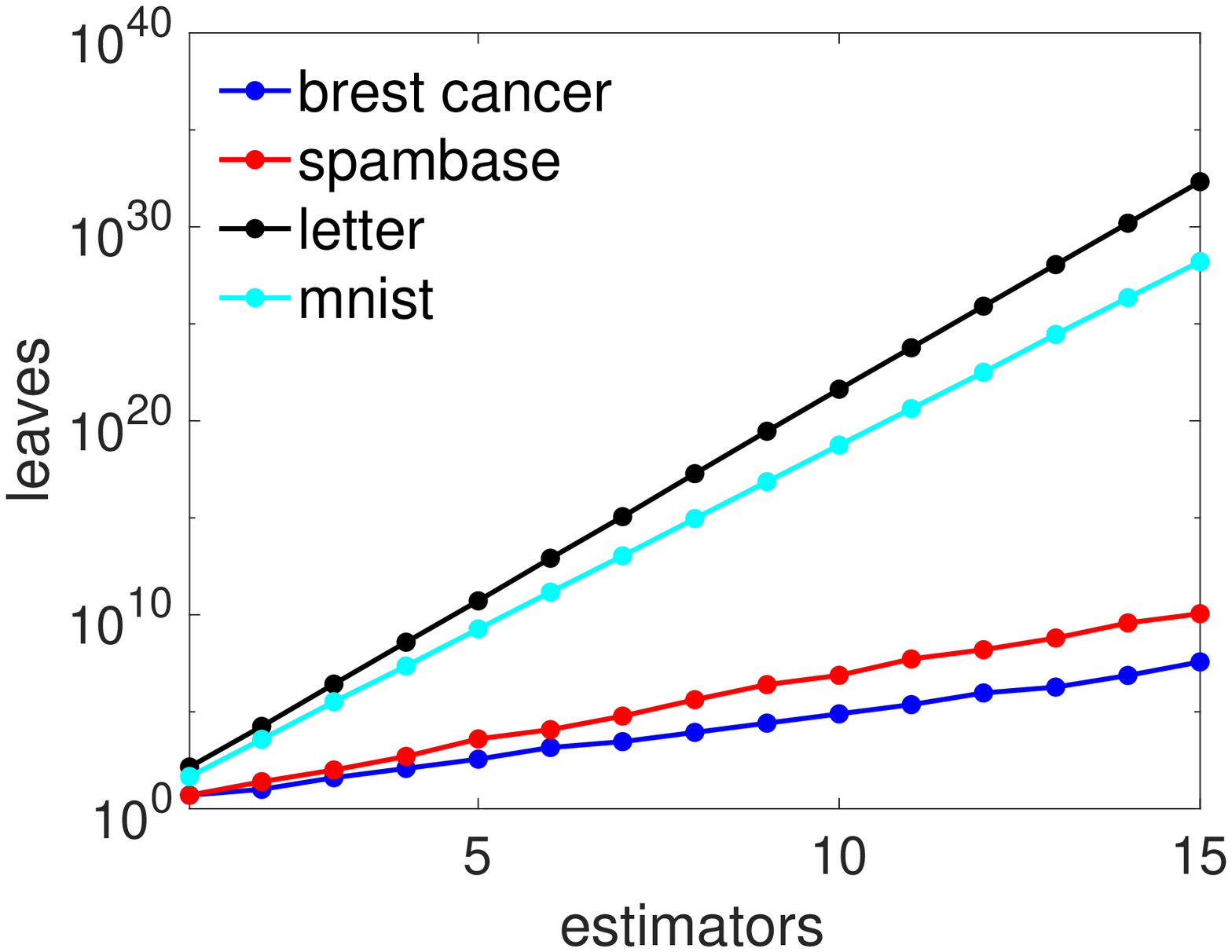}
  \end{tabular}
  \caption{Growth of the number of regions of a forest as a function of the number of trees $T$, for different datasets, for axis-aligned trees (top panels) and oblique trees (bottom panels). Within each pair of panels, on the left panel we plot the number of nonempty regions (solid lines) and of live regions (dashed lines); on the right panel, the upper bound for the number of nonempty regions. All regions are capped to a maximum of $5 \cdot 10^6$ (axis-aligned) and $10^6$ (oblique). The axis-aligned forests use fully-grown trees with an average depth of 8.9, 33.5, 27.6, 35.1 and 48.9, for Breast cancer, Spambase, Letter, MNIST and Adult, respectively. The oblique forests have a fixed depth of 8.}
  \label{f:num-regions}
\end{figure}

Fig.~\ref{f:num-regions} shows the results for Random Forests \cite{Breiman01a} on several small datasets, for which it is computationally feasible to count the regions. The actual number of regions depends in a complex way on the dataset (size and dimensionality) and type of forest. For axis-aligned forests, the number of nonempty regions, while far smaller than the upper bound, does grow exponentially quickly and exceeds a million for just a handful of trees. For oblique forests, the growth is significantly slower but still exponential. This shows than an exhaustive search, even with clever speedups, will be intractable unless the forest is impractically small.

\begin{figure}[t]
  \centering
  \setlength{\fboxsep}{1ex}
  \framebox{%
    \begin{minipage}[t]{0.45\linewidth}
      \begin{tabbing}
        n \= n \= n \= n \= n \= \kill
        \textsc{Enumerating all nonempty regions} \` {\small\textsf{}} \\[1ex]
        {\textbf{input}} forest $F$ with $T$ trees \\
        $I_1 \gets$ all-ones vector of dimension $L_1$ \\
        {\textbf{for}} $t = 2,\dots,T$ \+ \\
        $I_t \gets$ all-zeros sparse array of $L_1 \times \dots \times L_t$ \\
        {\textbf{for}} each $(l_1,\dots,l_{t-1})$ with $I_{t-1}(l_1,\dots,l_{t-1}) = 1$ \+ \\
        {\textbf{for}} $l_t = 1,\dots,L_t$ \+ \\
        {\textbf{if}} $\text{region}(l_1,\dots,l_{t-1}) \cap \text{region}(l_t) \neq \varnothing$ {\textbf{then}} \+ \\
        $I_t(l_1,\dots,l_{t-1},l_t) \gets 1$ \- \- \- \\
        remove $I_{t-1}$ from memory \- \\
        {\textbf{return}} $I_T$
      \end{tabbing}
    \end{minipage}
  }
  \hfill
  \framebox{%
    \begin{minipage}[t]{0.45\linewidth}
      \begin{tabbing}
        n \= n \= n \= n \= n \= \kill
        \textsc{Enumerating all live regions} \` {\small\textsf{}} \\[1ex]
        {\textbf{input}} forest $F$ with $T$ trees, dataset $\X_{D\times N}$ \\
        $I,\ Y \gets$ all-zeros sparse array of $L_1 \times \dots \times L_T$ \\
        {\textbf{for}} $n = 1,\dots,N$ \+ \\
        {\textbf{for}} $t = 1,\dots,T$ \+ \\
        $l_t \gets$ leaf reached by $\x_n$ in tree $t$ \- \\
        {\textbf{if}} $I(l_1,\dots,l_T) = 0$ {\textbf{then}} \+ \\
        $I(l_1,\dots,l_T) \gets 1$ \\
        $Y(l_1,\dots,l_T) \gets F(l_1,\dots,l_T)$ \- \- \\
        {\textbf{return}} $I$ and $Y$, both sorted by $Y$ value
      \end{tabbing}
    \end{minipage}
  }
  \caption{Pseudocode for finding all nonempty regions (left) and all live regions (right), valid for both axis-aligned and oblique trees. We omit the construction of the arrays \A, \B\ and \RR\ (needed in fig.~\ref{f:pseudocode-LIRE-search}). \emph{Left}: $\text{region}(l_1,\dots,l_t) \equiv \cap^t_{i=1}{ \text{region}(l_i) }$ and $\text{region}(l_i)$ is the input space region of leaf $l_i$ in tree $i$ (defined by the intersection of the decision node hyperplanes along the path from the root to leaf $l_i$). $F(l_1,\dots,l_T)$ means the forest output for $\text{region}(l_1,\dots,l_T)$.}
  \label{f:pseudocode-LIRE-set}
\end{figure}

\section{An approximation: search only over the ``live'' regions}
\label{s:LIRE}

\begin{figure}[t]
  \centering
  \psfrag{s}[][]{$\overline{\x}$}
  \psfrag{x1}[r][r]{$\x^*$}
  \psfrag{x2}[r][r]{$\x'$}
  \begin{tabular}{@{}c@{\hspace{0.01\linewidth}}c@{\hspace{0.01\linewidth}}c@{}}
    & Dataset \\
    & \includegraphics*[width=0.33\linewidth]{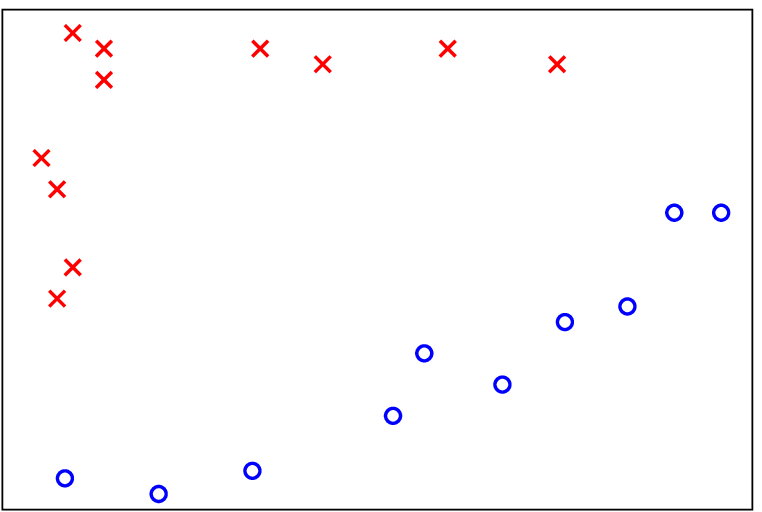} \\[1ex]
    Tree 1 & Tree 2 & Tree 3 \\
    \includegraphics*[width=0.33\linewidth]{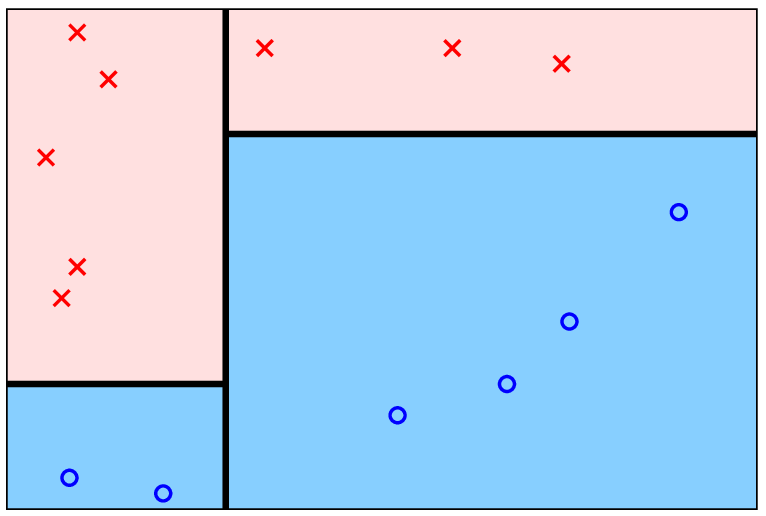} &
    \includegraphics*[width=0.33\linewidth]{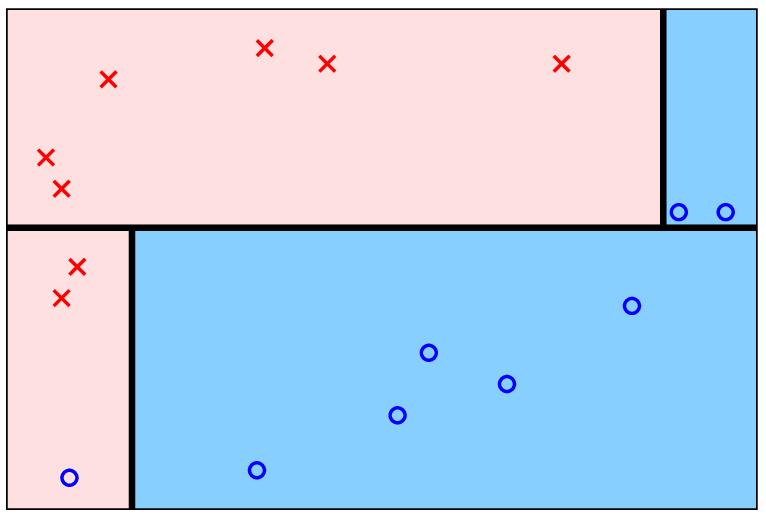} &
    \includegraphics*[width=0.33\linewidth]{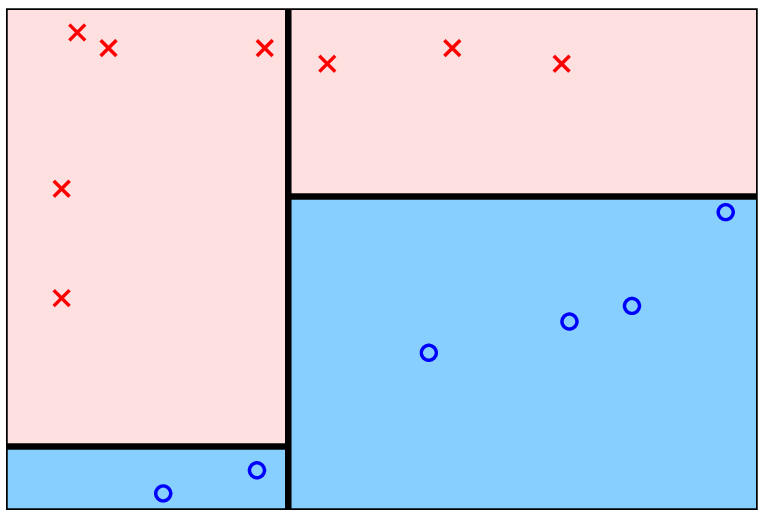} \\[1ex]
    Forest & Exact solution & LIRE solution \\
    \includegraphics*[width=0.33\linewidth]{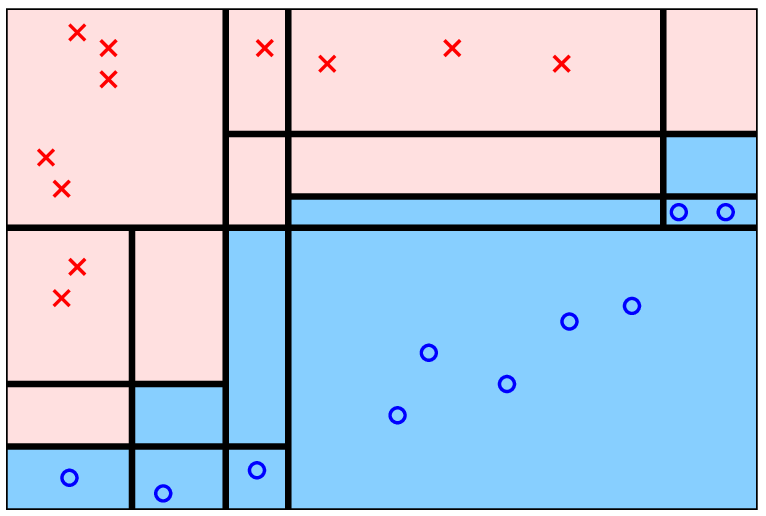} &
    \includegraphics*[width=0.33\linewidth]{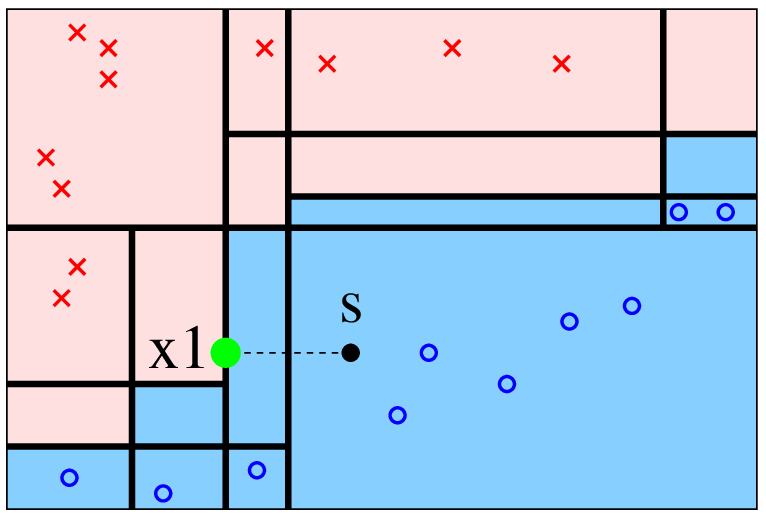} &
    \includegraphics*[width=0.33\linewidth]{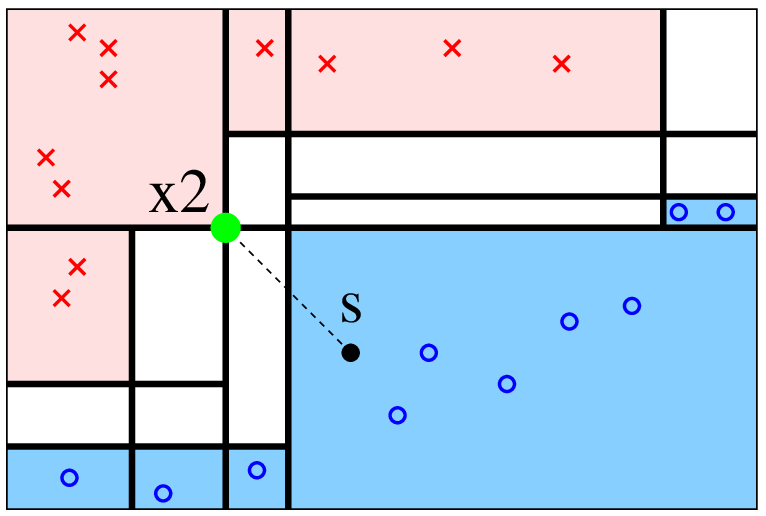}
  \end{tabular}
  \caption{Illustration of the LIRE idea. We show a forest of $T=3$ trees of depth $\Delta=2$, each trained on a subset of the training set. The plots show the regions of each tree and the forest, colored accordingly to the class label they predict. The live regions are those containing at least one data point. For the source instance $\overline{\x}$, the optimal counterfactual explanation is $\x^*$ (searching over all regions) and the approximate one with LIRE is $\x'$ (searching only over the live regions).}
  \label{f:diagram}
\end{figure}

Instead of considering all regions, we restrict the search to only those regions containing at least one actual data point (from the training, validation and test datasets used to train the forest). We call these \emph{live} regions, and call the procedure LIRE (for LIve REgion search). LIRE results in an approximate but fast search and, intuitively, retrieves realistic CEs, as described next. Fig.~\ref{f:diagram} illustrates the idea. Let the dataset have $N$ points and the number of live regions be $M \le N$.

\subsection{Faster computation}

The number of regions reduces from exponential to at most $N$, so the search is far faster and also has a very predictable runtime (unlike, for example, approaches based on mixed-integer programming, which have a wildly variable runtime). The number of live regions $M$ is at most $N$ because multiple points may belong to the same region (this is particularly so with oblique forests). A second reduction in the number of regions to search is due to the constraint $F(\x) \in \calS$ (for example, we may need to search only on regions of one target class).

Fig.~\ref{f:num-regions} shows the number of live regions. The growth behavior is very different from that of the nonempty regions, because it is upper bounded by $N$. For axis-aligned forests, the number of live regions reaches $N$ with just a handful of trees, so we can expect about $N$ regions with any practical forest (with nearly every region containing just one instance). For oblique forests, it takes well over 10 trees to approach $N$ regions, so practical oblique forests (which do not require as many trees as axis-aligned forests) may have quite less than $N$ regions, particularly with large datasets.

As expected, the resulting runtime of LIRE is very small (see experiments). For axis-aligned trees it takes less than 1 second in even our largest experiments; in this case, the search reduces to a special type of nearest-neighbor search, efficiently implementable using arrays and vectorization (see later). For oblique trees, each region requires solving a small QP or LP (having $T \Delta$  constraints on average, where $\Delta$ is the average leaf depth). Although this is more costly, the number of regions in an oblique forest is far smaller. In our experiments this takes at most a couple of minutes.

\subsection{Approximate solution}

The CE found by searching only on the live regions is suboptimal: it has a larger distance than the exact CE. We estimate this in our experiments by comparing with the exact CE (for small forests) and with other existing algorithms for CEs.

It is instructive to consider also an even simpler approximation to CEs: to ignore entirely the forest regions, and search directly in a dataset of instances (say, the training set), but labeled per the forest (not the ground truth). While this is very fast, it is never better than LIRE, because the live regions contain the data instances. In fact, as shown in our experiments, this approach produces CEs with quite a larger distance than LIRE.

\subsection{Realistic solution}

A difficult problem with counterfactual explanations with any type of model (not just forests) is that it is difficult to constrain the search space in problem~\eqref{e:CE} to find realistic instances. Although the input space is defined to be $\bbR^D$, most real-world data live in a manifold or subset of it. Some domain knowledge can be incorporated through simple constraints (e.g.\ a grayscale pixel should be in [0,1]), but this is insufficient to capture the subset of realistic instances. Intuitively, this requires estimating the density distribution of the data, a very hard problem in high dimensions, and then constraining problem~\eqref{e:CE} to that. We can see LIRE in this light as imposing a constraint based on a nonparametric, adaptive kernel density estimate: each live region (a box or polytope) sits on one point and has constant density; all other regions have zero density. The kernel is adaptive, rather than having a given form and bandwidth. This density estimate comes for free with the forest and blends conveniently into the optimization. This makes it more likely that a CE found by searching on the live regions will be more realistic than searching anywhere in the space. 

In summary, LIRE can be seen either as an approximate solution to searching CEs in the entire space, or as an exact solution of a CE subject to lying in high density regions, using a nonparametric density estimate. Either way, LIRE is extremely fast and scales to forests with practically useful sizes.

\section{Efficient implementation of the live region search}
\label{s:LIRE-effic}

\subsection{Constructing the set of live regions (offline)}

Obviously, we do not need to build the sparse tensor $I_T$ and test every nonempty region. All we have to do is feed each input instance to the forest and determine which leaf it reaches in each tree. The resulting region is live. See the pseudocode in fig.~\ref{f:pseudocode-LIRE-set} (right). This is done offline and has a complexity of $\calO(N T \Delta)$ where $\Delta$ is the average depth of a tree. The result is a list of $M \le N$ regions, each encoded by a leaf tuple $(l_1,\dots,l_T)$. For axis-aligned trees, each forest region is a box and can be compactly represented by two vectors $\aa_n,\b_n \in \bbR^D$ with $\aa_n \le \b_n$ elementwise, containing the lower and upper bounds along each dimension (or $\pm \infty$ if unbounded). For oblique trees, each region is defined by the intersection of all the constraints (hyperplanes) along one root-leaf path in each tree; computationally, it is better not to construct this list explicitly, instead reconstructing it on the fly during the live region search.

\subsection{Sorting or indexing the set of live regions (offline)}

Of all $M$ live regions, we need only search in those that satisfy the constraint $F(\x) \in \calS$, which are usually far less than $M$; this greatly accelerates the search. For example, in $K$-class classification, if our target is a specific class, we need only search $\frac{M}{K}$ regions (assuming uniform class populations). Determining the range of target regions to search can be done by a binary search in $\calO(\log{M})$ time if we presort the $M$ regions by their value of $F$; this is useful in regression. In $K$-class classification, we can pre-index the $M$ regions into $K$ groups (one per class) and determine the target in constant time. This supports complex targets such as $F(\x) \in \{1,3,7\}$ or $[2,5] \cup [7,\infty)$. Then, the actual search in the range of target regions is done sequentially as described next.

\subsection{Searching in each target live region}

\paragraph{Axes-aligned trees}

This can be implemented in a way that is very efficient in time and memory through arrays and vectorization, without any need for the tree structures. Firstly \cite{CarreirHada21a}, for any distance that is separable over dimensions (e.g.\ $\ell_1$, $\ell^2_2$, possibly weighted), the solution to ``$\x^* = \argmin_{\x}{ d(\x,\overline{\x}) }$ s.t.\ $\aa \le \x \le \b$'' (where \aa, \b\ are the lower and upper  bounds of a particular box) can be given in closed form as $\x^* = \text{median}(\aa,\b,\overline{\x})$ elementwise. (That is, for each dimension, $x^*$ is $a$ if $x^* \le a$, $b$ if $x^* \ge b$ and $\overline{x}$ otherwise.) However, it is more efficient to compute directly the \emph{distance-to-a-box} $d_{\text{box}}(\overline{\x},\smash{\binom{\aa}{\b}}) \equiv d(\x^*,\overline{\x})$. For the $\ell^2_2$ and $\ell_1$ distances this is:
\begin{align*}
  \label{e:dist-to-a-box}
  \norm{\x^* - \overline{\x}}^2_2 &= \1^T (\max(\aa - \overline{\x},\0) + \max(\overline{\x} - \b,\0))^2 \\
  \norm{\x^* - \overline{\x}}_1 &= \1^T (\max(\aa - \overline{\x},\0) + \max(\overline{\x} - \b,\0))
\end{align*}
where $\max(\cdot,\cdot)$ applies elementwise. This holds by noting that, for each dimension $d = 1,\dots,D$, $\abs{x^*_d - \overline{x}_d} = a_d - \overline{x}_d$ if $a_d - \overline{x}_d \ge 0$, $\overline{x}_d - b_d$ if $\overline{x}_d - b_d \ge 0$, and 0 otherwise. To preserve memory locality, this can be vectorized over the entire array of $\A_{D\times N} = (\aa_1,\dots,\aa_N)$ and $\B_{D\times N} = (\b_1,\dots,\b_N)$ to find the box $n$ with the smallest distance to $\overline{\x}$.%
\footnote{For ex., in Matlab for $\ell^2_2$: {\scriptsize\texttt{[d,n] = min(sum((max(bsxfun(@minus,A,x),0)+max(bsxfun(@minus,x,B),0)).\^{}2,2));}}.}
This shows that, in effect, the problem reduces to \emph{a form of nearest-neighbor search}, where we have a search set of $N$ multidimensional points $\smash{\binom{\aa_n}{\b_n}} \in \bbR^{2D}$ (each representing a box), a query $\overline{\x} \in \bbR^D$, and a distance $d_{\text{box}}(\overline{\x},\smash{\binom{\aa_n}{\b_n}})$ given by the distance-to-a-box. Fig.~\ref{f:pseudocode-LIRE-search} (left) gives the pseudocode.

\paragraph{Oblique trees}

In this case we cannot vectorize using arrays because each region $(l_1,\dots,l_T)$ has an irregular (polytope) shape, given by the constraints for each leaf $l_t$, $t = 1,\dots,T$ (which, in turn, are the constraints in the root-leaf path to $l_t$ in tree $t$). So we have to loop through each region, solve its QP or LP, and return the one with minimum distance; see pseudocode in fig.~\ref{f:pseudocode-LIRE-search} (right). As noted earlier, the advantage with oblique trees is that they use few, shallower trees, so the number of regions is much smaller.

\paragraph{Handling constraints}

In practice with CEs, we often impose additional, linear constraints on \x\ in problem~\eqref{e:CE}. For example, we can fix ($x_d = \cdot$) or bound ($\cdot \le x_d \le \cdot$) the values of some variables (this can also be used to force the solution to be interior to the regions rather than right on the boundary). Such constraints are simply handled individually in each region as in \cite{CarreirHada21a}. For axis-aligned trees, we can preprocess each box offline to shrink it correspondingly by intersecting it with the constraints and eliminating infeasible boxes (which makes the search even faster).

\begin{figure}[t]
  \centering
  \setlength{\fboxsep}{1ex}
  \framebox{%
    \begin{minipage}[t]{0.41\linewidth}
      \begin{tabbing}
        n \= n \= n \= n \= n \= \kill
        \textsc{\caja{t}{l}{Search for closest live region \\ (axis-aligned trees)}} \` {\small\textsf{}} \\[1ex]
        {\textbf{input}} $\A_{D\times M}$, $\B_{D\times M}$, $\overline{\x} \in \bbR^D$ \\
        $\delta \gets \infty$ \\
        {\textbf{for}} $n = 1,\dots,M$ \+ \\
        $\alpha \gets d_{\text{box}}(\overline{\x},\smash{\binom{\aa_n}{\b_n}})$ \\
        {\textbf{if}} $\alpha < \delta$ {\textbf{then}} \+ \\
        $i \gets n$ \\
        $\delta \gets \alpha$ \- \- \\
        $\x^* \gets \argmin_{\x}{ d(\x,\overline{\x}) }$ s.t.\ $\aa_i \le \x \le \b_i$ \\
        $\phantom{\x^*} \ = \text{median}(\aa_i,\b_i,\overline{\x})$ \\
        {\textbf{return}} $i$, $\x^*$, $d(\x^*,\overline{\x})$
      \end{tabbing}
    \end{minipage}
  }
  \hfill
  \hspace{1ex}
  \framebox{%
    \begin{minipage}[t]{0.47\linewidth}
      \begin{tabbing}
        n \= n \= n \= n \= n \= \kill
        \textsc{\caja{t}{l}{Search for closest live region \\ (oblique trees)}} \` {\small\textsf{}} \\[1ex]
        {\textbf{input}} forest of $T$ trees, $\RR_{T\times M}$ \\
        $\delta \gets \infty$ \\
        {\textbf{for}} $n = 1,\dots,M$ \+ \\
        $\alpha \gets \min_{\x}{ d(\x,\overline{\x}) }$ s.t.\ constraints for $R(:,n)$ \\
        {\textbf{if}} $\alpha< \delta$ {\textbf{then}} \+ \\
        $i \gets n$ \\
        $\delta \gets \alpha$ \- \- \\
        $\x^* \gets \argmin_{\x}{ d(\x,\overline{\x}) }$ s.t.\ constraints for $R(:,i)$ \\
        {\textbf{return}} $i$, $\x^*$, $d(\x^*,\overline{\x})$
      \end{tabbing}
    \end{minipage}
  }
  \caption{Pseudocode for the search for the closest live region to a source instance $\overline{\x}$ for axis-aligned (left) and oblique trees (right). We assume there are $M$ target regions which have been preselected into the lower/upper bound arrays \A\ and \B\ (for axis-aligned trees) or the array \RR\ (for oblique trees). $R(t,n)$ contains the index of the leaf in tree $t$ that participates in region $n$ and $R(:,n)$ stands for $\{R(1,n),\dots,R(T,n)\}$. Region $n$ is defined by the constraints (halfspaces) in the decision nodes from $t$'s root to the parent of leaf $R(t,n)$, for each leaf in $R(:,n)$. $d_{\text{box}}(\overline{\x},\smash{\binom{\aa_i}{\b_i}}) = d(\overline{\x},\text{median}(\aa_i,\b_i,\overline{\x}))$ represents the distance-to-a-box.}
  \label{f:pseudocode-LIRE-search}
\end{figure}

\subsection{Further accelerating the search}

The exhaustive search over all live regions is very fast. For example, sequentially searching $M = 10^6$ points in $D =$ 100 dimensions takes less than a second on a laptop. However, for very large data sets (say, a billion points), this will be too slow. One way to speed this up while finding the exact solution is by parallelizing the search, which can be done trivially over subsets of regions. Another one is by using a search tree, decorated at each node with bounding boxes, to prune sets of regions that are guaranteed not to be optimal. If we allow the search to be inexact, a simple approach is to use live regions for a random sample of data points. It should also be possible to adapt fast techniques to find approximate nearest neighbors in high dimensions. Note that LIRE is an anytime algorithm in that we can stop at any time and return a feasible solution.

\subsection{Computational complexity}

As noted earlier, determining the range of regions we need to search (say, the regions with a desired target class) takes negligible time: a logarithmic binary search if the list of regions has been sorted by forest output, or a constant-time lookup if it has been indexed. The cost is dominated by the exhaustive search over the range of regions. For axis-aligned trees, this is $\calO(MD)$ with $M$ regions and $D$ features, with a small constant factor due to the distance-to-a-box computation. For oblique trees, we have to solve $M$ QPs ($\ell^2_2$) or LPs ($\ell_1$). Each has $D$ variables and $T \Delta$ constraints on average (assuming an average leaf depth $\Delta$). In both cases, the search can be trivially parallelized.

\section{Experiments}
\label{s:expts}

In this section, we used Random Forests (where each tree is grown in full, i.e., not pruned), with individual trees trained by CART \cite{Breiman_84a} if axis-aligned and by TAO \cite{CarreirTavall18a,Carreir22a} if oblique. All runtimes were obtained in a single core (without parallel processing). The appendices give details about the experiments, as well as more results (e.g.\ with AdaBoost forests, and the training and test error of the forests we trained). Here, we comment on the main results.

\subsection{LIRE as an approximate CE}

\begin{figure}[p]
  \centering
  \psfrag{exact}[l][l][0.8]{all regions}
  \psfrag{trptrestriction}[l][l][0.8]{LIRE}
  \psfrag{whatif}[l][bl][0.8]{\caja[0.2]{c}{c}{ dataset search}}
  \begin{tabular}{@{\hspace{0.03\linewidth}}c@{\hspace{0.0\linewidth}}c@{\hspace{0.0\linewidth}}c@{\hspace{0.0\linewidth}}c@{}}
    {Breast cancer} & {Spambase} & {Letter} &{Adult}\\
    \psfrag{errors}[c][]{\raisebox{3ex}[0pt][0pt]{\makebox[0pt]{\hspace{-8ex}axis-aligned trees}}avg.\ distance}
    \psfrag{estimators}{}
    \includegraphics*[width=.2425\linewidth]{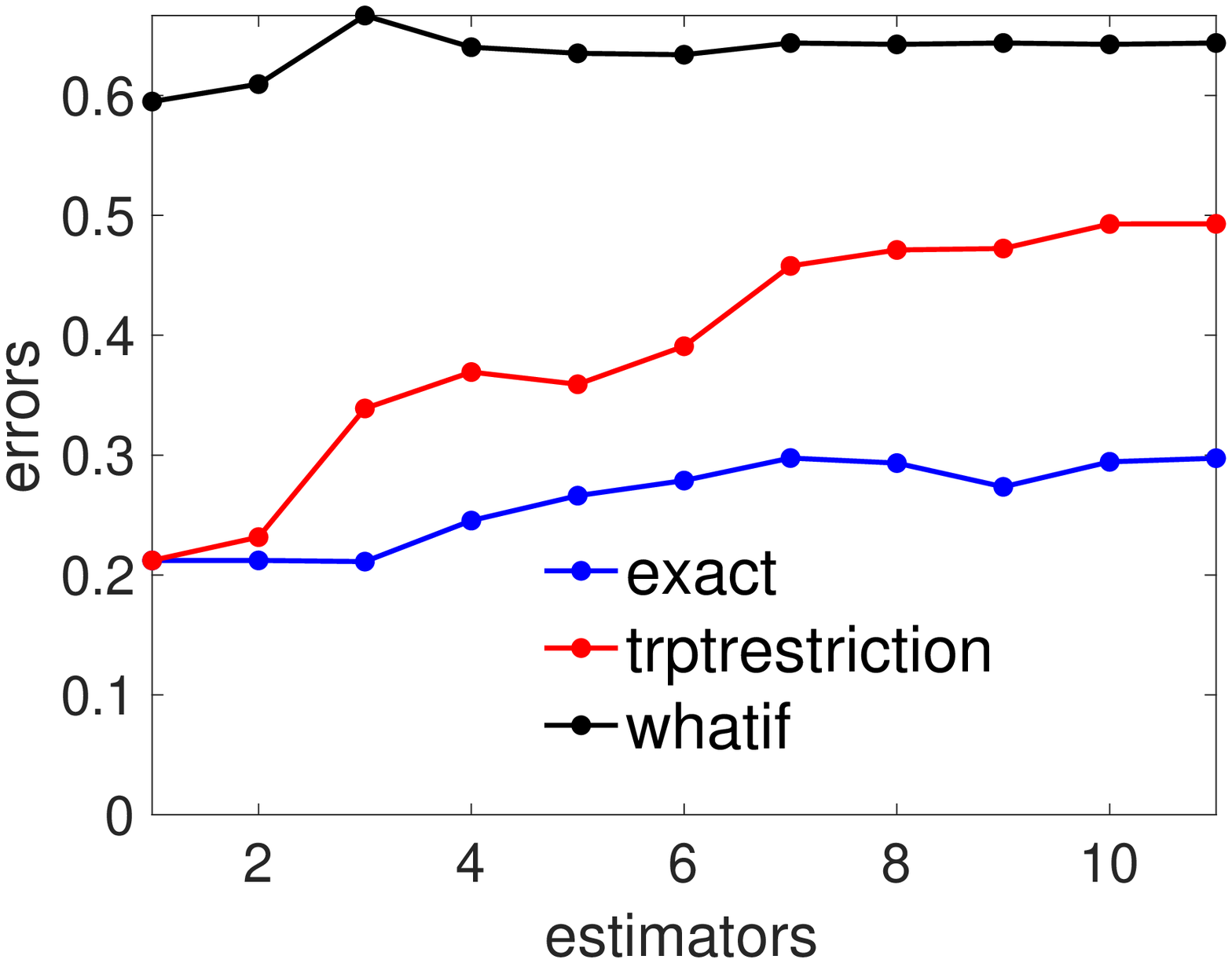}&
    \psfrag{estimators}{}
    \psfrag{errors}{}
    \psfrag{runtime}{}
    \includegraphics*[width=.2425\linewidth,bb=13 4 512 393]{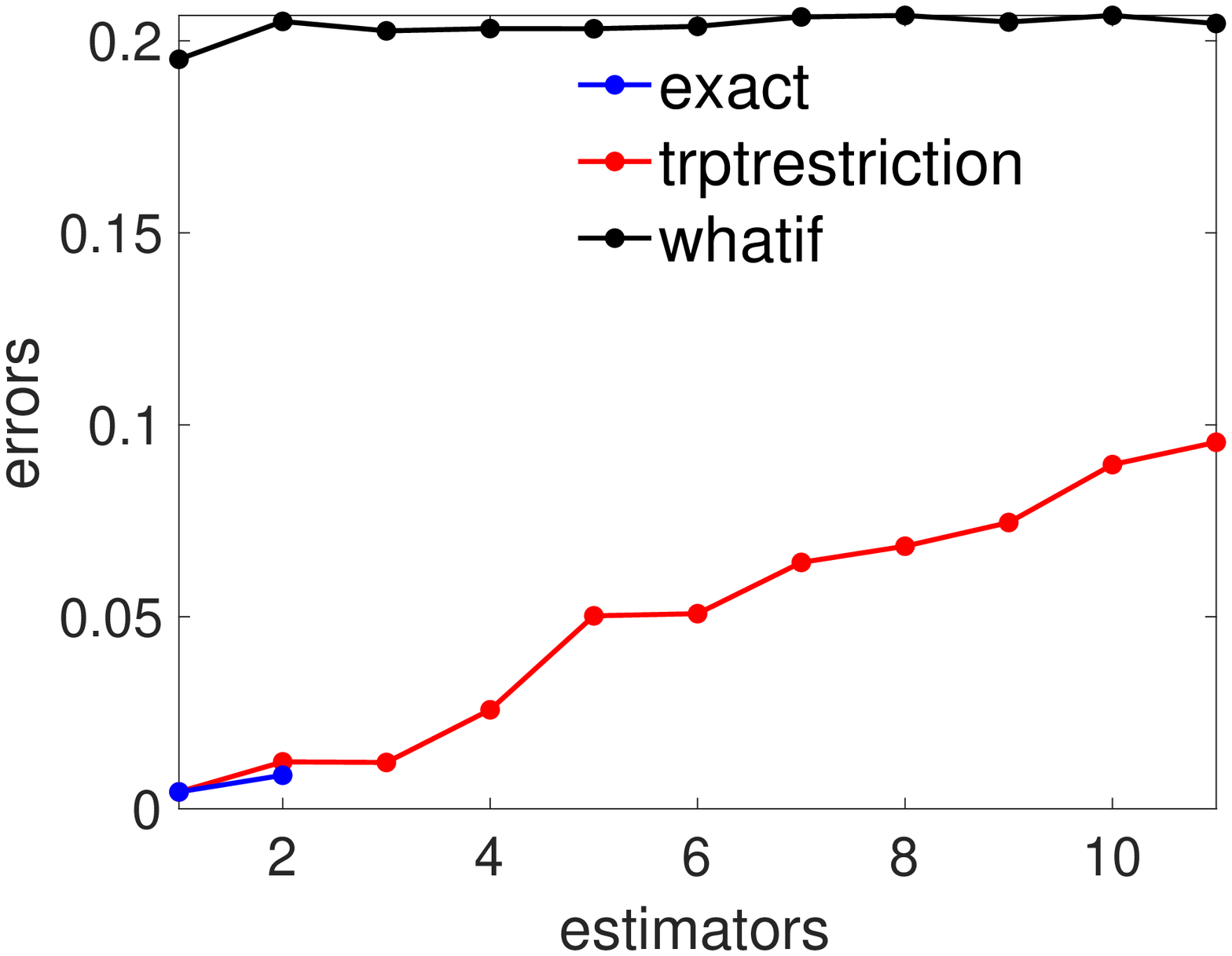}&
    \psfrag{estimators}{}
    \psfrag{errors}{}
    \psfrag{runtime}{}
    \includegraphics*[width=.2425\linewidth]{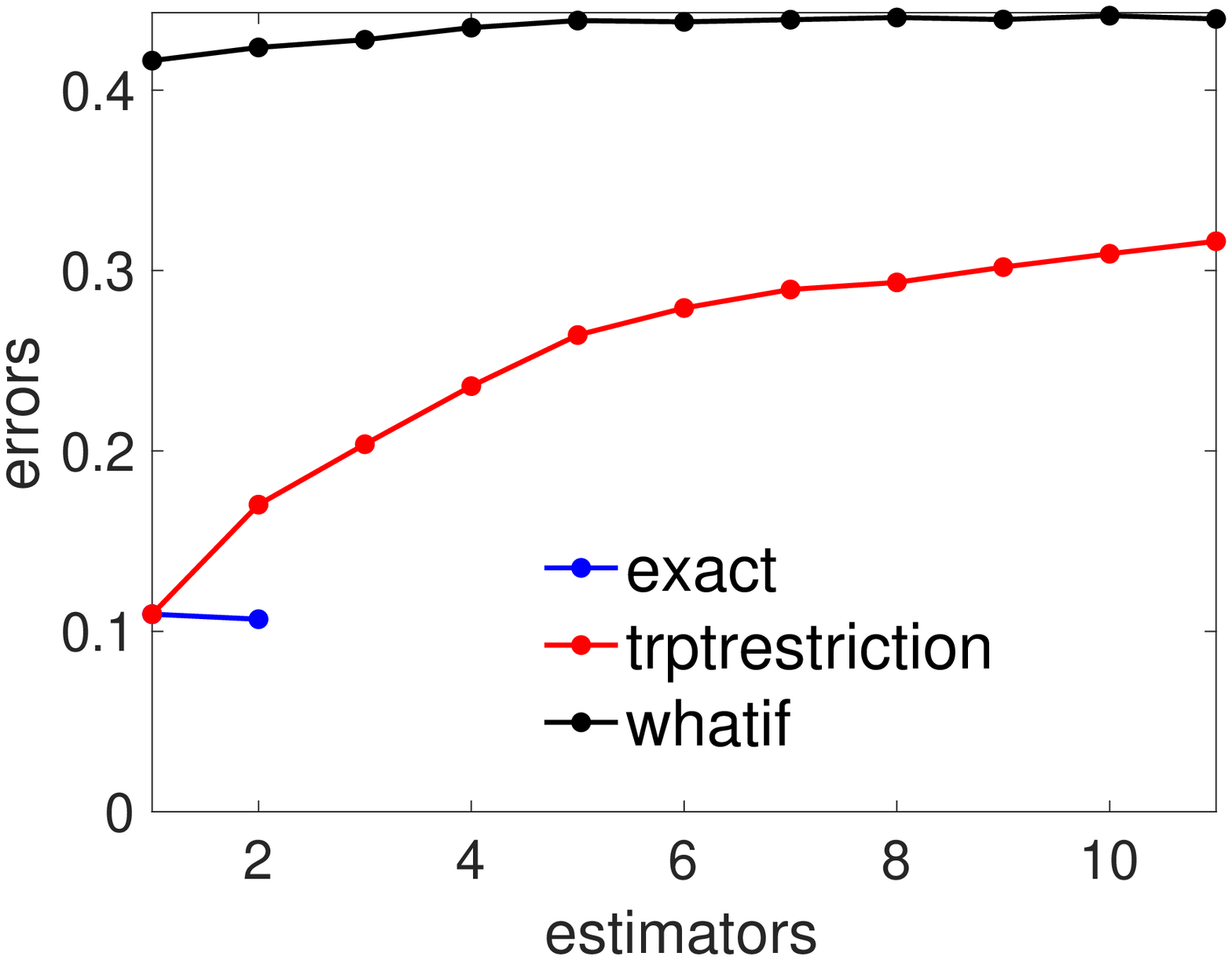}&
    \psfrag{estimators}{}
    \psfrag{errors}{}
    \psfrag{runtime}{}
    \includegraphics*[width=.2425\linewidth,bb=13 4 512 393]{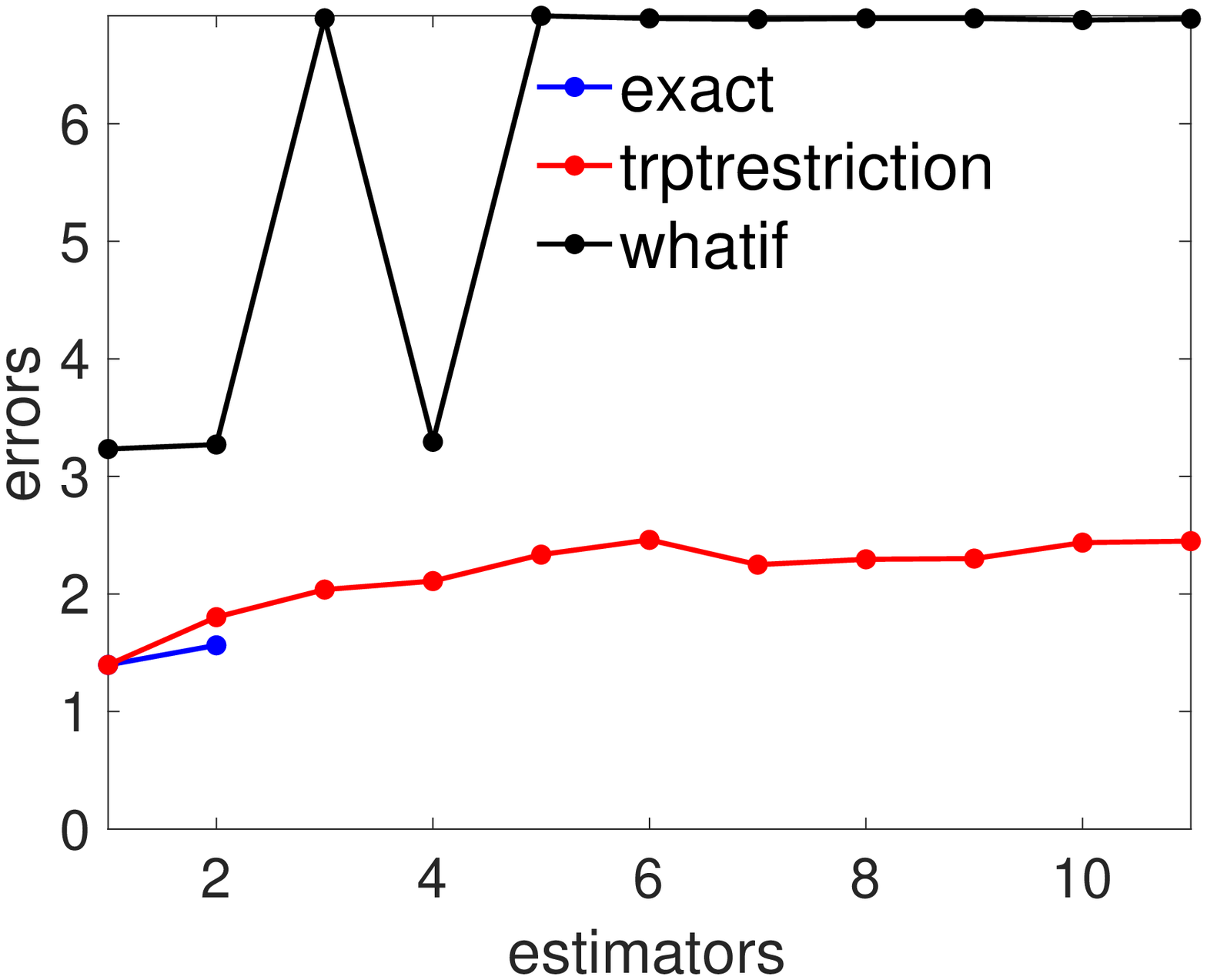}\\[-2ex]
    \psfrag{runtime}[c][]{avg.\ runtime (s)}
    \psfrag{estimators}[c][][1]{\# trees}
    \includegraphics*[width=.2425\linewidth]{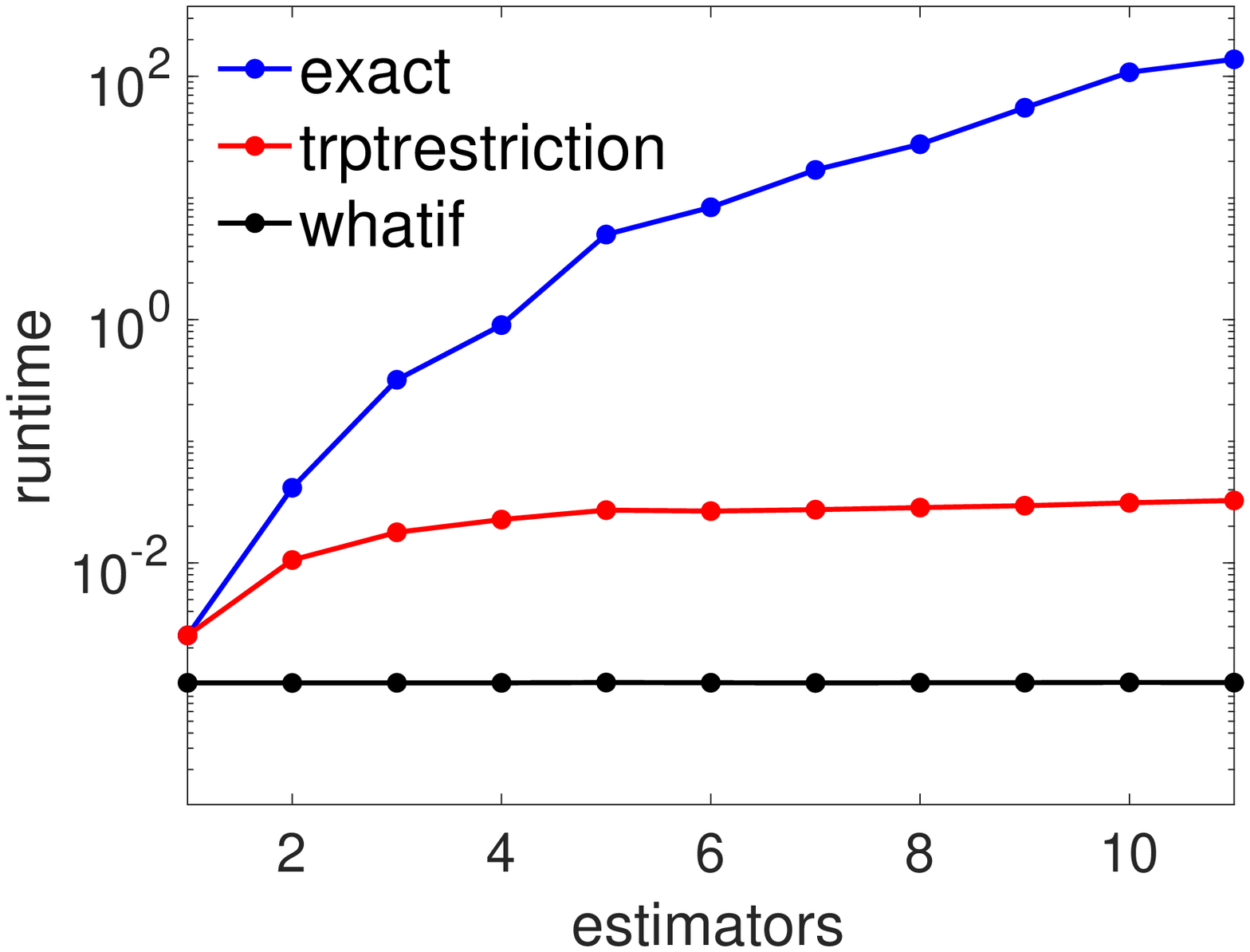}&
    \psfrag{estimators}[c][][1]{\# trees}
    \psfrag{errors}{}
    \psfrag{runtime}{}
    \includegraphics*[width=.2425\linewidth]{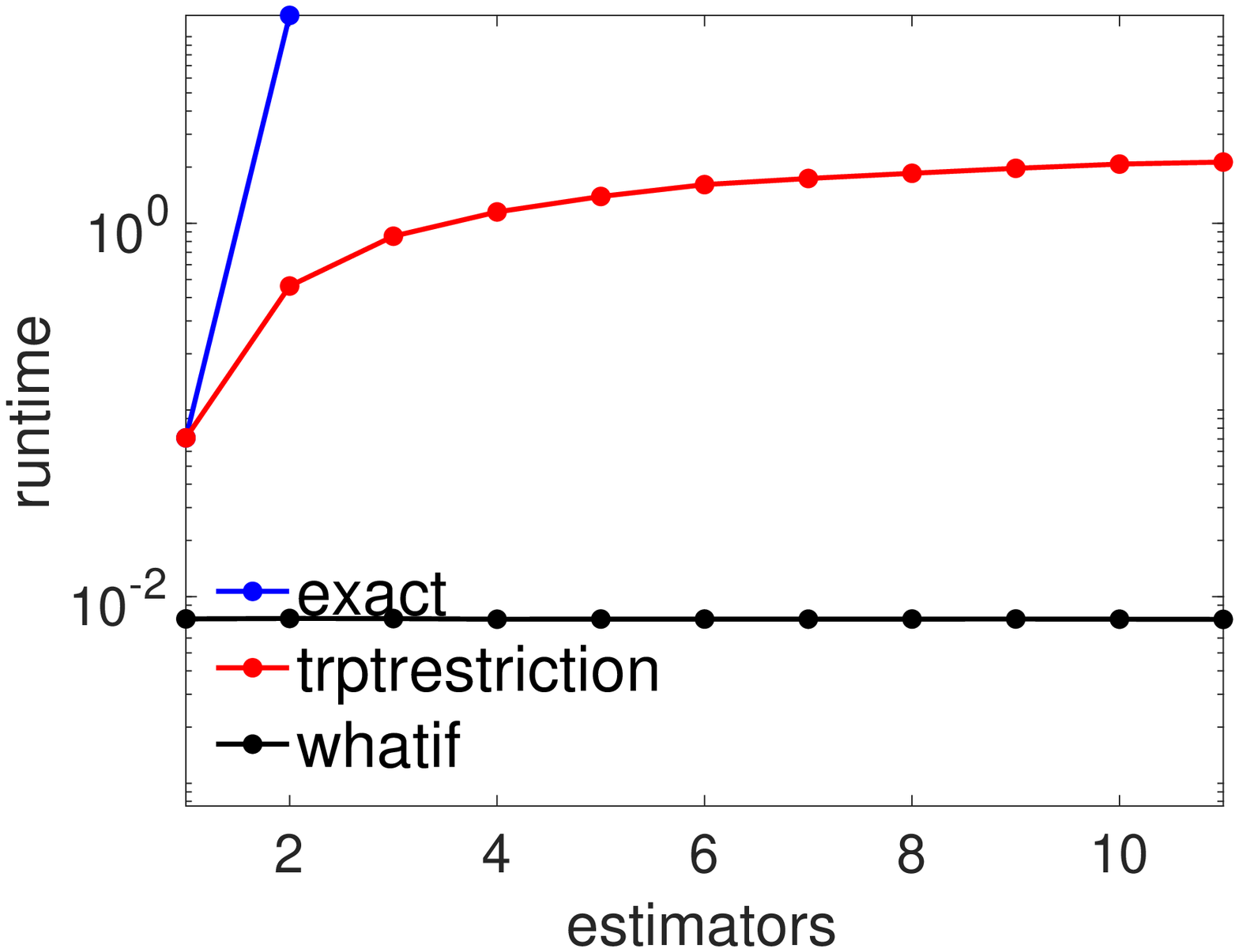}&
    \psfrag{estimators}[c][][1]{\# trees}
    \psfrag{errors}{}
    \psfrag{runtime}{}
    \includegraphics*[width=.2425\linewidth]{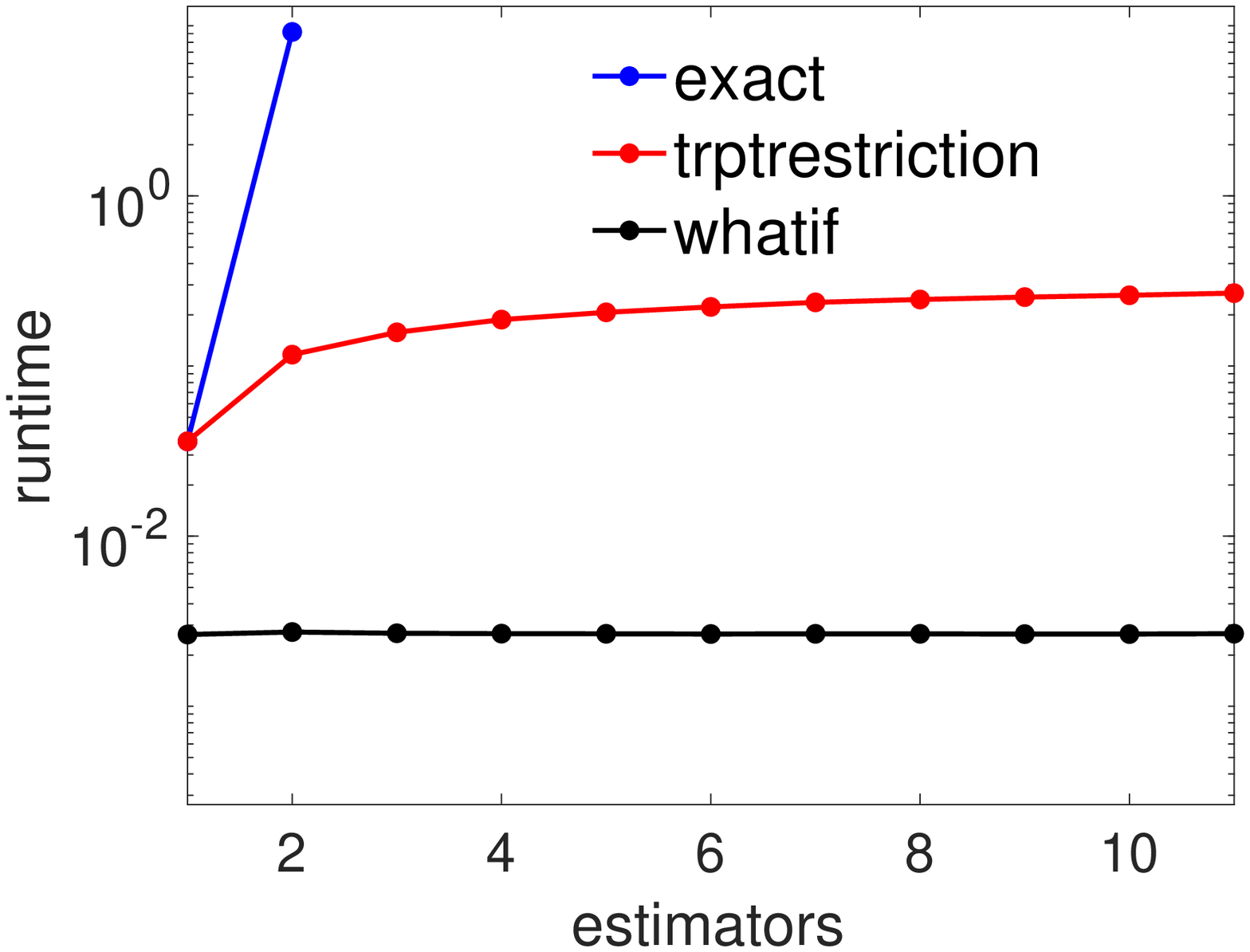}&
    \psfrag{estimators}[c][][1]{\# trees}
    \psfrag{errors}{}
    \psfrag{runtime}{}
    \includegraphics*[width=.2425\linewidth]{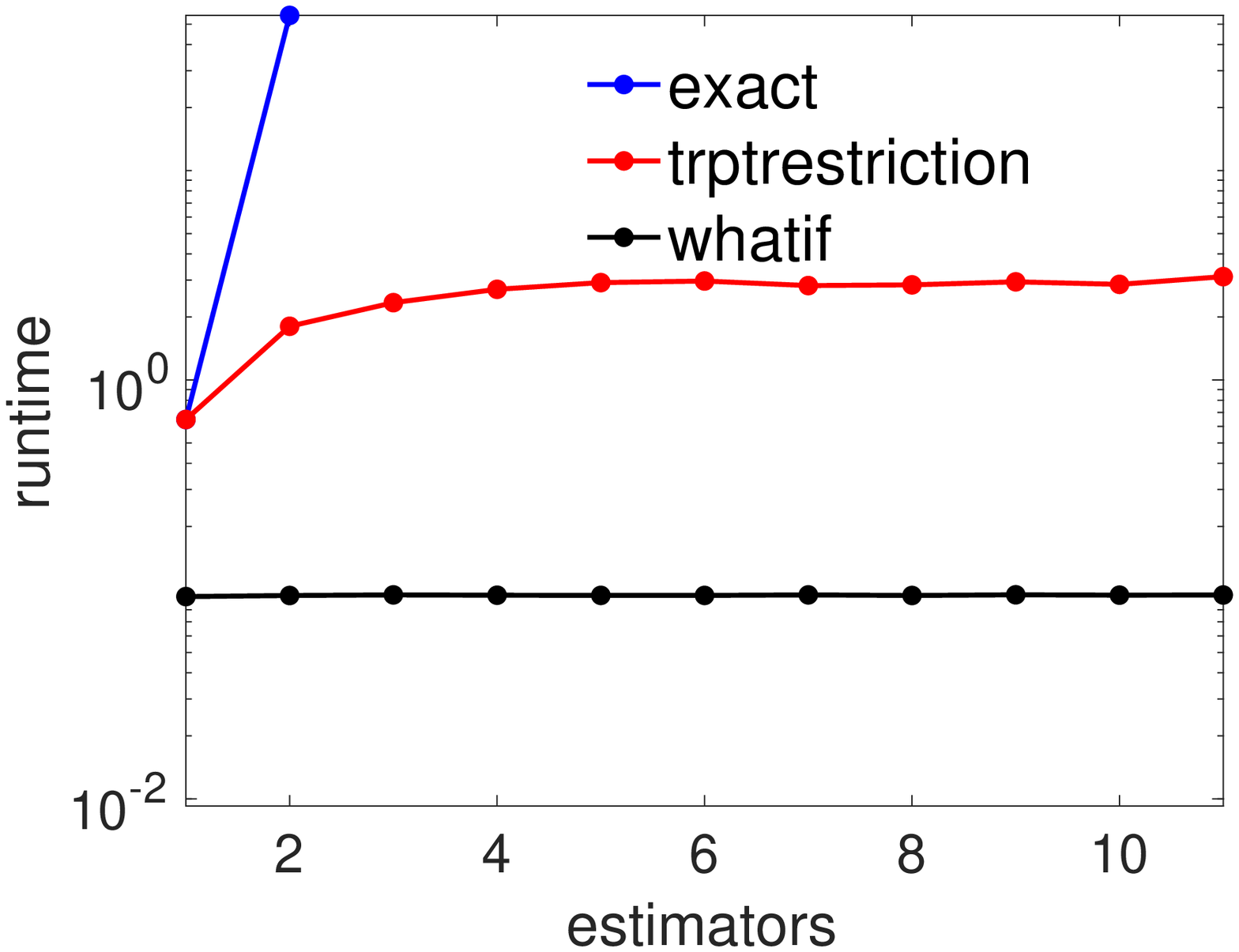} \\[2ex]
    {Breast cancer} & {Spambase} & {Letter} &{MNIST}\\
    \psfrag{errors}[c][]{\raisebox{3ex}[0pt][0pt]{\makebox[0pt]{\hspace{-8ex}oblique trees}}avg.\ distance}
    \psfrag{estimators}{}
    \includegraphics*[width=.2425\linewidth]{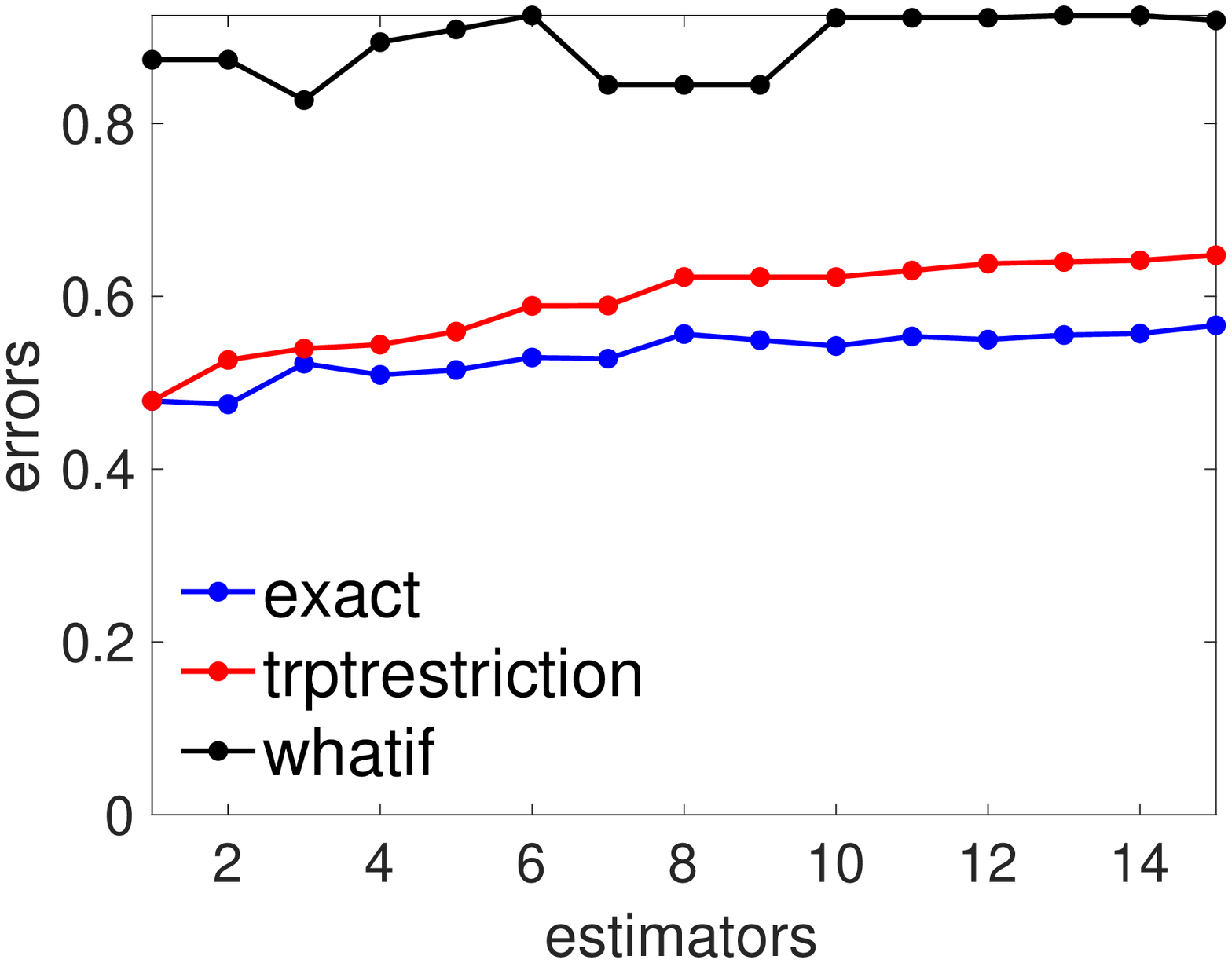}&
    \psfrag{estimators}{}
    \psfrag{errors}{}
    \psfrag{runtime}{}
    \includegraphics*[width=.2425\linewidth,bb=13 4 512 393]{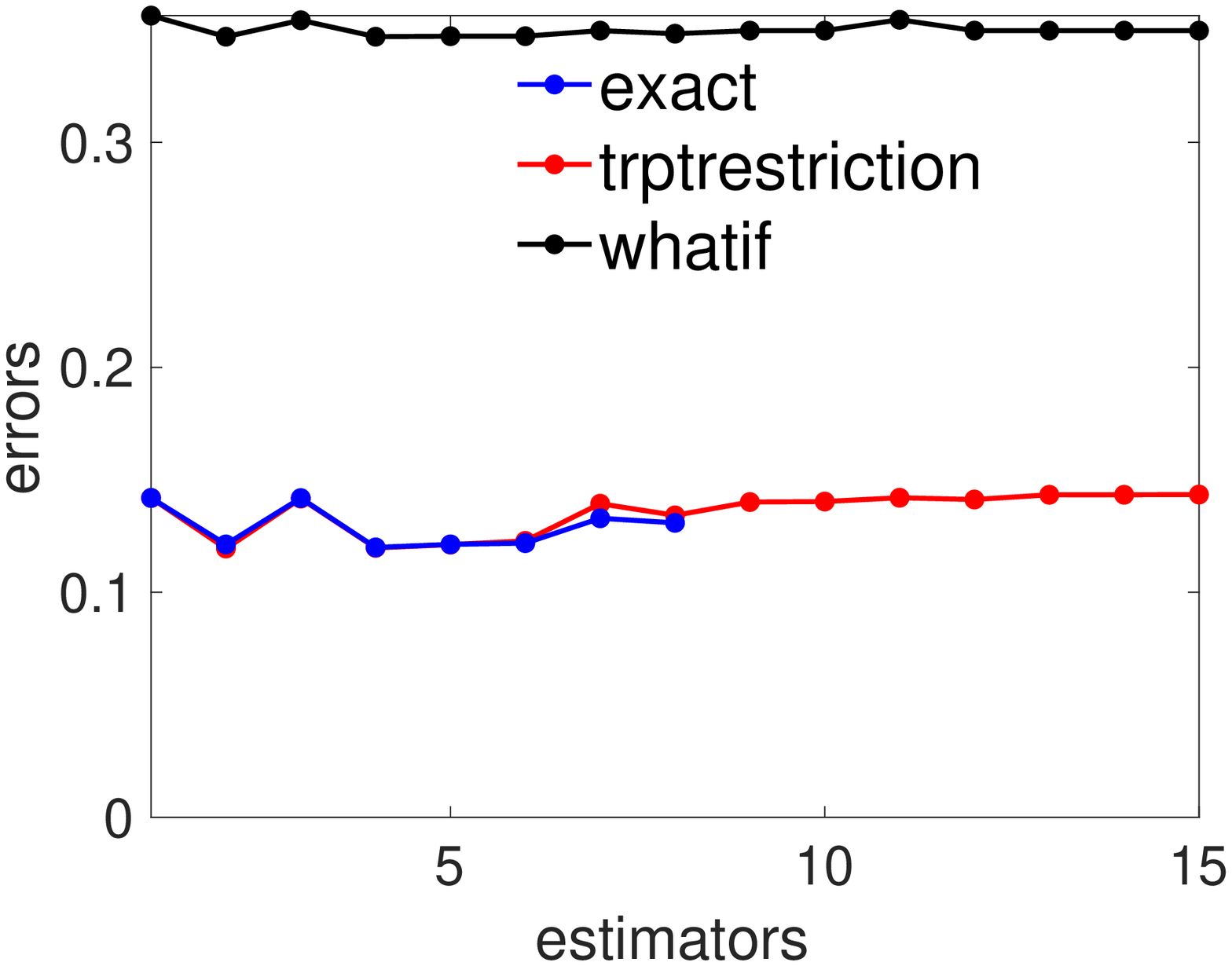}&
    \psfrag{estimators}{}
    \psfrag{errors}{}
    \psfrag{runtime}{}
    \includegraphics*[width=.2425\linewidth]{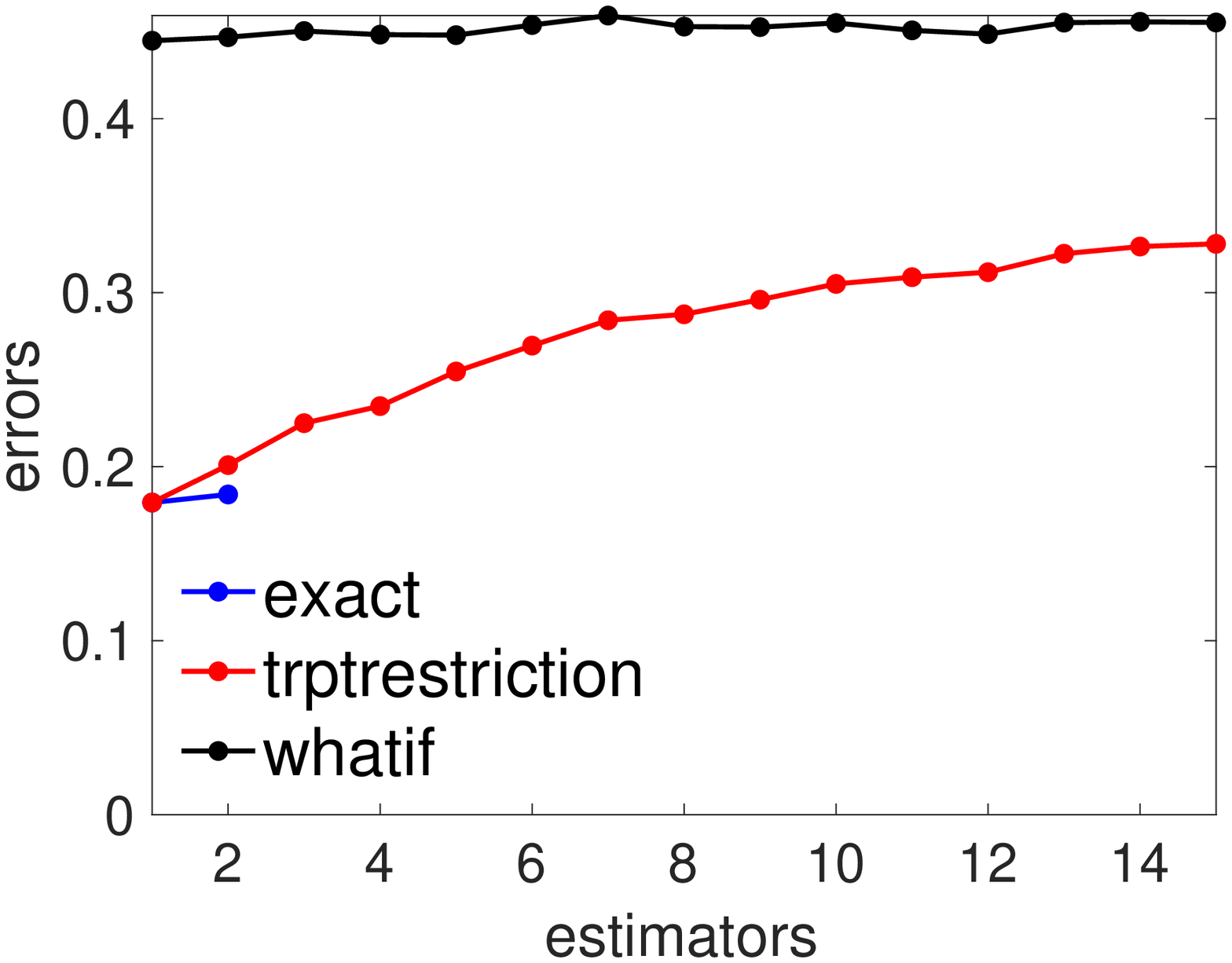}&
    \psfrag{estimators}{}
    \psfrag{errors}{}
    \psfrag{runtime}{}
    \includegraphics*[width=.2425\linewidth,bb=13 4 512 393]{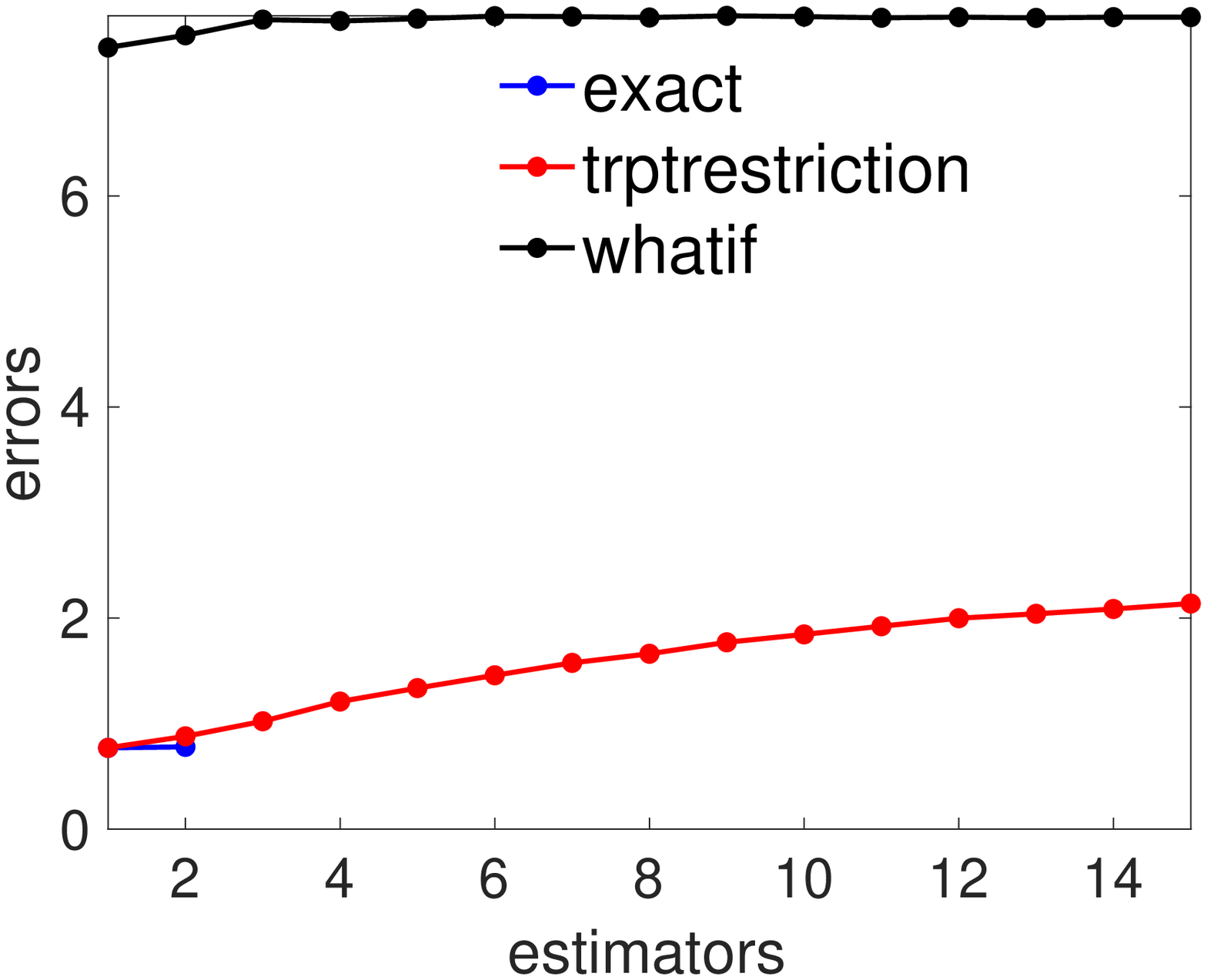}\\[-2ex]
    \psfrag{runtime}[c][]{avg.\ runtime (s)}
    \psfrag{estimators}[c][][1]{\# trees}
    \includegraphics*[width=.2425\linewidth]{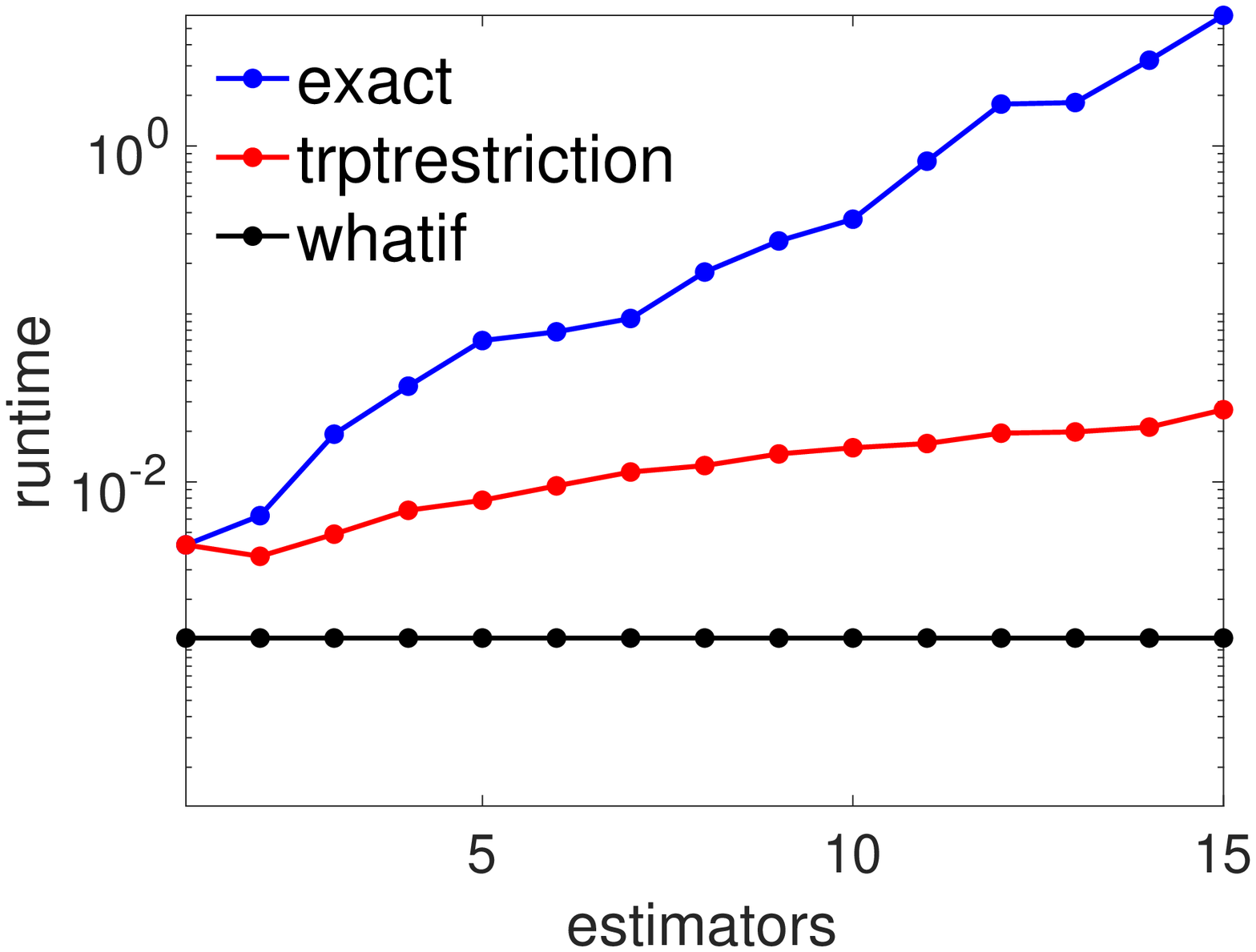}&
    \psfrag{estimators}[c][][1]{\# trees}
    \psfrag{errors}{}
    \psfrag{runtime}{}
    \includegraphics*[width=.2425\linewidth]{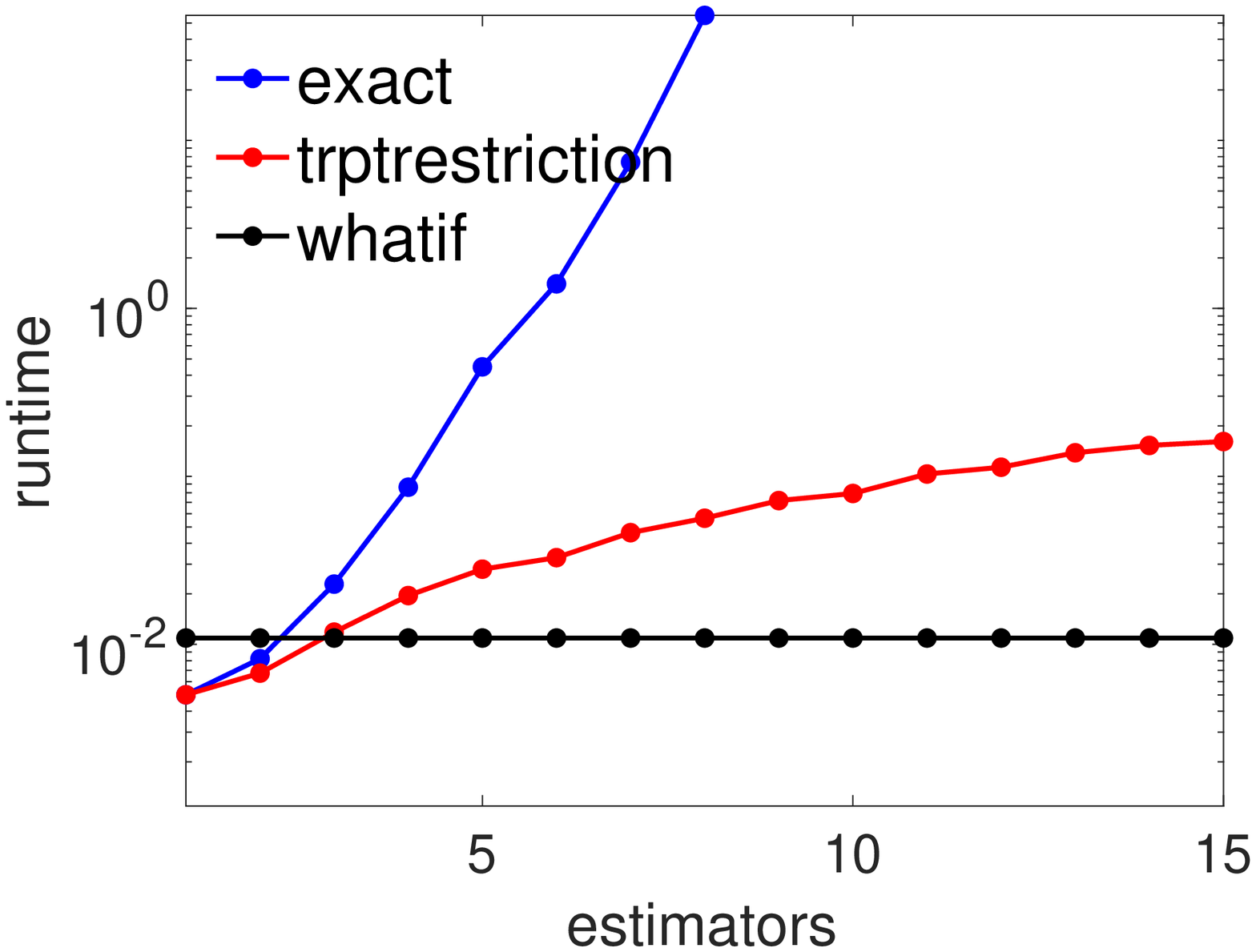}&
    \psfrag{estimators}[c][][1]{\# trees}
    \psfrag{errors}{}
    \psfrag{runtime}{}
    \includegraphics*[width=.2425\linewidth]{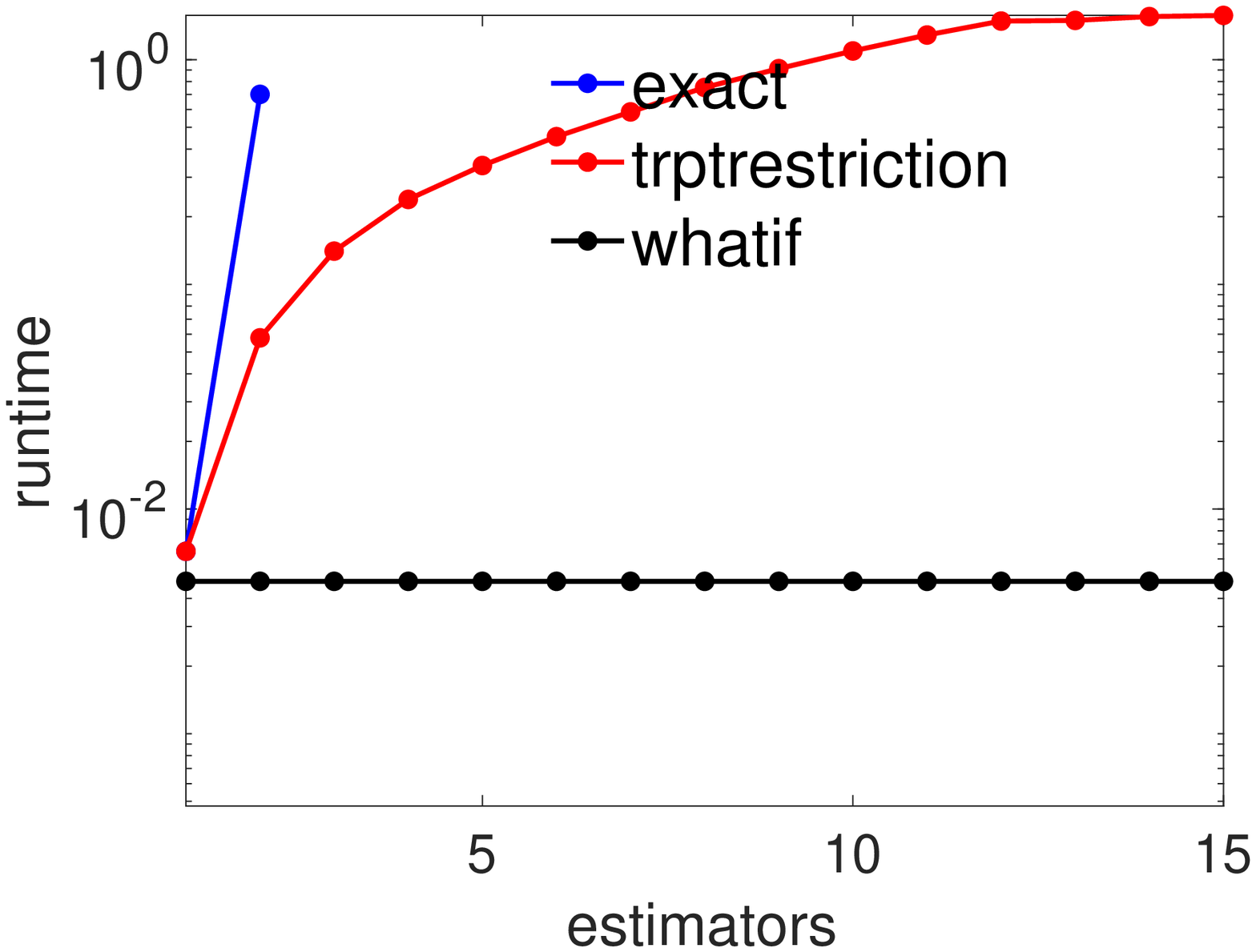}&
    \psfrag{estimators}[c][][1]{\# trees}
    \psfrag{errors}{}
    \psfrag{runtime}{}
    \includegraphics*[width=.2425\linewidth]{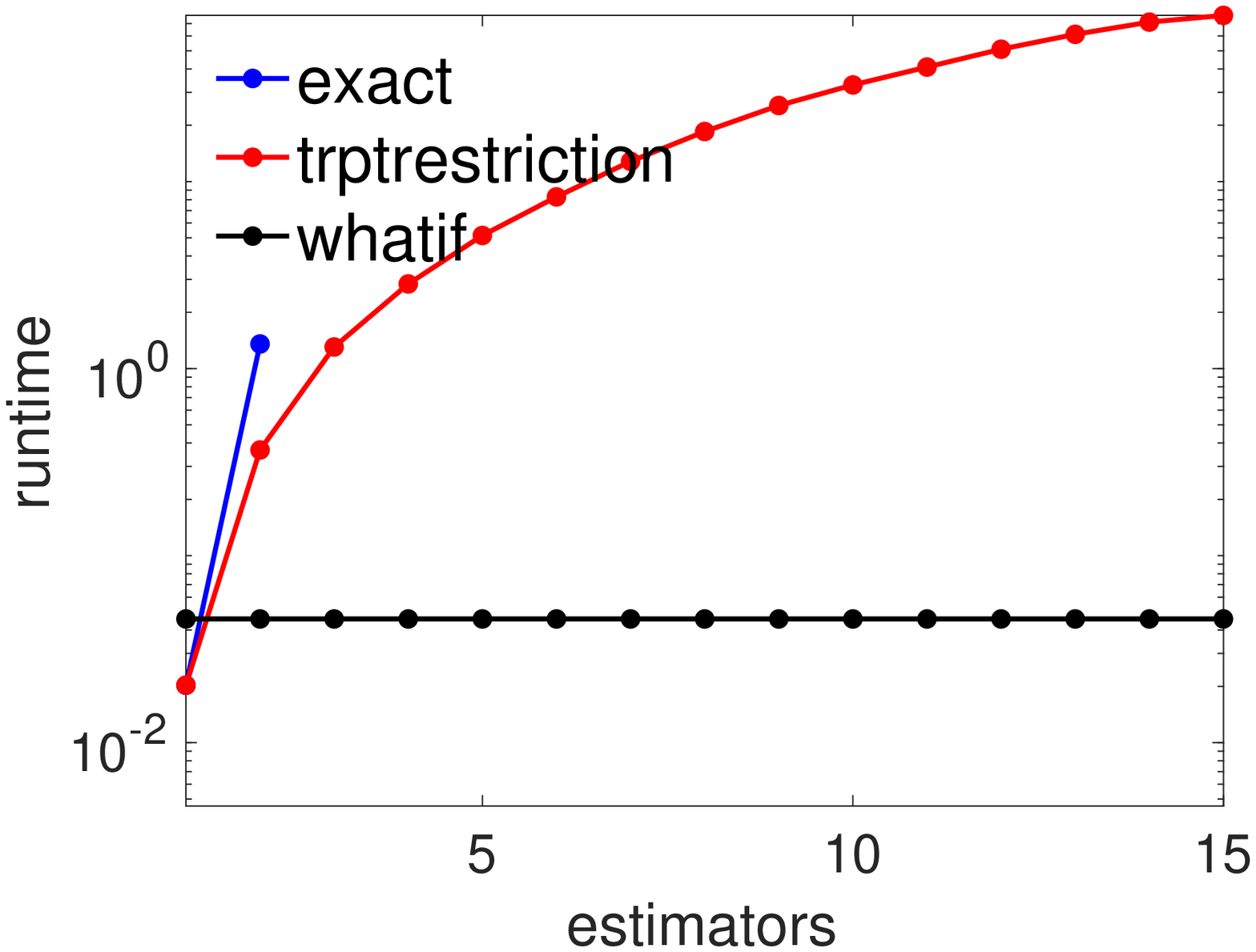}
  \end{tabular}
  \caption{Performance of three types of search: on all nonempty regions, on the live regions (LIRE) and on the dataset points, as a function of the number of trees $T$, for different datasets, for axis-aligned trees (top 2 rows) and oblique trees (bottom 2 rows). We show the $\ell_2$ distance of the CE found ($\norm{\x^*-\overline{\x}}_2$) and the runtime (seconds) to solve the CE problem. The curves are the average for 5 source instances.}
  \label{f:exact-vs-LIRE}
\end{figure}

In order to estimate how good an approximation LIRE is to the exact solution, we do an exhaustive search on all the nonempty regions in small problems for which the latter is computationally feasible. Fig.~\ref{f:exact-vs-LIRE} shows that the approximation (in terms of the distance to the source instance) is quite good, though it degrades as the number of trees increases---since the number of nonempty regions continues to increase exponentially while the number of live regions is capped at $N$. The approximation is quite better for oblique forests than for axis-aligned ones, in agreement with the fact that the number of regions grows more slowly for oblique forests. Importantly, note than LIRE is far better than searching directly on the dataset instances. This, and its very fast runtime, makes LIRE highly practical in order to get a fast, relatively accurate estimate of the optimal CE.

\subsection{LIRE vs other algorithms}

Table~\ref{t:comparison-axis-aligned} compares LIRE with searching on the dataset instances, Feature Tweak \cite{Tolomei_17a} and OCEAN \cite{ParmenVidal21a}. We use several classification datasets of different size, dimensionality and type, and axis-aligned forests (Random Forests) of different size. Unlike previous works on forest CEs, we consider quite larger, high-dimensional datasets and forests---having up to 1\,000 trees, with thousands of leaves per tree. This is important because, to achieve competitive performance in practice, the number of trees may need to be quite large. For small problems, OCEAN (which does an exhaustive search) finds the best solution, but its runtime quickly shoots up and becomes intractable for most cases (see detailed comments in the appendix). LIRE is extremely fast even for large problems, comparable to searching on the dataset instances, but finding better CEs. Also, it is guaranteed to find a feasible solution, i.e., producing the desired prediction.

Table~\ref{t:comparison-oblique}, for oblique trees on some classification datasets, compares LIRE only with the dataset search, since no other algorithm is applicable. Again, LIRE finds better CEs and is reasonably fast, although for large problems its runtime grows from seconds to minutes.

Table~\ref{t:comparison-RFReg} shows results with axis-aligned forests (Random Forests) in regression datasets. From each dataset we randomly select 5 source instances and for each source instance ($\overline{\x}$) we solve two CE problems by setting two different intervals \calS\ in eq.~\eqref{e:CE} as follows: $\calS\ = [a_1, b_1] \subset \bbR$, where $F(\overline{\x}) > b_1$; and $\calS\ = [a_2, b_2] \subset \bbR$, where $F(\overline{\x}) < a_2$. This way, for each dataset, we solve 10 CE problems. The conclusions are as for the classification datasets.

\begin{table}[p]
  \caption{Comparison of different CE algorithms: LIRE, search on dataset points, Feature Tweak \cite{Tolomei_17a} and OCEAN \cite{ParmenVidal21a}, for different datasets and Random Forests, for axis-aligned trees, and optimizing the $\ell_2$ (above) and $\ell_1$ (below) distance. We show the resulting distance $\norm{\x^* - \overline{\x}}_2$ or $\norm{\x^* - \overline{\x}}_1$ and the runtime in seconds (average $\pm$ stdev over 10 source instances). All distances are normalized so that LIRE has unit distance. For LIRE we give the number of live regions and for Feature Tweak the percentage of times the CE found is feasible i.e., it is predicted to be the desired class (all other algorithms are always feasible). For each dataset we give its size, dimensionality and number of classes $(N,D,K)$; for each forest we give its number of trees and average tree depth and number of leaves $(T,\Delta,L)$. ``timeout'' means runtime over 500 s. The best (smallest) distance is in \textbf{boldface}.}
  \label{t:comparison-axis-aligned}
  \centering 
  \footnotesize
  \vspace*{1ex}
  \begin{tabular}{@{}c@{\hspace{1ex}}|@{}c@{\hspace{1ex}}c@{\hspace{1ex}}r@{$\pm$}l@{\hspace{1ex}}|@{}c@{\hspace{1ex}}r@{$\pm$}l@{\hspace{1ex}}|@{}r@{$\pm$}l@{\hspace{1ex}}r@{$\pm$}l@{\hspace{1ex}}c@{\hspace{1ex}}|@{}r@{$\pm$}l@{\hspace{1ex}}r@{$\pm$}l@{}}

    Dataset & \multicolumn{4}{c|}{LIRE} & \multicolumn{3}{c|}{dataset search} & \multicolumn{5}{c|}{Feature Tweak} & \multicolumn{4}{c}{OCEAN} \\
    $(N,D,K)$ & ~regions & time (s) & \multicolumn{2}{c|}{$\ell_2$} & ~time (s) & \multicolumn{2}{c|}{$\ell_2$} & \multicolumn{2}{c}{time (s)} & \multicolumn{2}{c}{$\ell_2$} & feasible & \multicolumn{2}{c}{time (s)} & \multicolumn{2}{c}{$\ell_2$} \\
    $(T,\Delta,L)$ & \multicolumn{2}{c}{} & \multicolumn{2}{c|}{$\ell_1$} & & \multicolumn{2}{c|}{$\ell_1$} & \multicolumn{2}{c}{} & \multicolumn{2}{c}{$\ell_1$} & & \multicolumn{2}{c}{} & \multicolumn{2}{c}{$\ell_1$} \\
    \toprule
    & & & \multicolumn{2}{c|}{} & & \multicolumn{2}{c|}{} & \multicolumn{2}{c}{} & \multicolumn{2}{c}{} & \\
    & 337 &  2$\times 10^{-4}$ & 1.00&0.95 & 1$\times 10^{-4}$ & 1.17&1.08 & 8.6&1.4 & 1.25&0.82 & 100\% & 2.1&1.0 &  \textbf{0.48}& \textbf{0.32} \\ 
    \raisebox{0ex}[0pt][0pt]{\caja[1.1]{b}{c}{breast cancer\\ (559,9,2) \\ (100,9.1,51.9)}}& 337 &  2$\times 10^{-4}$ & 1.00&0.51 & 1$\times 10^{-4}$ & 1.25&0.56 & 8.4&1.6 & 1.04&0.16 & 100\% & 4.6&2.1 &  \textbf{0.45}& \textbf{0.15} \\ [3.3ex]

    & 432& 6$\times 10^{-4}$ & 1.00&0.68 &2$\times 10^{-4}$& 1.23&0.54 & 6.6&1.1 & 1.04&0.98& 100\% &3.3&2.9 & \textbf{0.47}& \textbf{0.45}\\
    \raisebox{0ex}[0pt][0pt]{\caja[1.1]{b}{c}{climate \\ (432,18,2) \\ (100,8.7,48.7)}}& 432& 6$\times 10^{-4}$& 1.00&0.65 &2$\times 10^{-4}$& 2.67&0.29 & 7.2&1.8 &1.14&0.95& 100\% &4.1&3.7 & \textbf{0.38}& \textbf{0.28}\\[3.3ex]

    & 3211& 6$\times 10^{-3}$& 1.00&0.90 & 2$\times 10^{-4}$ & 1.09&0.90 & 89.6&13.2 & 0.54&0.36 & 100\% &75.5&82.7 & \textbf{0.09}& \textbf{0.04}\\
    \raisebox{0ex}[0pt][0pt]{\caja[1.1]{b}{c}{spambase \\ (3.6k,57,2) \\ (100,31.8,596.4)}}& 3211& 6$\times 10^{-3}$& 1.00&0.63 & 2$\times 10^{-4}$ & 1.15&0.65 & 91.2&14.6& \textbf{0.37}& \textbf{0.29} & 100\% &\multicolumn{4}{c}{timeout}\\[3.3ex]

     & 1162 & 3$\times 10^{-4}$ &  \textbf{1.00}& \textbf{0.78}& 2$\times 10^{-4}$ & 1.27&1.15 & 15.9&8.5 & 1.12&1.03 & 67\% &\multicolumn{4}{c}{timeout}\\
    \raisebox{0ex}[0pt][0pt]{\caja[1.1]{b}{c}{yeast \\ (1162,8,10) \\ (100,23.6,734.2) }}& 1162 & 3$\times 10^{-4}$ &  \textbf{1.00}& \textbf{0.76}& 2$\times 10^{-4}$ & 1.33&0.96 & 14.8&8.1 & 1.15&0.57 & 67\% &\multicolumn{4}{c}{timeout}\\[3.3ex]

    & 14532 &  9$\times 10^{-4}$ &  \textbf{1.00}& \textbf{0.26} & 6$\times 10^{-5}$& 1.32&0.97 & 58.2&18.5 & 1.09&0.82 & 100\% &\multicolumn{4}{c}{timeout} \\
    \raisebox{0ex}[0pt][0pt]{\caja[1.1]{b}{c}{letter \\ (16k,16,26)\\ (100,27.9,4201)} }& 14532 &  9$\times 10^{-4}$ &  \textbf{1.00}& \textbf{0.61} & 6$\times 10^{-5}$ & 1.96&0.68 & 61.1&18.9 & 1.42&0.37 & 100\% &\multicolumn{4}{c}{timeout} \\[3.3ex] 

    & 55000 & 2$\times 10^{-1}$&  \textbf{1.00}& \textbf{0.73} & 4$\times 10^{-2}$& 1.41&0.89& 147.8&48.5 & \multicolumn{2}{c}{--} & 0\%&\multicolumn{4}{c}{timeout}\\
    \raisebox{0ex}[0pt][0pt]{\caja[1.1]{b}{c}{MNIST \\ (55k,784,10) \\ (100,34.1,9369) } }& 55000 & 2$\times 10^{-1}$ &  \textbf{1.00}& \textbf{0.44}& 4$\times 10^{-2}$ & 1.88&0.58& 151.8&51.4 & \multicolumn{2}{c}{--}&0\%&\multicolumn{4}{c}{timeout}\\[3.3ex]

    & 103692 & 2$\times 10^{-1}$&  \textbf{1.00}& \textbf{0.80}& 1$\times 10^{-2}$ & 1.17&0.87 & \multicolumn{5}{c|}{timeout} &\multicolumn{4}{c}{timeout}\\
    \raisebox{0ex}[0pt][0pt]{ \caja[1.1]{b}{c}{MiniBooNE \\ (104051,50,2) \\ (100,34.5,9616) } }& 103692 & 2$\times 10^{-1}$ &  \textbf{1.00}& \textbf{0.32} & 1$\times 10^{-2}$ & 1.42&0.38 & \multicolumn{5}{c|}{timeout} &\multicolumn{4}{c}{timeout}\\[3.3ex]

    & 8607 & 8$\times 10^{-1}$ & \textbf{1.00}& \textbf{0.66}&  4$\times 10^{-1}$ & 11.21&4.44& \multicolumn{5}{c|}{timeout} &\multicolumn{4}{c}{timeout}\\
    \raisebox{0ex}[0pt][0pt]{\caja[1.1]{b}{c}{Swarm \\ (18647,2400,2) \\ (100,31.6,1464) } }& 8607 & 8$\times 10^{-1}$ & \textbf{1.00}& \textbf{0.41}&  4$\times 10^{-1}$ & 27.00&6.31 & \multicolumn{5}{c|}{timeout} &\multicolumn{4}{c}{timeout}\\[3.3ex]

    & 12901 & 1.3 &  \textbf{1.00}& \textbf{0.70}&  4$\times 10^{-1}$ &5.31&2.10& \multicolumn{5}{c|}{timeout} &\multicolumn{4}{c}{timeout}\\
    \raisebox{0ex}[0pt][0pt]{\caja[1.1]{b}{c}{Swarm \\ (18647,2400,2) \\ (1000,31.3,1468) } }& 12901 & 1.3 &  \textbf{1.00}& \textbf{0.53}&  4$\times 10^{-1}$ &18.59&4.34& \multicolumn{5}{c|}{timeout} &\multicolumn{4}{c}{timeout} \\
    \bottomrule
  \end{tabular}
\end{table} 

\begin{table}[p]
  \caption{Like table~\ref{t:comparison-axis-aligned} but using a Random Forest of oblique trees (trained with TAO \cite{CarreirTavall18a,Carreir22a}).}
  \label{t:comparison-oblique}
  \centering 
  \vspace*{1ex}
  \begin{tabular}{@{}ccc|c@{\hspace{1ex}}c@{\hspace{1ex}}r@{$\pm$}l@{\hspace{2ex}}|c@{\hspace{1ex}}r@{$\pm$}l@{}}
    Dataset & $(N,D,K)$ & $(T,\Delta,L)$ & \multicolumn{4}{c|}{LIRE} & \multicolumn{3}{c}{dataset search} \\
    & & & regions & time (s) & \multicolumn{2}{c|}{$\ell_2$} & time (s) & \multicolumn{2}{c}{$\ell_2$} \\
    \toprule  
    breast cancer & (559,9,2) & (30,2.3,5.8) & 60 & 0.07 & \textbf{1.00}&\textbf{0.52} & 0.001 & 1.23&0.95  \\
    spambase & (3.6k,57,2) & (30,2.6,7.8) & 214 & 0.36 &  \textbf{1.00}&\textbf{0.71}& 0.011 & 2.57&2.42  \\
    letter & (16k,16,26) & (30,8.0,289.2) & 13238 & 2.63 & \textbf{1.00}&\textbf{0.70} & 0.004 & 1.21&0.83  \\
    MNIST & (55k,784,10) & (30,8.0,148.6) & 50711 & 151.97 & \textbf{1.00}&\textbf{0.83} & 0.040 & 3.03&1.81
  \end{tabular}
\end{table}

\begin{table}[p]
  \caption{Like table~\ref{t:comparison-axis-aligned} but for regression.}
  \label{t:comparison-RFReg}
  \centering 
  \vspace*{1ex}
  \begin{tabular}{@{}c|c@{\hspace{1ex}}c@{\hspace{1ex}}r@{$\pm$}l@{\hspace{2ex}}|c@{\hspace{1ex}}r@{$\pm$}l@{}}
    Dataset& \multicolumn{4}{c|}{LIRE} & \multicolumn{3}{c}{dataset search} \\
     $(N,D,K)$ & regions & time (s) & \multicolumn{2}{c|}{$\ell_2$} & time (s) & \multicolumn{2}{c}{$\ell_2$} \\
      $(T,\Delta,L)$ & && \multicolumn{2}{c|}{$\ell_1$} & &\multicolumn{2}{c}{$\ell_1$} \\
    \toprule  
      abalone & \multicolumn{4}{c|}{} & \multicolumn{3}{c}{}\\
     (3341,10,1) & 1097 &1$\times 10^{-4}$ & \textbf{1.00}& \textbf{0.91} & 5$\times 10^{-5}$ & 1.31 & 1.14\\   
     (100,24.7,1277) & 1097 &1$\times 10^{-4}$ & \textbf{1.00}& \textbf{0.95} & 5$\times 10^{-5}$ & 1.48 & 1.35 \\[1ex]     
     cpuact & \multicolumn{4}{c|}{} & \multicolumn{3}{c}{}\\
     (6553,21,1)  & 6553 &2$\times 10^{-4}$ & \textbf{1.00}& \textbf{0.84} & 7$\times 10^{-5}$ & 1.29 & 0.91\\
     (100,29.6,4133) & 6553 &2$\times 10^{-4}$ & \textbf{1.00}& \textbf{0.91} & 7$\times 10^{-5}$ & 1.36 & 0.93\\[1ex]
     ailerons & \multicolumn{4}{c|}{} & \multicolumn{3}{c}{}\\
     (7154,40,1))  & 1106 &2$\times 10^{-4}$ & \textbf{1.00}& \textbf{0.46} & 1$\times 10^{-4}$ & 5.04 & 4.56\\  
     (100,7.9,18.1) &1106&2$\times 10^{-4}$ & \textbf{1.00}& \textbf{0.47} & 1$\times 10^{-4}$ & 4.58 & 3.86\\[1ex]
     CT slice & \multicolumn{4}{c|}{} & \multicolumn{3}{c}{}\\
    (42800,384,1)   & 42416 &4$\times 10^{-2}$ & \textbf{1.00}& \textbf{0.21} & 1$\times 10^{-4}$ & 2.01 & 0.61\\  
     (100,25.0,22550)&42416&4$\times 10^{-2}$ & \textbf{1.00}& \textbf{0.33} & 1$\times 10^{-4}$ & 3.42 & 1.68\\[1ex]
     CT slice & \multicolumn{4}{c|}{} & \multicolumn{3}{c}{}\\
    (42800,384,1)   & 42691 &5$\times 10^{-2}$ & \textbf{1.00}& \textbf{0.19} & 1$\times 10^{-4}$ & 1.56 & 0.48\\  
     (1000,25.0,22550)&42691&5$\times 10^{-2}$ & \textbf{1.00}& \textbf{0.32} & 1$\times 10^{-4}$ & 2.24 & 1.06\\    
   \end{tabular}
\end{table}

\subsection{Realistic CEs}

\begin{figure}[p]
  \centering
  \scriptsize
  \newcommand{\mysize}{0.113}
  \begin{tabular}{@{}c@{\hspace{0.04\linewidth}}c@{\hspace{0.02\linewidth}}c@{\hspace{0.02\linewidth}}c@{\hspace{0.04\linewidth}}c@{\hspace{0.02\linewidth}}c@{\hspace{0.02\linewidth}}c@{}}
    & \multicolumn{3}{c}{\normalsize $\ell_2$} & \multicolumn{3}{c}{\normalsize $\ell_1$} \\
    \cline{2-4}\cline{5-7}
    \normalsize $\overline{\x}$ & \normalsize LIRE & \normalsize LIRE $\x_n$ & \normalsize dataset & \normalsize LIRE & \normalsize LIRE $\x_n$ & \normalsize dataset \\
    \toprule
    (\underline{1},1.0,0.0) & \textbf{(\underline{8},0.0,1.0),2.8}&  (\underline 8,0.0,1.0),8.6&  (\underline 8,0.1,0.7),5.3 &  \textbf{(\underline 8,0.0,1.0),18.0} &  (\underline 8,0.0,1.0),94.7 &  (\underline 8,0.1,0.7),43.8 \\
    \includegraphics*[width=\mysize \linewidth]{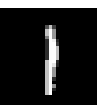}&
    \includegraphics*[width=\mysize \linewidth]{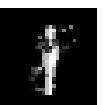}&
    \includegraphics*[width=\mysize \linewidth]{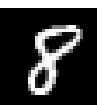}&
    \includegraphics*[width=\mysize \linewidth]{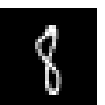} &
    \includegraphics*[width=\mysize \linewidth]{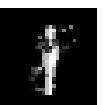}&
    \includegraphics*[width=\mysize \linewidth]{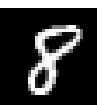}&
    \includegraphics*[width=\mysize \linewidth]{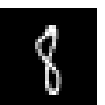} \\

    (\underline{1},1.0,0.0) &  \textbf{(\underline 2,0.1,0.8),3.1} &  (\underline 2,0.1,0.8),4.7 &  (\underline 2,0.1,0.8),4.7
    & \textbf{(\underline 2,0.1,0.8),18.3}&  (\underline 2,0.1,0.8),36.8 &  (\underline 2,0.1,0.8),36.8 \\
    \includegraphics*[width=\mysize \linewidth]{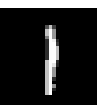}&
    \includegraphics*[width=\mysize \linewidth]{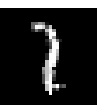}&
    \includegraphics*[width=\mysize \linewidth]{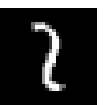}&
    \includegraphics*[width=\mysize \linewidth]{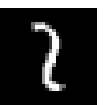} &
    \includegraphics*[width=\mysize \linewidth]{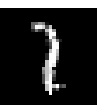}&
    \includegraphics*[width=\mysize \linewidth]{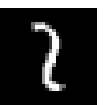}&
    \includegraphics*[width=\mysize \linewidth]{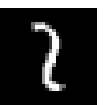} \\

    (\underline{1},1.0,0.0) &  \textbf{(\underline 3,0.0,1.0),3.2} &  (\underline 3,0.0,1.0),9.5 &  (\underline 3,0.2,0.7),4.8 &  \textbf{(\underline 3,0.0,1.0),20.2} &  (\underline 3,0.0,1.0),100.4 &  (\underline 3,0.2,0.7),35.8 \\
    \includegraphics*[width=\mysize \linewidth]{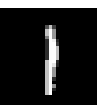}&
    \includegraphics*[width=\mysize \linewidth]{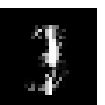}&
    \includegraphics*[width=\mysize \linewidth]{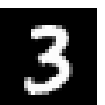}&
    \includegraphics*[width=\mysize \linewidth]{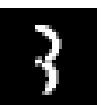} &
    \includegraphics*[width=\mysize \linewidth]{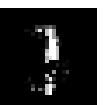}&
    \includegraphics*[width=\mysize \linewidth]{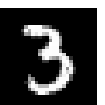}&
    \includegraphics*[width=\mysize \linewidth]{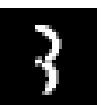} \\

    (\underline{1},1.0,0.00) &  \textbf{(\underline 4,0.1,0.7)},3.4 &  (\underline 4,0.1,0.7),4.1&  (\underline 4,0.1,0.7),4.1 &  \textbf{(\underline 4,0.1,0.7),18.6} &  (\underline 4,0.1,0.7),29.4 &  (\underline 4,0.1,0.7),28.9 \\
    \includegraphics*[width=\mysize \linewidth]{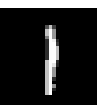}&
    \includegraphics*[width=\mysize \linewidth]{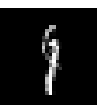}&
    \includegraphics*[width=\mysize \linewidth]{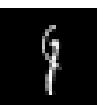}&
    \includegraphics*[width=\mysize \linewidth]{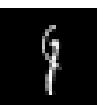} &
    \includegraphics*[width=\mysize \linewidth]{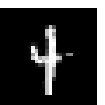}&
    \includegraphics*[width=\mysize \linewidth]{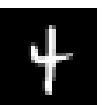}&
    \includegraphics*[width=\mysize \linewidth]{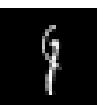} \\

    (\underline{4},0.9,0.0) &  \textbf{(\underline 3,0.0,0.9),5.4} &  (\underline 3,0.0,0.9),9.8 &  (\underline 3,0.1,0.7),7.3 &  \textbf{(\underline 3,0.0,0.9),42.2}&  (\underline 3,0.0,0.9),95.9 &  (\underline 3,0.1,0.7),76.7 \\
    \includegraphics*[width=\mysize \linewidth]{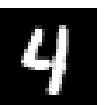}&
    \includegraphics*[width=\mysize \linewidth]{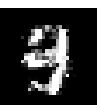}&
    \includegraphics*[width=\mysize \linewidth]{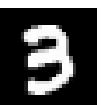}&
    \includegraphics*[width=\mysize \linewidth]{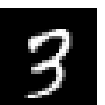} &
    \includegraphics*[width=\mysize \linewidth]{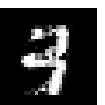}&
    \includegraphics*[width=\mysize \linewidth]{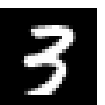}&
    \includegraphics*[width=\mysize \linewidth]{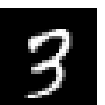} \\

    (\underline{4},0.9,0.00) &  \textbf{(\underline 1,0.0,0.7),7.3} &  (\underline 1,0.0,0.7),9.4 &  (\underline 1,0.1,0.8),8.5 &  \textbf{(\underline 1,0.1,0.8),72.6} &  (\underline 1,0.1,0.8),100.4 &  (\underline 1,0.1,0.8),93.3 \\
    \includegraphics*[width=\mysize \linewidth]{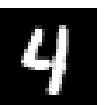}&
    \includegraphics*[width=\mysize \linewidth]{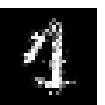}&
    \includegraphics*[width=\mysize \linewidth]{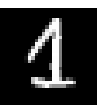}&
    \includegraphics*[width=\mysize \linewidth]{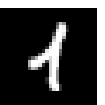} &
    \includegraphics*[width=\mysize \linewidth]{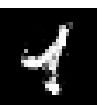}&
    \includegraphics*[width=\mysize \linewidth]{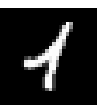}&
    \includegraphics*[width=\mysize \linewidth]{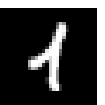} \\

    (\underline{7},0.9,0.0) &  \textbf{(\underline 1,0.1,0.7),6.1} &  (\underline 1,0.1,0.7),9.2 &  (\underline 1,0.1,0.7),7.6 &  \textbf{(\underline 1,0.1,0.7),55.1} &  (\underline 1,0.1,0.7),95.5 &  (\underline 1,0.1,0.7),81.2 \\
    \includegraphics*[width=\mysize \linewidth]{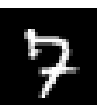}&
    \includegraphics*[width=\mysize \linewidth]{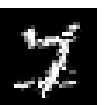}&
    \includegraphics*[width=\mysize \linewidth]{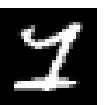}&
    \includegraphics*[width=\mysize \linewidth]{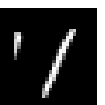} &
    \includegraphics*[width=\mysize \linewidth]{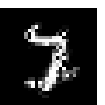}&
    \includegraphics*[width=\mysize \linewidth]{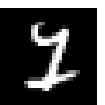}&
    \includegraphics*[width=\mysize \linewidth]{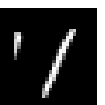} \\
    
    (\underline{2},0.9,0.0) &  \textbf{(\underline 6,0.0,1.0),3.6} &  (\underline 6,0.0,1.0),9.2 &  (\underline 6,0.1,0.7),6.6 &  \textbf{(\underline 6,0.0,1.0),23.4} &  (\underline 6,0.0,1.0),109.2 &  (\underline 6,0.0,0.6),81.2 \\
    \includegraphics*[width=\mysize \linewidth]{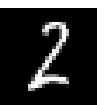}&
    \includegraphics*[width=\mysize \linewidth]{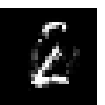}&
    \includegraphics*[width=\mysize \linewidth]{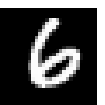}&
    \includegraphics*[width=\mysize \linewidth]{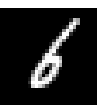} &
    \includegraphics*[width=\mysize \linewidth]{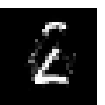}&
    \includegraphics*[width=\mysize \linewidth]{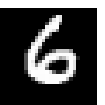}&
    \includegraphics*[width=\mysize \linewidth]{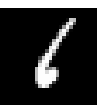} \\
    
    (\underline{4},1.0,0.0) &  \textbf{(\underline 7,0.0,0.7),5.0} &  (\underline 7,0.0,0.7),7.5 &  (\underline 7,0.2,0.6),7.0 &  \textbf{(\underline 7,0.0,0.9),38.9} &  (\underline 7,0.0,0.9),87.2 &  (\underline 7,0.1,0.8),71.3 \\
    \includegraphics*[width=\mysize \linewidth]{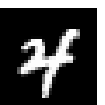}&
    \includegraphics*[width=\mysize \linewidth]{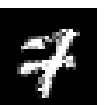}&
    \includegraphics*[width=\mysize \linewidth]{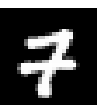}&
    \includegraphics*[width=\mysize \linewidth]{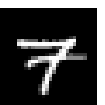} &
    \includegraphics*[width=\mysize \linewidth]{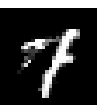}&
    \includegraphics*[width=\mysize \linewidth]{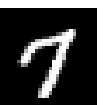}&
    \includegraphics*[width=\mysize \linewidth]{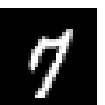} \\
  \end{tabular}
  \vspace*{-2ex}
  \caption{MNIST handwritten digit image CEs. Each row is a different source instance and target class (the class labels are underlined). For each instance, we show the source instance $\overline{\x}$, and the CEs for: LIRE; the dataset instance in the region found by LIRE (``LIRE $\x_n$''); and the search over all dataset instances (for both optimizing the $\ell_2$ distance in columns 2--4, and the $\ell_1$ distance in columns 5--7). The best (closest) CE is in \textbf{boldface}. For each CE, we give numbers like ``(\underline{8},0.0,1.0),2.8'' meaning the class predicted (8), the forest-predicted probability for the source and target class (0.0,1.0) and the distance (2.8).}
  \label{f:MNIST}
\end{figure}

Fig.~\ref{f:MNIST} shows actual source and CE instances for the MNIST dataset of handwritten digit images, optimizing for the $\ell_2$ and $\ell_1$ distances. Besides LIRE and the search in the dataset instances, we also show the instance (or, if there are several, the closest one to the source instance) in the closest live region that LIRE finds. Note that the latter two, being actual data points, are realistic by definition, but they are also necessarily farther than the LIRE CE. We can see that the LIRE CE also looks like quite a realistic image, sometimes altering strokes of the source digit so as to make it look like the target class.

\clearpage

\section{Conclusion}

Decision forests define a piecewise constant function with an exponential number of regions in feature space, so searching for a counterfactual explanation exhaustively is impractical unless the forest is very small (in number of trees and of leaves). However, if we restrict the search to only those regions containing at least an actual data point (``live'' regions), then the search becomes not only practical but very fast, even suitable for interactive use in some cases. This can also be seen as a realistic formulation of counterfactual explanations where the solution is constrained to lie in high-density regions of feature space, and the live regions act as a nonparametric density estimate. We are working on scaling the search to even larger forests and datasets using pruning heuristics and approximate nearest-neighbor search techniques.

\section{Acknowledgments}

Work funded in part by NSF award IIS--2007147.

\appendix

\section{Details about the experiments' setup}

\subsection{Datasets}

All datasets are from the UCI Machine Learning Repository \cite{Lichman13a} unless otherwise indicated.

\subsubsection{Classification datasets}

For classification we use the following nine datasets:
\begin{description}
\item[Breast Cancer] The task is to classify whether the cancer is malignant or benign. There are 699 instances, and each instance has 9 real-valued attributes. Since there is no separate test dataset, we randomly divide the entire data into training (80\%) and test (20\%).
\item[Spambase] This dataset consists of a collection of emails, and the task is to create a spam filter that can tell whether an email is spam or not. There are 4\,601 instances, and each instance has 56 real-valued attributes. Since there is no separate test dataset, we randomly divide the entire data into training (80\%) and test (20\%).
\item[Yeast] This has 1\,453 instances, each with 8 real-valued attributes, in 10 classes. Similar to other datasets, we use 20\% inputs as the test set and the rest as the training set.
\item[Climate]The task is to predict climate model simulation outcomes (fail or succeed) using 18 real-valued attributes. There are 540 instances in total, out of which we use 80\% for training and the rest for testing.
\item[Letter] The objective of this dataset is to classify 26 capital letters in the English alphabet. It has separate 4\,000 test instances along with 16\,000 training instances. Each instance has 16 real-valued attributes. The character images were based on 20 different fonts and each letter within these 20 fonts was randomly distorted to produce a file of 20\,000 unique stimuli. Each stimulus was converted into 16 primitive numerical attributes (statistical moments and edge counts), which were then scaled to fit into a range of integer values from 0 through 15.
\item[MNIST] The dataset (from \cite{Lecun_98a}) consists of grayscale images of handwritten digits and the task is to classify them as 0 to 9. There are 60\,000 training images and 10\,000 test images. Each image is of size $28\times 28$ with gray scales in [0,1].
\item[Miniboon] The dataset is taken from the MiniBooNE experiment and is used to distinguish electron neutrinos (signal) from muon neutrinos (background). There are 104K training and 26\,013 test inputs, each containing 50 real-valued features. 
\item[Swarm] There are 23\,309 instances, each with 2\,400 real features, and 2 classes. Similar to other datasets, we use 20\% inputs as the test set and the rest as the training set.
\item[Adult] It is a dataset with mixed type attributes. The prediction task is to determine whether a person makes over 50K a year. There are 12 attributes, out of which 4 are continuous, and the rest are categorical. In all our experiments we convert each categorical attribute to one-hot encoding attribute. Thus each instance has 102 attributes. There are 30\,162 training instances, and separate 15\,062 test instances. 
\end{description}

\subsubsection{Regression datasets}

Since all datasets except Ailerons do not contain separate test sets, we use 80\% inputs as training and the rest as the test.  
\begin{description}
\item[Abalone] The task is to predict the age of abalone from physical measurements. There are a total 4\,177 inputs, each with 10 real-valued attributes. 
\item[Cpuact] The task (from the DELVE data collection\footnote{\url{http://www.cs.toronto.edu/\textasciitilde delve/data/comp-activ/desc.html}}) is to predict the portion of time that CPUs run in user mode given different system measures. The total number of inputs is 8\,192, and each input contains 21 real-valued attributes. 
\item[Ailerons] Aircraft control action prediction\footnote{\url{https://www.dcc.fc.up.pt/\textasciitilde ltorgo/Regression/DataSets.html}}. The attributes describe the status of the aircraft, and the target is the command given to its ailerons. It has separate 6\,596 test instances along with 7\,154 training instances. Each instance has 40 real-valued attributes. 
\item[CT slice ] The attributes are histogram features (in polar space) of a Computer Tomography (CT) slice. The task is to predict the relative location of the image on the axial axis (in the range [0 180]). There are 53\,500 inputs in total, each with 384 real-valued attributes. 
\end{description}

\subsection{Decision forest models}
\label{s:exp-setup}

All our experiments are implemented in Python and run in a single core (without parallel processing). For classification, we present results on two types of decision forests:
\begin{description}
\item[Random Forest] We consider a Random Forest of axis-aligned or oblique trees, where individual trees were trained as follows.
  \begin{description}
  \item[Axis-aligned trees] We used the \texttt{scikit-learn} Random Forest classifier, where individual trees are trained using CART. In our experiments, we used the default parameters, and each tree is grown in full (i.e., not pruned). The only parameter we change is the number of trees in the Random Forest. 
  \item[Oblique trees] Each tree is trained using the TAO algorithm \cite{CarreirTavall18a,Carreir22a}. We use our own implementation based on \cite{ZharmagCarreir20a}. Each tree in the Random Forest is initialized with a full tree of depth 8 and random parameters in all our experiments. We apply an $\ell_1$ penalty on the weights of the decision nodes (which are then somewhat sparse), with a penalty hyperparameter equal to 1 for all datasets.
  \end{description}
\item[AdaBoost forest] For AdaBoost forests, we again use the \texttt{scikit-learn} AdaBoost classifier, where individual trees are axis-aligned and trained using CART. For all datasets, we use a maximum depth of 8 except for the Breast-cancer dataset, where it is 6. The number of trees is selected based on the experiment, and the rest of the parameters are default. 
\end{description}
For regression, we use forests of axis-aligned decision trees only (not oblique). We use the \texttt{scikit-learn} Random Forest regression model, where individual trees are trained using CART. Similar to the classification case, we used the default parameters, and each tree is grown in full. The only parameter we change is the number of trees in the Random Forest. 

\subsection{Counterfactual explanation algorithms: implementations and comments}

We implement LIRE (both for axis-aligned and oblique) and the dataset search in Python. For other algorithms, we use the implementation available online. For both Optimal Action Extraction (OAE) \citep{Cui_15a} and OCEAN \citep{ParmenVidal21a}, we use the implementation by the author of OCEAN\footnote{\url{https://github.com/vidalt/OCEAN}}. For Feature Tweak \citep{Tolomei_17a}, we use the same implementation as in the OCEAN paper\footnote{\url{https://github.com/upura/featureTweakPy}}. OCEAN uses Gurobi as solver.

In table~\ref{t:comparison-axis-aligned}, we write ``timeout'' for a method in a dataset if, for any of the randomly picked source instances from that dataset, the method fails to generate a counterfactual within 500 seconds (a very long time, which would make practical use very inconvenient). Whenever that happens, we stop the experiments for that dataset with that method. For OAE, this happened in all our experiments, so we did not add it to table~\ref{t:comparison-axis-aligned}.

In table~\ref{t:comparison-RFReg}, we do not show OAE because the OCEAN implementation does not support regression (although \citet{Cui_15a} claim that OAE can be used for regression).

OCEAN can only handle source instances (or solution instances) where the numerical (continuous) feature values are in the [0,1] interval. This is achieved by rescaling the training set range of each feature (the minimum and maximum value of each feature). However, test instances may well have features whose values exceed the range in the training set, so after the rescaling they will not lie in [0,1]. During our experiments, we observed that in such cases, sometimes OCEAN fails to return the counterfactual and terminates the process with a message ``MY MILP IS WRONG''. We did not report such cases in the main paper.

Even though OCEAN relies on an highly optimized, commercial MIO solver (Gurobi), its runtime explodes unless one uses very small forests and feature dimensionality---from a few seconds in \cite{ParmenVidal21a} to usually exceeding the 500 seconds' timeout in our experiments. The experimental results shown by \citet{ParmenVidal21a} used Random Forests of shallow trees and datasets of low feature dimensionality. This gives an incomplete, overly optimistic picture of the runtime one can typically expect in realistic situations. In Random Forests each tree is grown fully (without pruning) \cite{Breiman01a}, otherwise one hurts the accuracy of the forest, which defeats its purpose. Fully-grown trees are quite deep with most datasets (see fig.~\ref{f:forest-accuracy-size}, for example). OCEAN's runtime seems to grow exponentially with the tree depth \cite{ParmenVidal21a}.

While \citet{ParmenVidal21a} claim that OCEAN can handle multiclass forests, their results only showed binary classification forests. In our experience with multiclass forests, the runtime of OCEAN blows up with even very small forests. Even for a low-dimensional dataset like Yeast (only 8 features), the runtime exceeds 500 seconds.

Another important, undesirable behavior in MIO approaches such as OCEAN is their large runtime variability, which makes it hard to predict how long it will take to solve a CE, and often results in runtimes that are considerably nonmonotonic as a function of the problem size. This is a consequence of the fact that MIO solvers essentially perform a brute-force search for an NP-hard problem, so we have to expect worst-case exponential runtimes. This large runtime and the effect of the many heuristics that an MIO solver tries in pruning the branch-and-bound search results in a large runtime variability when solving problems of the same size (e.g.\ different source instances or target classes for the CE). Even further, MIO solvers are known to be extremely variable in runtime even on the same exact problem and the same computer, software and software settings. For example, simply changing the representation of the problem (e.g.\ the ordering of the variables or the constraints) can significantly affect the runtime \citep{LodiTramon13a}. Indeed, in our experiments (for the few cases where it runs within the time limit), we observe OCEAN has a standard deviation comparable to the mean. In contrast, the runtime for our algorithm has a far smaller variation, as expected from its computational complexity analysis.

Finally, we also tried to compare with the code for FOCUS\footnote{\url{https://github.com/a-lucic/focus}} \cite{Lucic_22a}. This is an approximate CE method which replaces the hard splits of the decision trees with soft (sigmoid) functions and uses a gradient-based algorithm to optimize the CE problem. FOCUS requires tuning four hyperparameters to generate any single CE. Despite our best efforts, we were unable to generate CEs for all the 10 instances per dataset we report in table~\ref{t:comparison-axis-aligned}, so we do not include FOCUS in our results.

\section{Additional experimental results}
\label{s:expts-add}

\subsection{Accuracy and size of the forests we used}

\begin{figure}[p]
 \centering
 \psfrag{randomforest}[l][l][0.9]{Axis-aligned (RF)}
  \psfrag{adaboost}[l][l][0.9]{Axis-aligned (Adaboost)}
  \psfrag{taoforestd8}[l][l][0.9]{Oblique (RF, depth 8)}
  \psfrag{taoforestd6}[l][l][0.9]{Oblique (RF, depth 6)}
  \psfrag{taoforestd10}[l][l][0.9]{Oblique (RF, depth 10)}
  \psfrag{taoforestd4}[l][l][0.9]{Oblique (RF, depth 4)}
  \psfrag{testerror}[c][][1]{Test error (\%)}
  \psfrag{estimators}{}
  \begin{tabular}{@{}c@{\hspace{2ex}}c@{\hspace{1ex}}c@{}}
    \raisebox{15ex}[0pt][0pt]{\rotatebox{90}{Breast cancer}} &
    \psfrag{trerror}[cb][][1]{Training error (\%)}
    \includegraphics*[width=.4\linewidth,clip,bb=16 4 530 396]{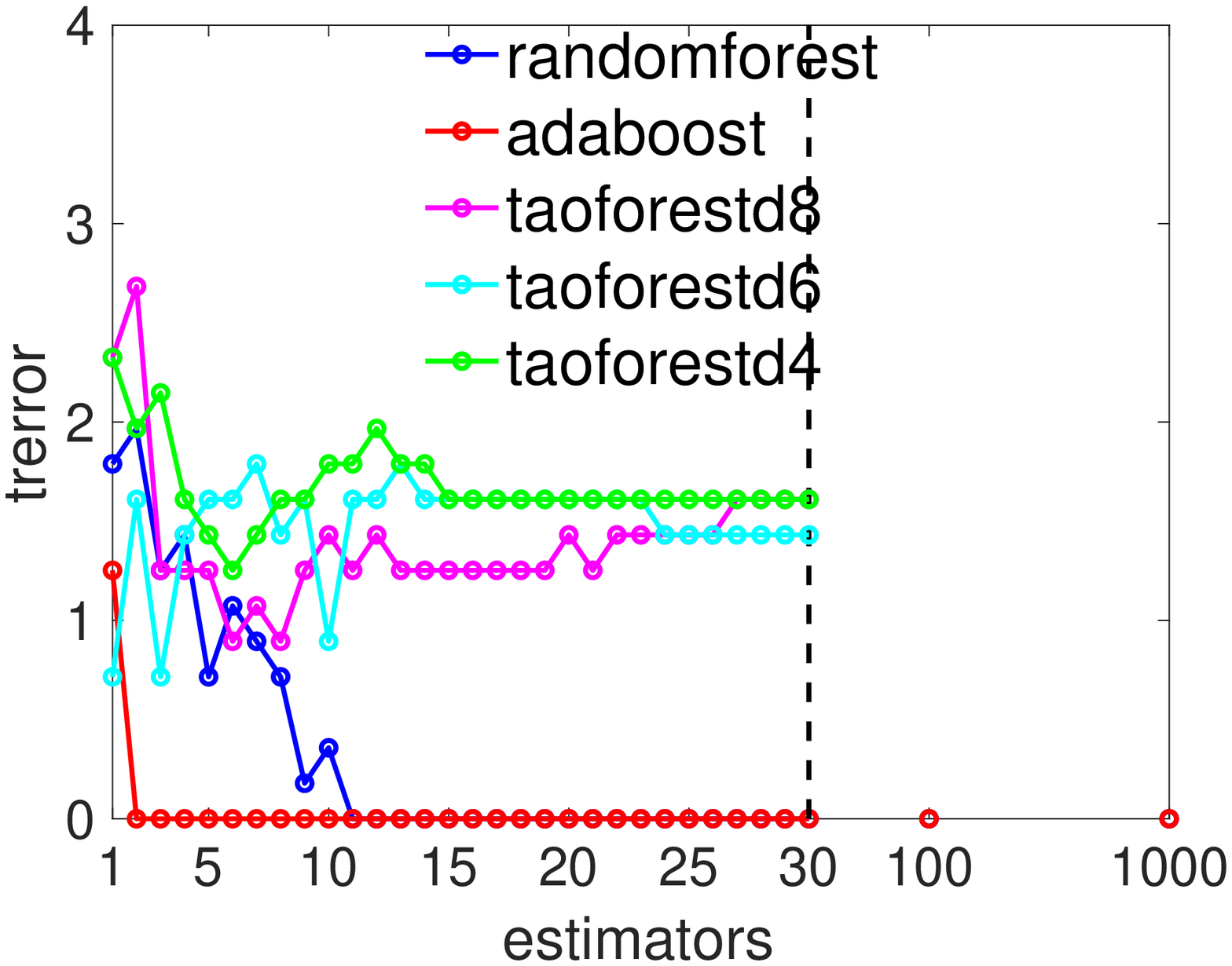}&
    \includegraphics*[width=.4\linewidth]{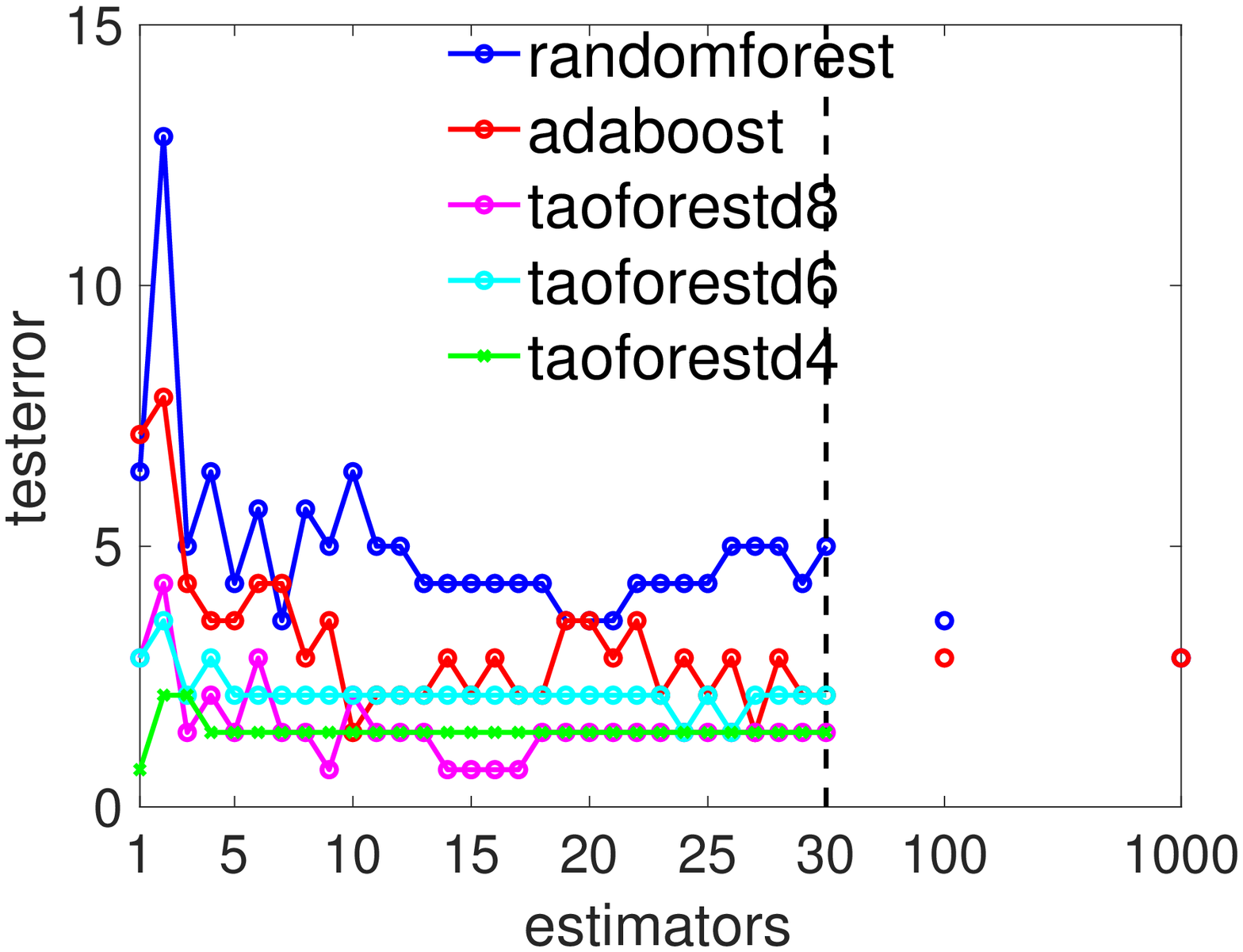}\\
    \raisebox{15ex}[0pt][0pt]{\rotatebox{90}{Spambase}} &
    \psfrag{trerror}[c][][1]{Training error (\%)}
    \includegraphics*[width=.4\linewidth]{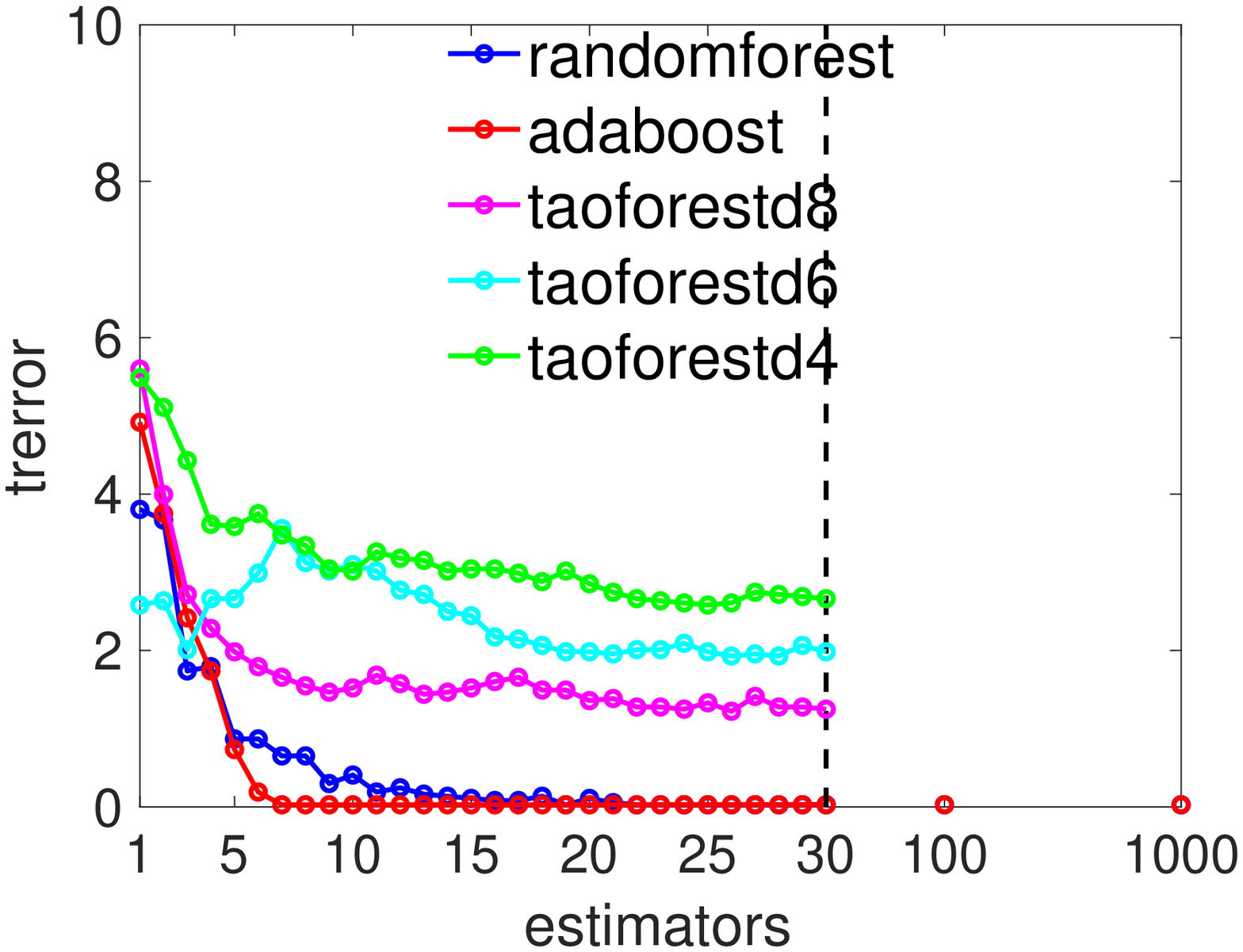}&
    \includegraphics*[width=.4\linewidth]{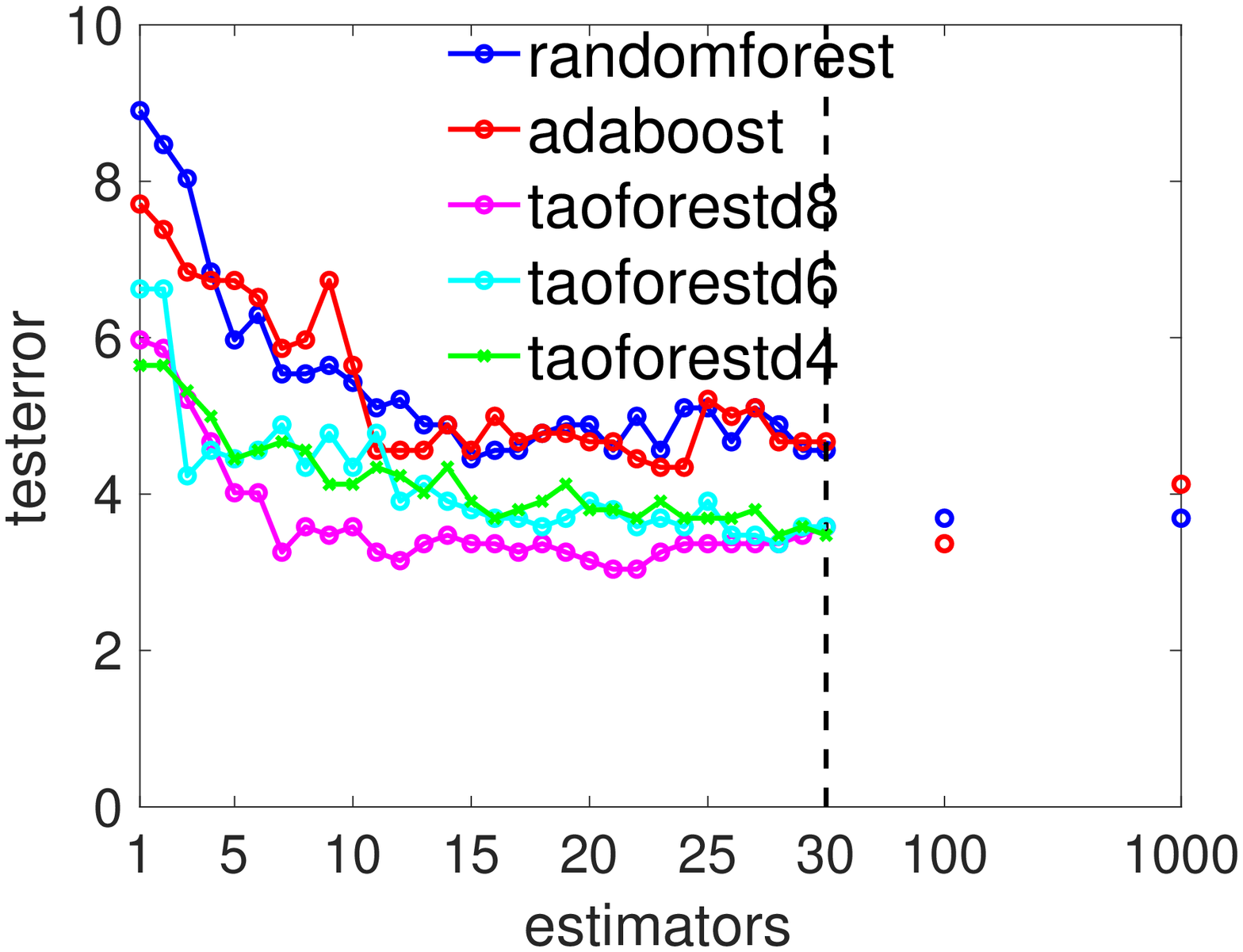}\\
    \raisebox{15ex}[0pt][0pt]{\rotatebox{90}{Letter}} &
    \psfrag{trerror}[c][][1]{Training error (\%)}
    \includegraphics*[width=.4\linewidth]{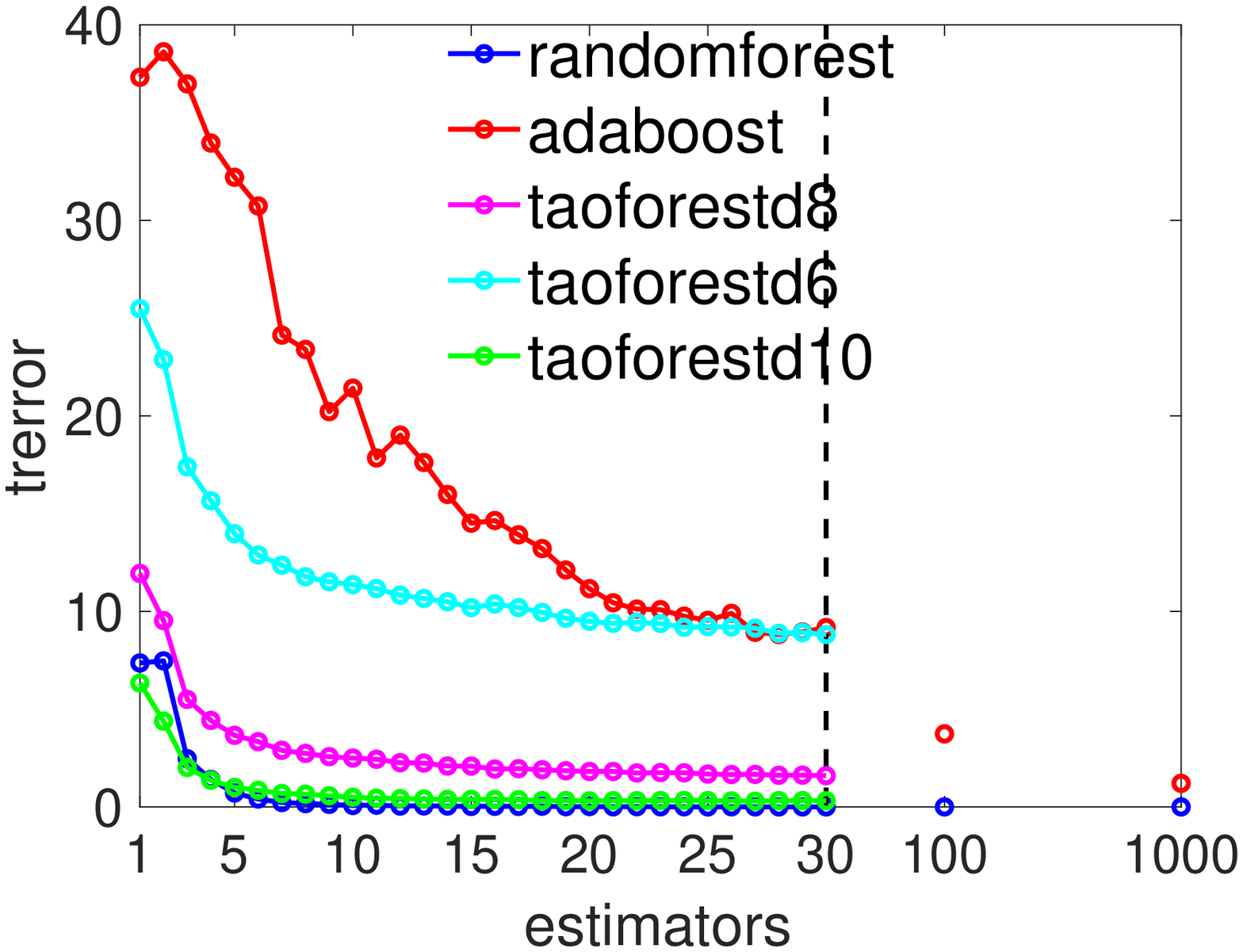}&
    \includegraphics*[width=.4\linewidth]{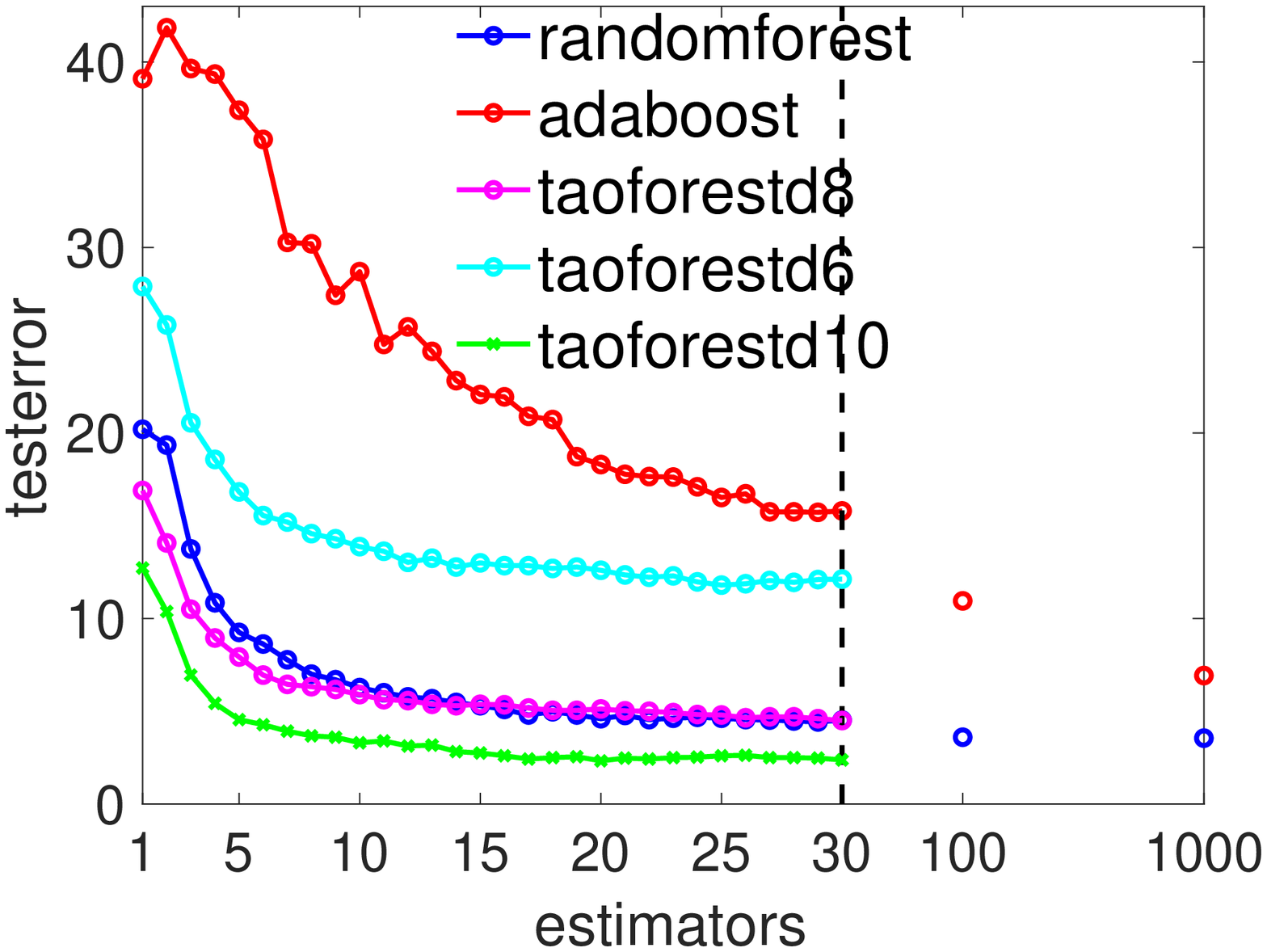}\\
    \raisebox{15ex}[0pt][0pt]{\rotatebox{90}{MNIST}} &
    \psfrag{trerror}[c][][1]{Training error (\%)}
    \psfrag{estimators}[c][][1]{\# trees}
    \includegraphics*[width=.4\linewidth]{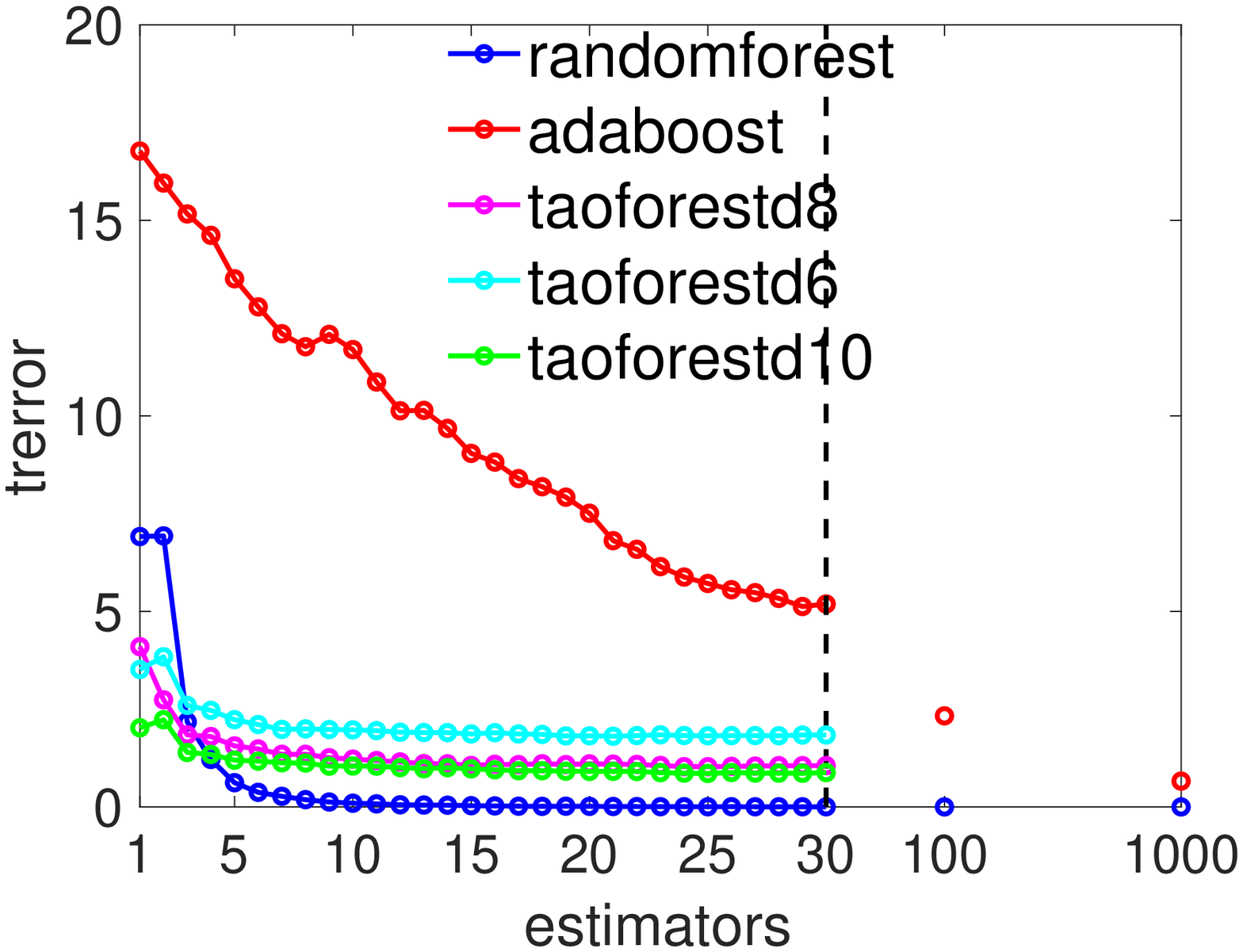}&
    \psfrag{estimators}[c][][1]{\# trees}
    \includegraphics*[width=.4\linewidth]{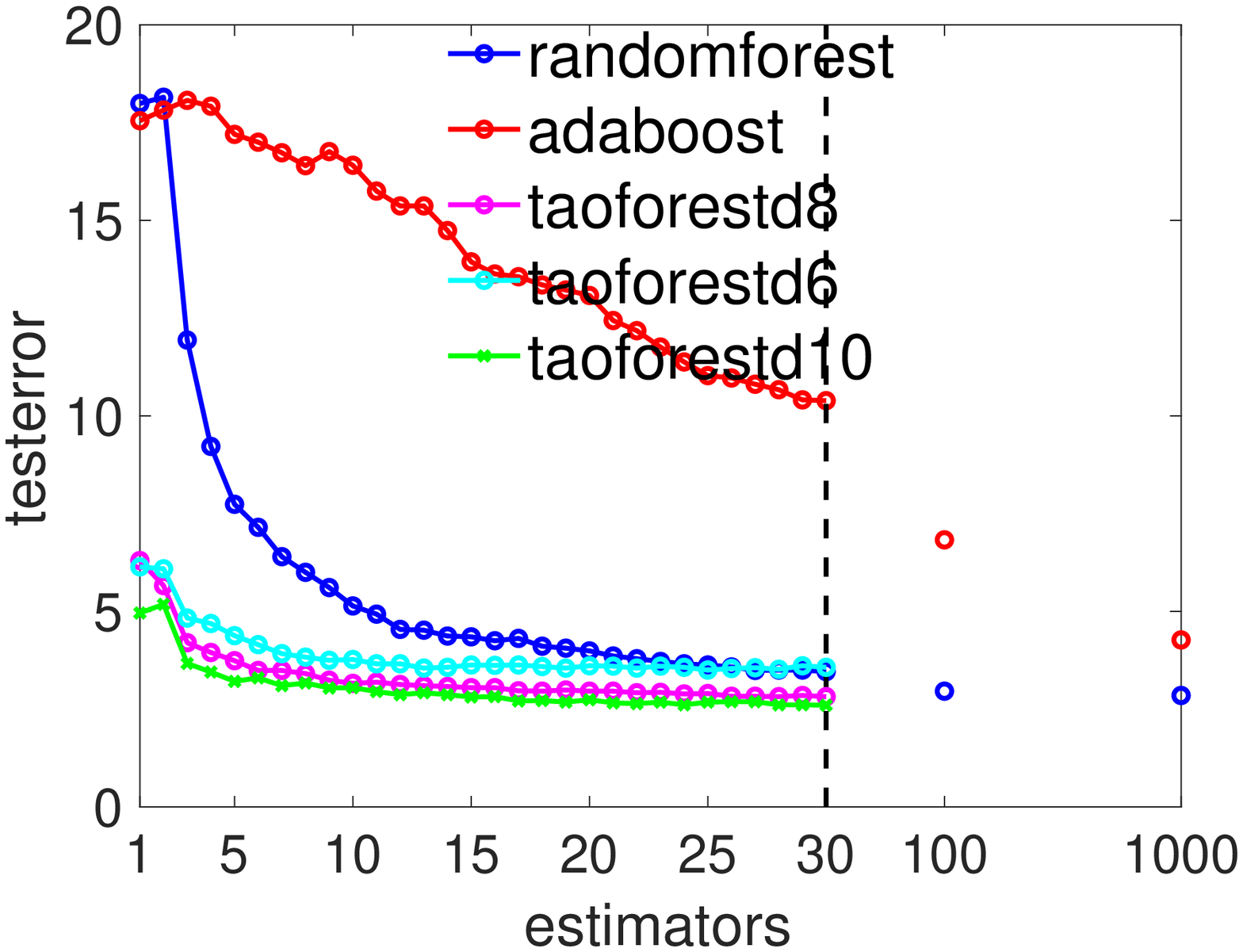}
  \end{tabular}
  \caption{Training and test error of different types of forests (Random Forests, AdaBoost) with different types of trees (axis-aligned, oblique), for different forests sizes (number of trees, depth). Beyond 30 trees, we just show the accuracy of the final forest, to avoid clutter. For axis-aligned Random Forests, the average depth of each tree in the forest for the Breast cancer, Spambase, Letter and MNIST datasets is 9.0, 32.0, 27.9 and 33.8, respectively. For axis-aligned AdaBoost forests, the depth of each tree in the forest is 8, except for Breast cancer, where it is 6.}
  \label{f:forest-accuracy-size}
\end{figure}

Fig.~\ref{f:forest-accuracy-size} shows the training and test error for the forests we used in our experiments. Importantly, observe that for forests with axis-aligned trees (Random Forests, AdaBoost), in order to achieve good accuracy the number of trees may need to reach hundreds or thousands and the depth of each tree may have to be considerable. This is particularly so for Random Forests, which require growing a tree in full, without pruning, for good performance (and this is the default in any practical implementation, such as \texttt{scikit-learn}). Unless the dataset is very small in both dimensionality and sample size, one can expect Random Forests to have trees with depth upwards of 10 (and, consequently, hundreds or thousands of nodes). Previous papers, particularly those using a mixed-integer programming formulation, report experimental results using very small forests (up to depth 5 in \cite{ParmenVidal21a}). Such small forests would not be competitive in terms of accuracy. With larger, more practically useful forests, solving counterfactuals exactly as in such approaches is infeasible, because they essentially do a brute-force search in an exponentially large search space.

For forests of oblique trees (trained with the TAO algorithm), we can achieve even better accuracy than for axis-aligned trees using a much smaller number of trees with much smaller depth (up to 30 trees of depth 8 to 10 in fig.~\ref{f:forest-accuracy-size}). This is important because, while the region counterfactual requires solving a QP or LP, the number of live regions in the forest is quite smaller, so the runtime remains moderate. Note that the only counterfactual explanation algorithm considered here that can handle oblique forests is LIRE.

\subsection{AdaBoost forests}

\begin{figure}[p]
  \centering
  \psfrag{estimators}[c][][1]{\# trees}
  \begin{tabular}{@{\hspace{0.03\linewidth}}c@{\hspace{0.01\linewidth}}c@{}}
    \psfrag{regions}[cb][b]{\caja{b}{c}{axis-aligned trees \\[1ex] \# regions}}
    \includegraphics*[width=.48\linewidth]{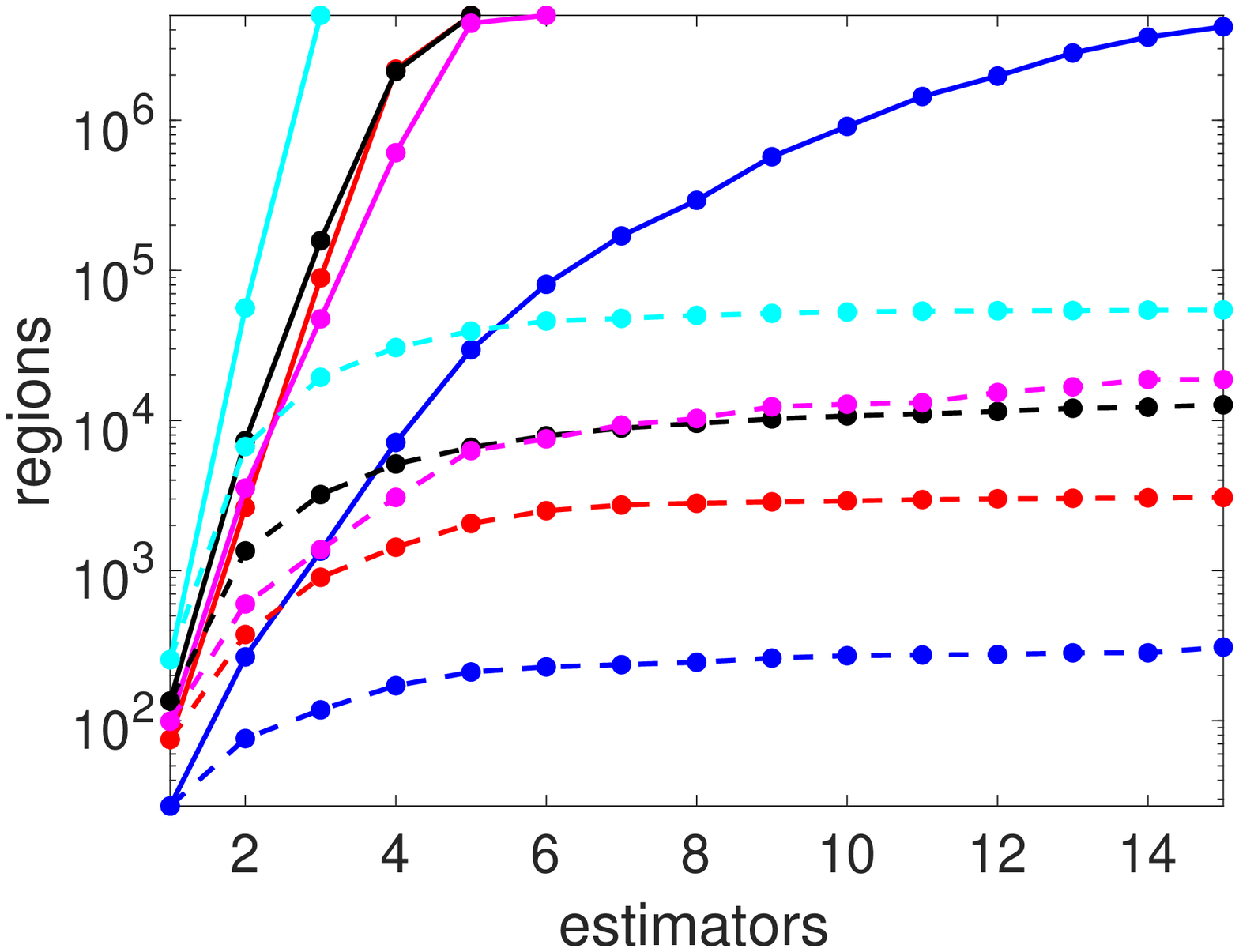}&
    \psfrag{brest cancer}[l][l]{Breast cancer}
    \psfrag{spambase}[l][l]{Spambase}
    \psfrag{letter}[l][l]{Letter}
    \psfrag{mnist}[l][l]{MNIST}
    \psfrag{adult}[l][l]{Adult}
    \psfrag{leaves}[c][][1]{Upper bound}
    \includegraphics*[width=.48\linewidth]{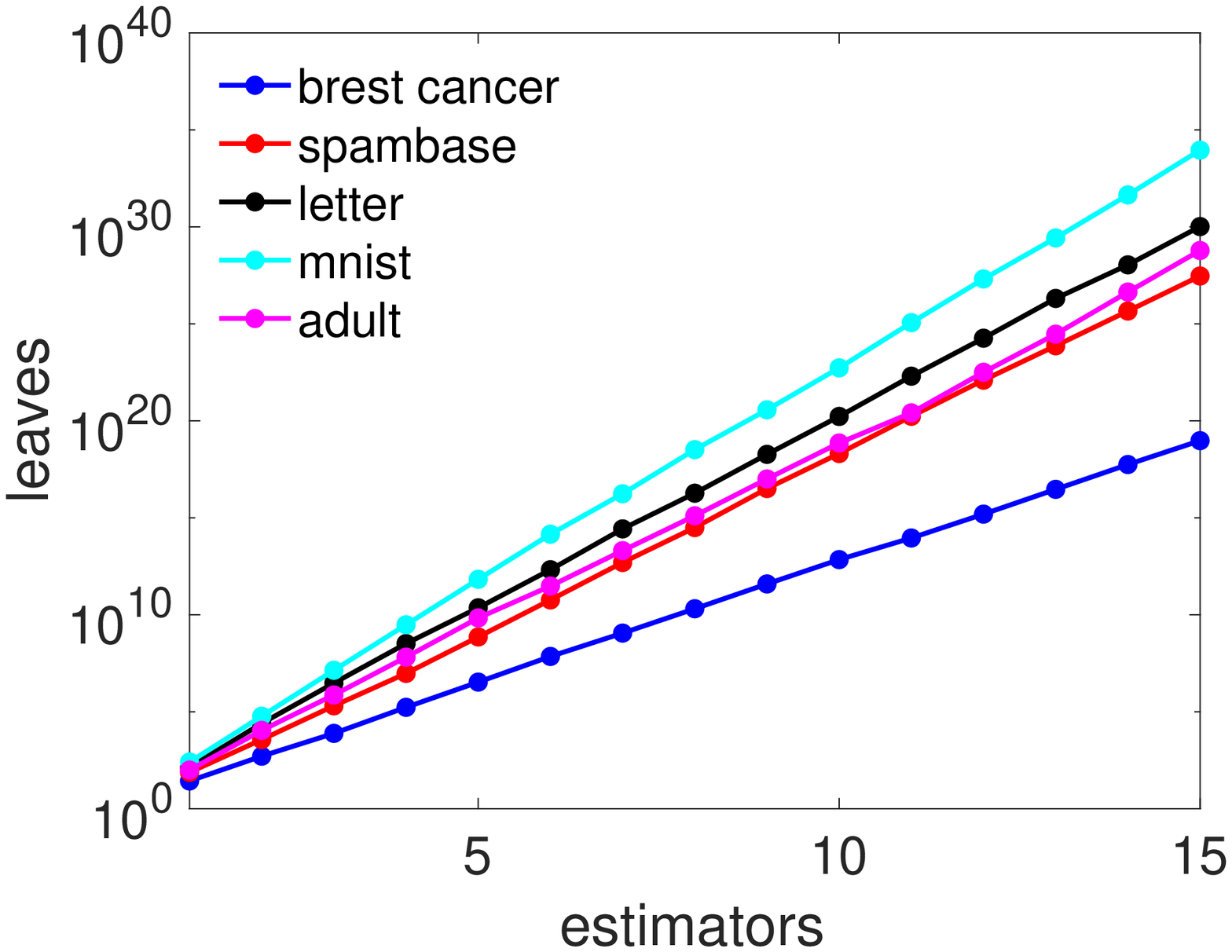}
  \end{tabular}
  \caption{Growth of the number of regions of an AdaBoost forest as a function of the number of trees $T$, for different datasets, for axis-aligned trees. On the left, we plot the number of nonempty regions (solid lines) and live regions (dashed lines); on the right, the upper bound for the number of nonempty regions. All regions are capped to a maximum of $5\cdot 10^6$.}
  \label{f:BoostregionsVsestimatorsComp}
\end{figure}

\begin{figure}[p]
  \centering
  \psfrag{exact}[l][l][0.8]{All regions}
  \psfrag{trptrestriction}[l][l][0.8]{\caja[0.2]{c}{c}{LIRE}}
  \psfrag{whatif}[l][bl][0.8]{\caja[0.2]{c}{c}{ dataset search}}
  \begin{tabular}{@{}c@{\hspace{0ex}}c@{\hspace{0ex}}c@{\hspace{0ex}}c@{}}
    {Breast cancer} & {Spambase} & {Letter} &{MNIST}\\
    \psfrag{errors}[c][]{avg.\ distance}
    \psfrag{estimators}{}
    \includegraphics*[width=.25\linewidth]{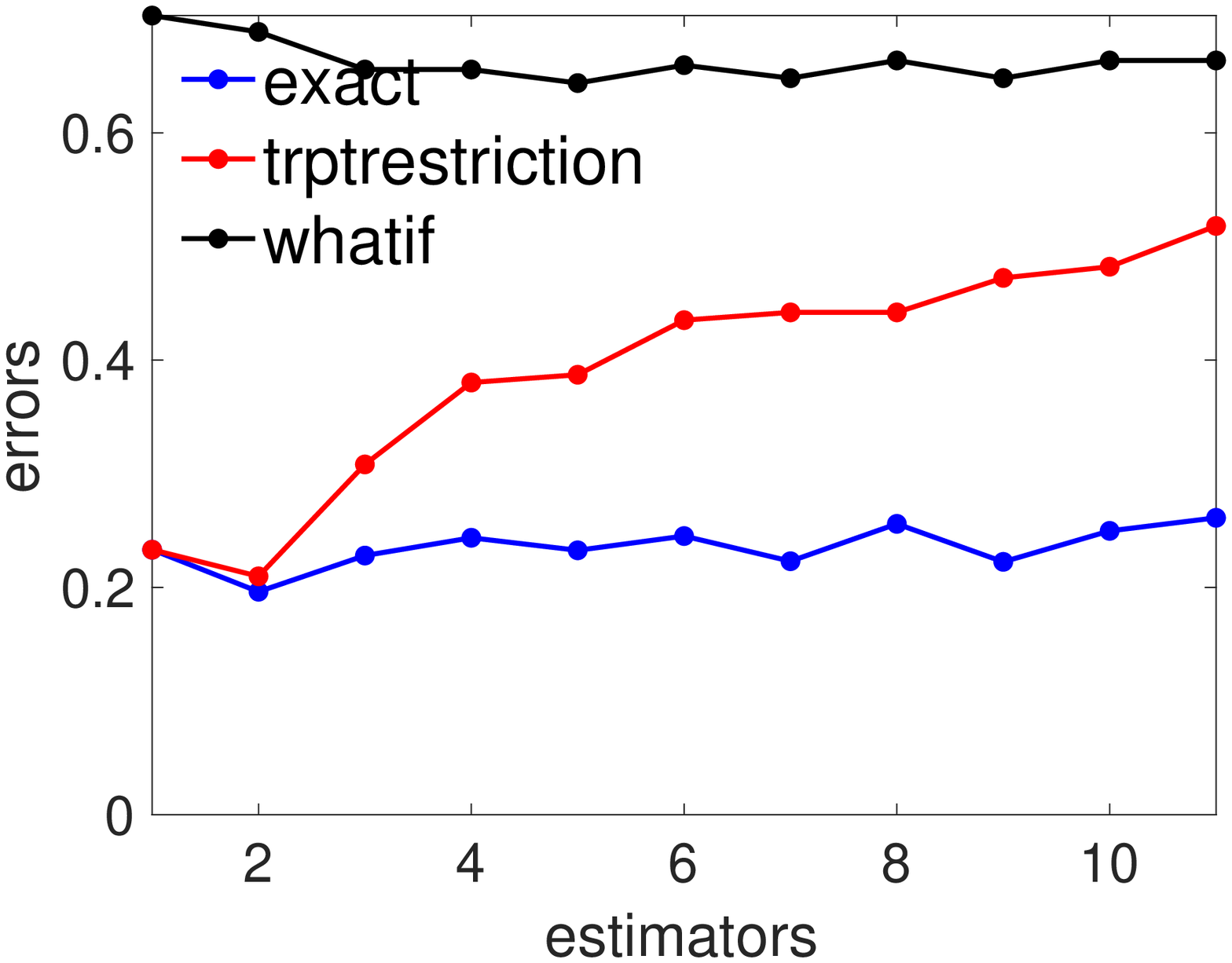}&
    \psfrag{errors}{}
    \psfrag{estimators}{}
    \includegraphics*[width=.25\linewidth,bb=13 4 512 393]{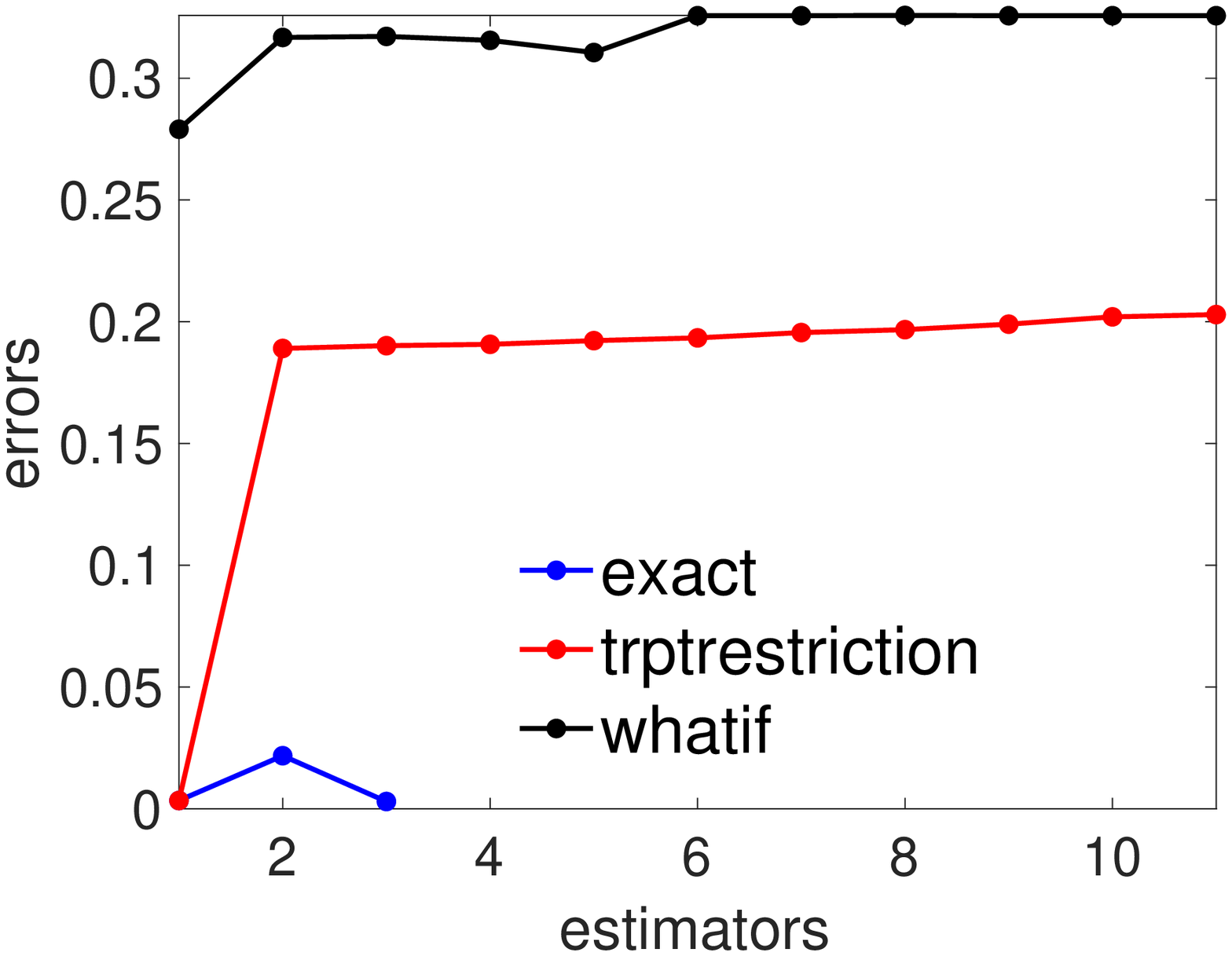}&
    \psfrag{errors}{}
    \psfrag{estimators}{}
    \includegraphics*[width=.25\linewidth]{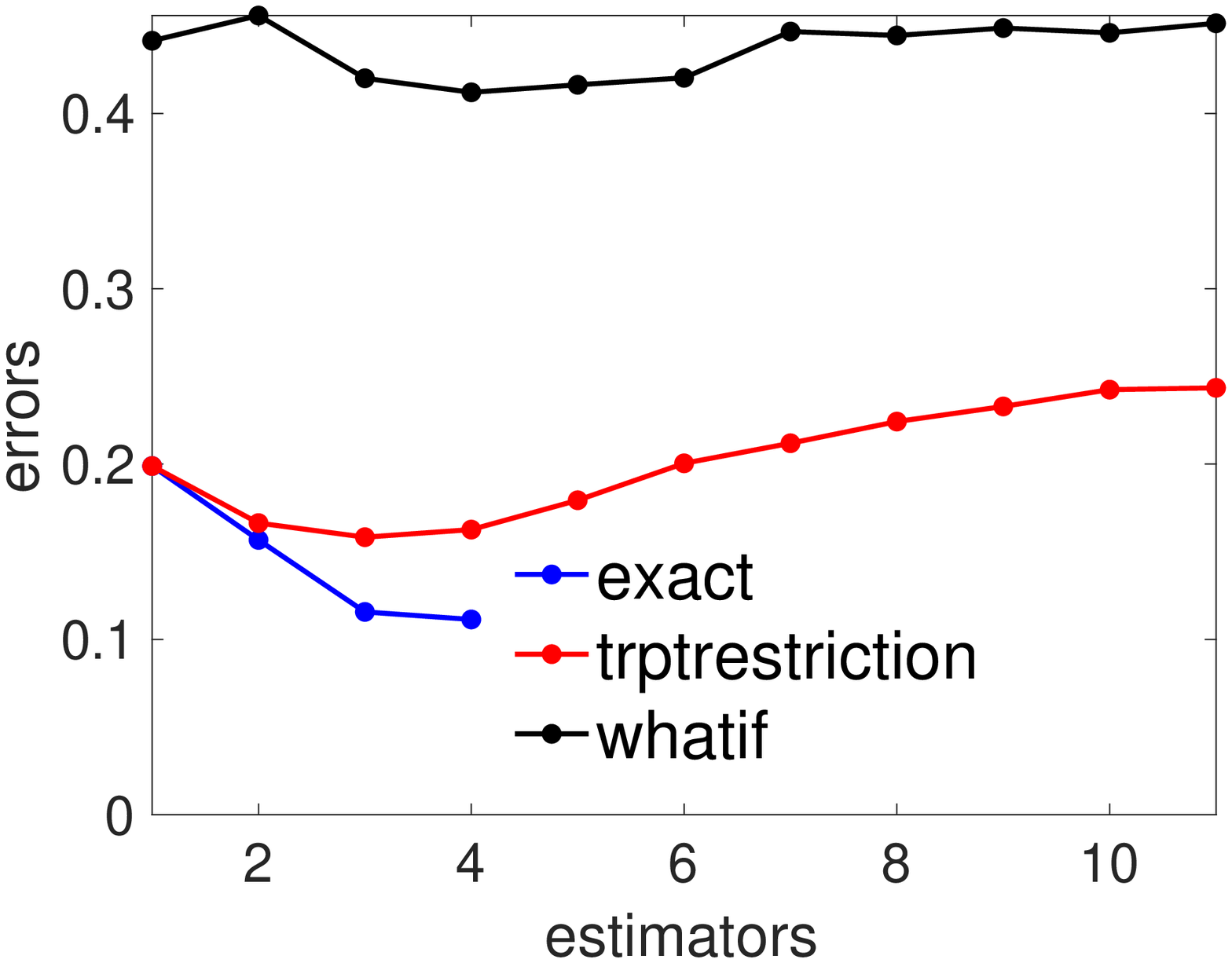}&
    \psfrag{errors}{}
    \psfrag{estimators}{}
    \includegraphics*[width=.25\linewidth,bb=13 4 512 393]{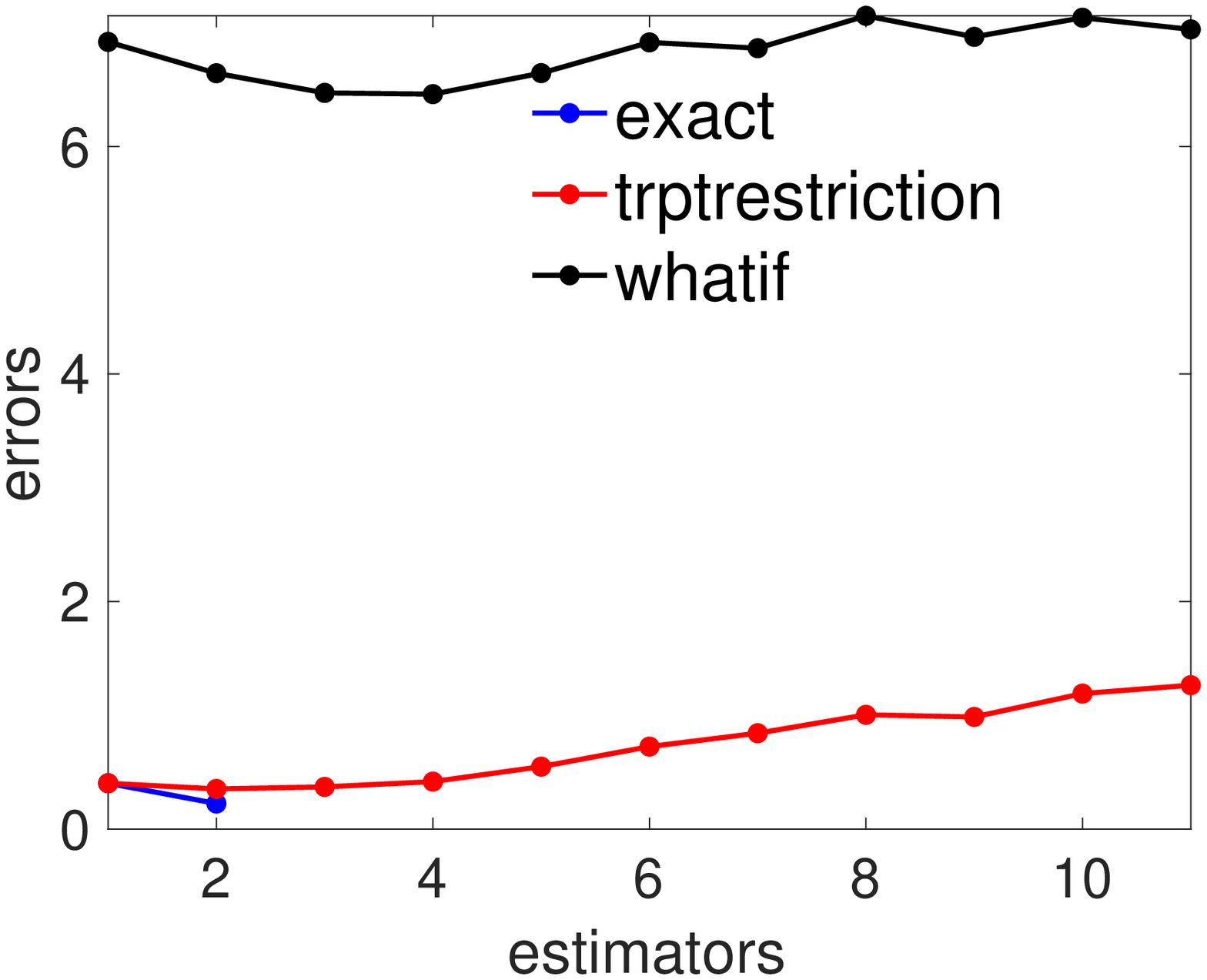}\\[-2ex]
    \psfrag{runtime}[c][]{avg.\ runtime (s)}
    \psfrag{estimators}[c][][1]{\# trees}
    \includegraphics*[width=.25\linewidth]{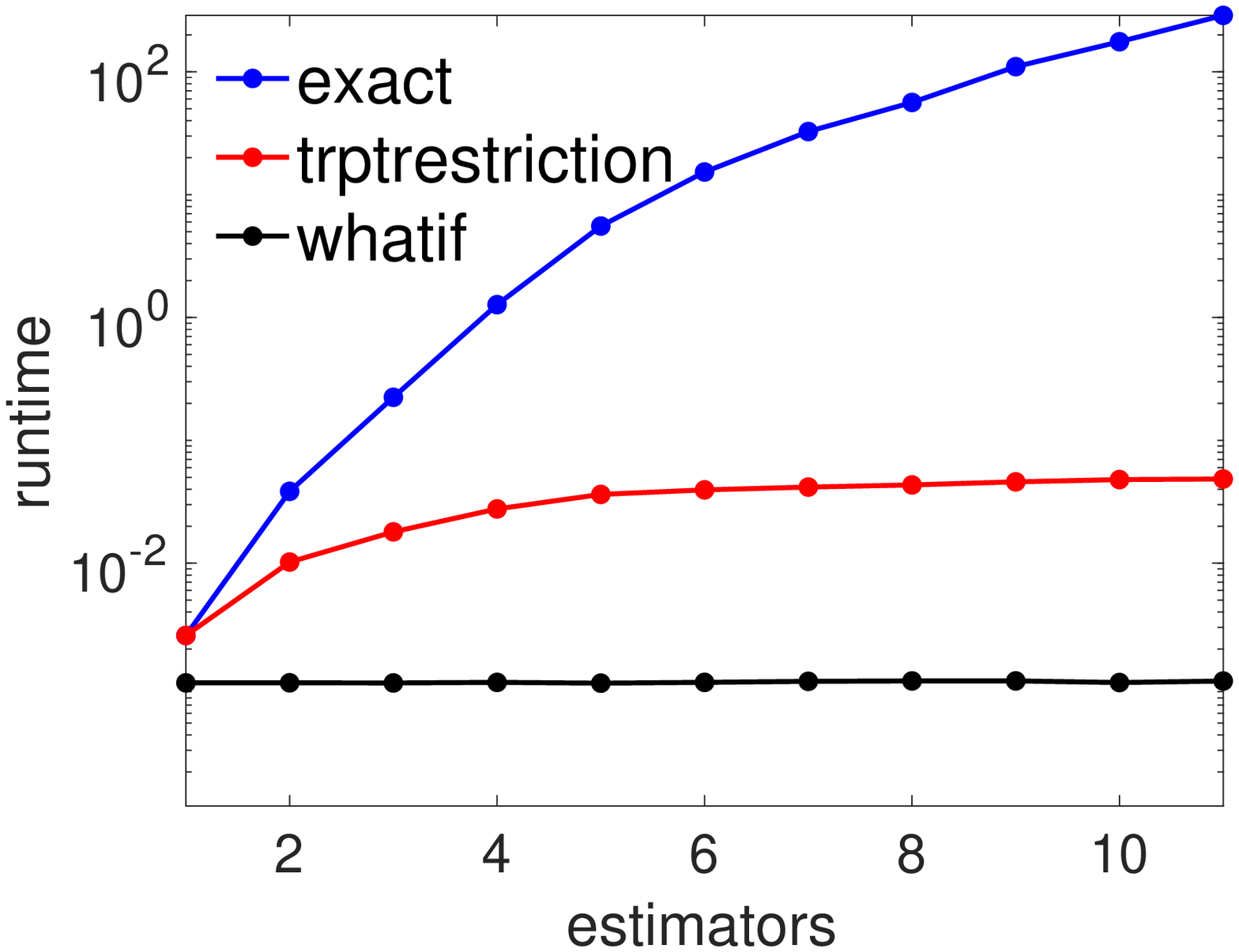}&
    \psfrag{runtime}{}
    \psfrag{estimators}[c][][1]{\# trees}
    \includegraphics*[width=.25\linewidth]{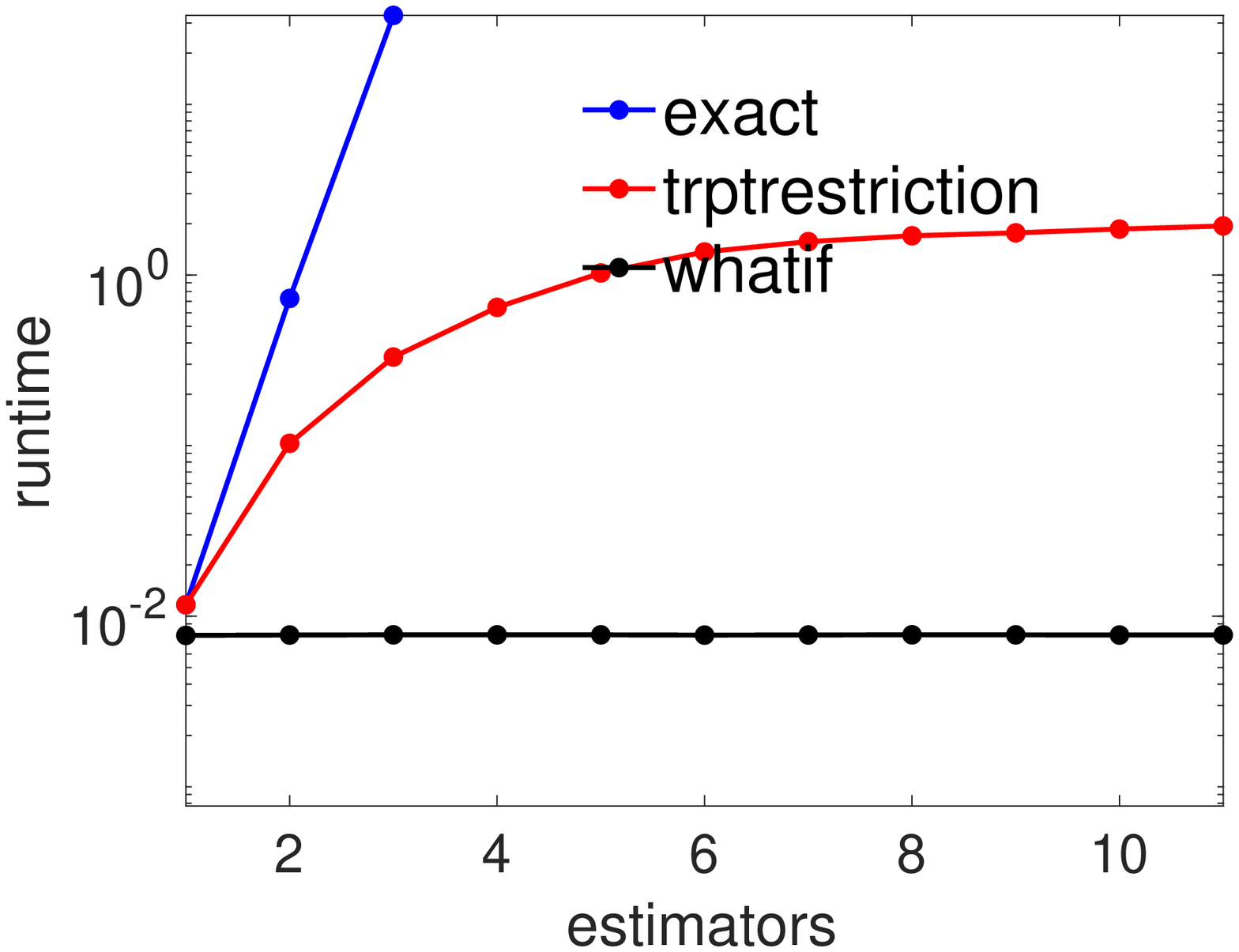}&
    \psfrag{runtime}{}
    \psfrag{estimators}[c][][1]{\# trees}
    \includegraphics*[width=.25\linewidth]{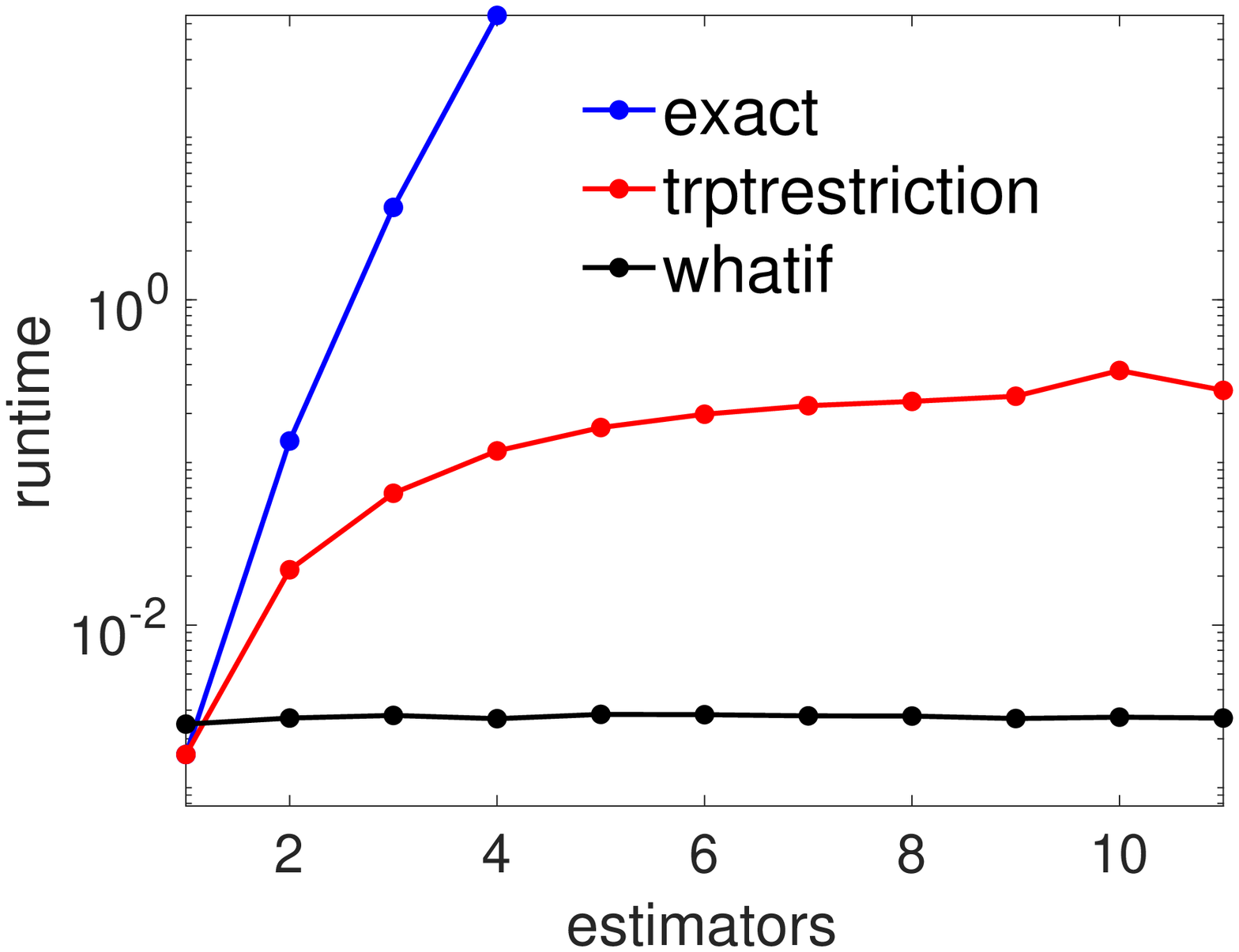}&
    \psfrag{runtime}{}
    \psfrag{estimators}[c][][1]{\# trees}
    \includegraphics*[width=.25\linewidth]{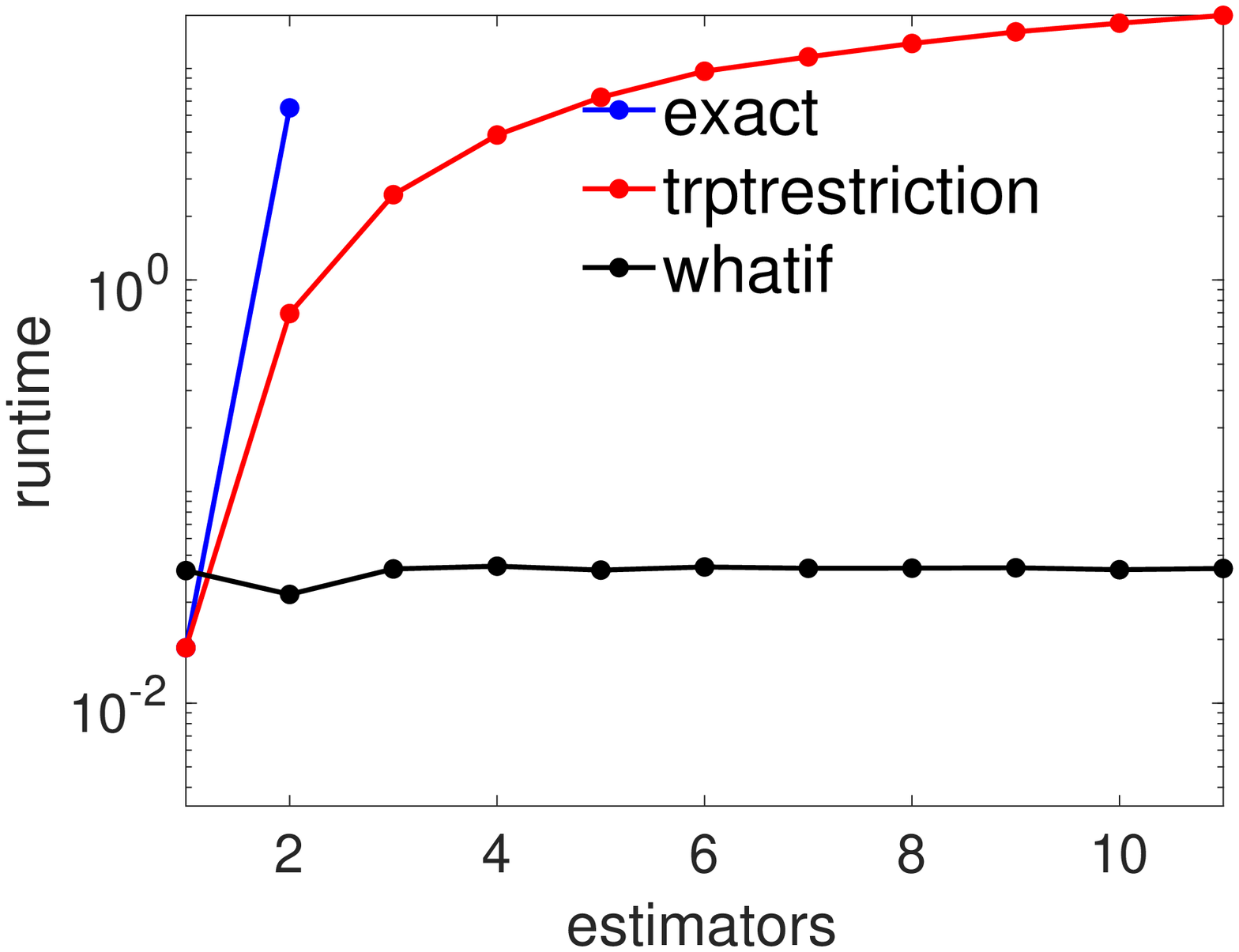}
  \end{tabular}
  \caption{Like fig.~\ref{f:exact-vs-LIRE} but for AdaBoost forest (axis-aligned trees).}
  \label{f:BoostCompErrosVsestimators}
\end{figure}

In this section, we present results with AdaBoost forests. As described in section~\ref{s:exp-setup} we use the \texttt{scikit-learn} AdaBoost classifier, where individual trees are axis-aligned and trained using CART. The results are qualitatively very similar to those shown in section~\ref{s:expts} for Random Forests. Specifically:
\begin{itemize}
  \item Growth of the number of regions of an AdaBoost forest as a function of the number of trees $T$ (fig.~\ref{f:BoostregionsVsestimatorsComp}). The number of non-empty regions grows exponentially as the number of trees in the forest increases, while the growth of live regions is upper bounded by the number of training instances.
  \item Approximation error between LIRE and using all non-empty regions for different datasets (fig.~\ref{f:BoostCompErrosVsestimators}). The approximation (in terms of the distance to the source instance) is quite good, though it degrades as the number of trees increases.
  \item Comparison of LIRE with dataset search (table~\ref{t:comparison-Boost}). LIRE finds better CEs and is very fast.
\end{itemize}

\begin{table}[p]
  \caption{Like table~\ref{t:comparison-oblique} but for the AdaBoost forest.}
  \label{t:comparison-Boost}
  \centering 
  \vspace*{1ex}
  \begin{tabular}{@{}ccc|c@{\hspace{1ex}}c@{\hspace{1ex}}r@{$\pm$}l@{\hspace{2ex}}|c@{\hspace{1ex}}r@{$\pm$}l@{}}
    Dataset & $(N,D,K)$ & $(T,\Delta,L)$ & \multicolumn{4}{c|}{LIRE} & \multicolumn{3}{c}{dataset search} \\
    & & & regions & time (s) & \multicolumn{2}{c|}{$\ell_2$} & time (s) & \multicolumn{2}{c}{$\ell_2$} \\
    \toprule  
    breast cancer & (559,9,2) & (100,5.9,32.3) & 376 &  2$\times 10^{-4}$ & \textbf{1.00}& \textbf{0.89} & 1$\times 10^{-4}$& 1.21&1.00 \\
    spambase & (3.6k,57,2) & (100,7.9,104.8) & 3226 &  6$\times 10^{-3}$&   \textbf{1.00}& \textbf{0.63} & 2$\times 10^{-4}$ & 1.41&0.51  \\
    letter & (16k,16,26) & (100,8.0,157.6) & 14877 & 9$\times 10^{-4}$ &  \textbf{1.00}& \textbf{0.76} & 6$\times 10^{-5}$& 1.32&0.91   \\
    MNIST & (55k,784,10) & (100,8.0,250.5) & 55000 & 2$\times 10^{-1}$ &  \textbf{1.00}& \textbf{0.58} & 4$\times 10^{-2}$ & 1.66&0.98
  \end{tabular}
\end{table}

\subsection{Growth of regions in Random Forest with depth of the trees}

\begin{figure}
  \centering
  \psfrag{depth}[c][][1]{depth}
  \begin{tabular}{@{\hspace{0.03\linewidth}}c@{\hspace{0.01\linewidth}}c@{}}
    \psfrag{regions}[cb][b]{\caja{b}{c}{axis-aligned trees \\[1ex] \# regions}}
    \includegraphics*[width=.48\linewidth]{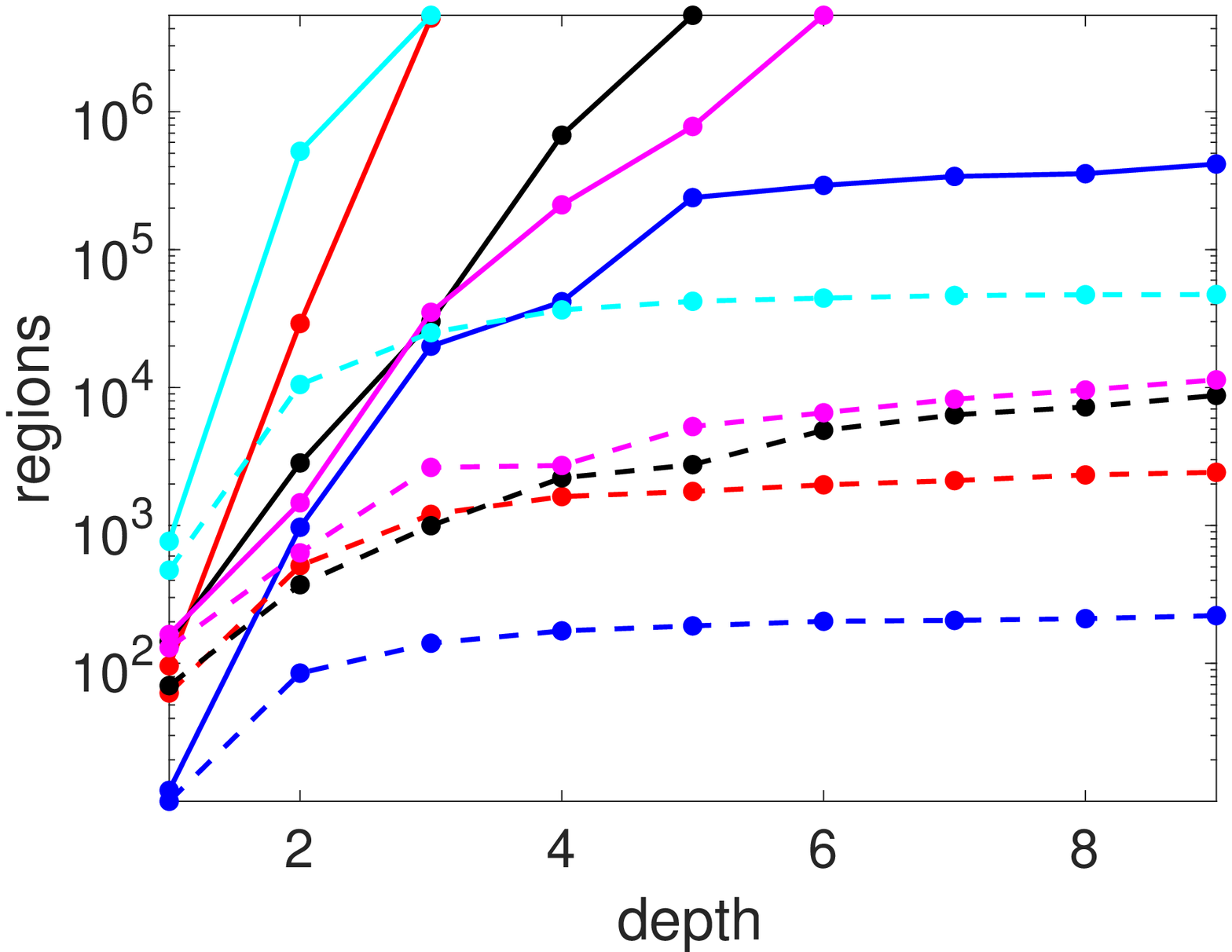}&
    \psfrag{brest cancer}[l][l]{Breast cancer}
    \psfrag{spambase}[l][l]{Spambase}
    \psfrag{letter}[l][l]{Letter}
    \psfrag{mnist}[l][l]{MNIST}
    \psfrag{adult}[l][l]{Adult}
    \psfrag{leaves}[c][][1]{Upper bound}
    \psfrag{estimators}[c][][1]{depth}
    \includegraphics*[width=.48\linewidth]{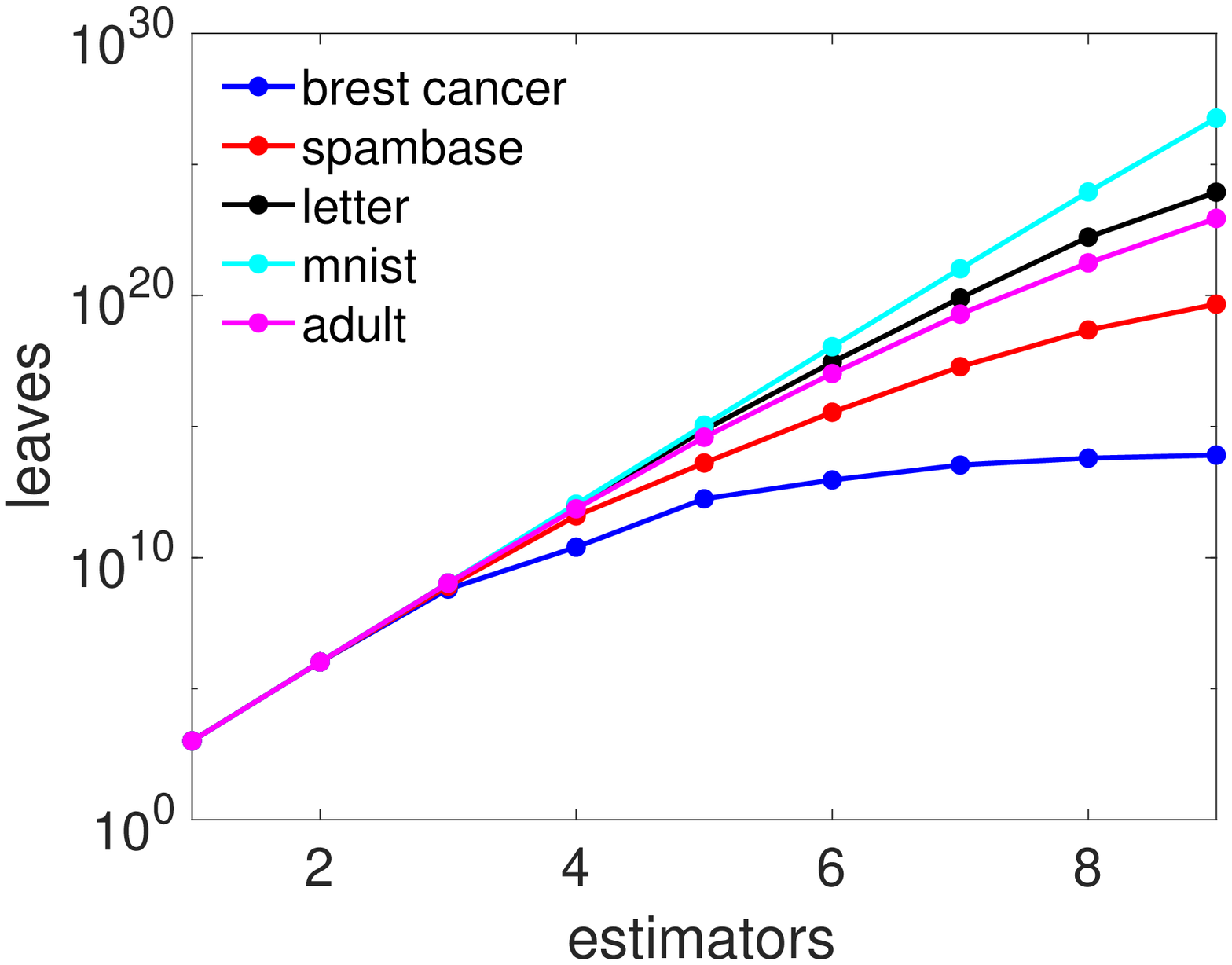}
  \end{tabular}
  \caption{Growth of the number of regions of a Random forest as a function of maximum depth of the trees $T$, for different datasets, for axis-aligned trees. On the left, we plot the number of nonempty regions (solid lines) and live regions (dashed lines); on the right, the upper bound for the number of nonempty regions. In each case, the number of trees in the forest is 10, and all regions are capped to a maximum of $5\cdot 10^6$.}
  \label{f:RFregionsVsdepthComp}
\end{figure}

In this section, we demonstrate how the number of regions in the Random Forest grows as a function of the depth of the individual trees. The experiment setup is the same as in fig.~\ref{f:num-regions}. The only difference is that here we change the maximum depth of the trees in the forest while keeping the number of trees in the forest constant. As shown in fig.~\ref{f:RFregionsVsdepthComp} the number of regions grows similarly to fig.~\ref{f:num-regions}. The number of live regions is bounded by $N$ (number of training instances), reaching $N$ even for smaller depths. Similarly, the number of non-empty regions hits the cap even for small depths for almost all datasets. 

\subsection{Realistic counterfactuals}

\begin{figure}
  \centering
  \scriptsize
  \newcommand{\mysize}{0.113}
  \begin{tabular}{@{}c@{\hspace{0.04\linewidth}}c@{\hspace{0.02\linewidth}}c@{\hspace{0.02\linewidth}}c@{}}
    & \multicolumn{3}{c}{\normalsize $\ell_2$}  \\
    \cline{2-4}
    \normalsize $\overline{\x}$ & \normalsize LIRE & \normalsize LIRE $\x_n$ & \normalsize dataset  \\
    \toprule
    (\underline{1},1.00,0.00) &  \textbf{(\underline 4,0.03,0.77),3.32} &  (\underline 4,0.03,0.70),4.41&  (\underline 4,0.1,0.70),31.98 \\
    \includegraphics*[width=\mysize \linewidth]{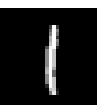}&
    \includegraphics*[width=\mysize \linewidth]{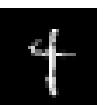}&
    \includegraphics*[width=\mysize \linewidth]{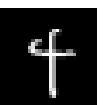}&
    \includegraphics*[width=\mysize \linewidth]{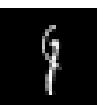}\\
    &  \normalsize all regions && \\
    & \textbf{(\underline 4,0.04,0.28),3.06} && \\
    &\includegraphics*[width=\mysize \linewidth]{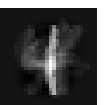}&&
  \end{tabular}
  \caption{The first row is the same as fig.~\ref{f:MNIST}, where the source class is 1 and the target class is 4, and only contains CEs generated by optimizing the $\ell_2$ distance. The second row contains one CE that is not in the live region and is closer to the source.}
  \label{f:MNIST-Allregion}
\end{figure}

In fig~\ref{f:MNIST-Allregion}, we show one example where generating a counterfactual that is closer to the source does not provide a realistic counterfactual. LIRE generates a counterfactual which is worse in terms of distance from the source but is more realistic, adding a horizontal stroke to convert the digit 1 into a digit 4.

\clearpage


\begin{thebibliography}{34}
\providecommand{\natexlab}[1]{#1}
\providecommand{\url}[1]{\texttt{#1}}
\expandafter\ifx\csname urlstyle\endcsname\relax
  \providecommand{\doi}[1]{doi: #1}\else
  \providecommand{\doi}{doi: \begingroup \urlstyle{rm}\Url}\fi

\bibitem[Breiman(2001)]{Breiman01a}
L.~Breiman.
\newblock Random forests.
\newblock \emph{Machine Learning}, 45\penalty0 (1):\penalty0 5--32, Oct. 2001.

\bibitem[Breiman et~al.(1984)Breiman, Friedman, Olshen, and Stone]{Breiman_84a}
L.~J. Breiman, J.~H. Friedman, R.~A. Olshen, and C.~J. Stone.
\newblock \emph{Classification and Regression Trees}.
\newblock Wadsworth, Belmont, Calif., 1984.

\bibitem[Carreira-Perpi{\~n}{\'a}n(2022)]{Carreir22a}
M.~{\'A}. Carreira-Perpi{\~n}{\'a}n.
\newblock The {Tree} {Alternating} {Optimization} {(TAO)} algorithm: A new way
  to learn decision trees and tree-based models.
\newblock arXiv, 2022.

\bibitem[Carreira-Perpi{\~n}{\'a}n and Hada(2021)]{CarreirHada21a}
M.~{\'A}. Carreira-Perpi{\~n}{\'a}n and S.~S. Hada.
\newblock Counterfactual explanations for oblique decision trees: Exact,
  efficient algorithms.
\newblock In \emph{Proc. of the 35th AAAI Conference on Artificial Intelligence
  (AAAI 2021)}, pages 6903--6911, Online, Feb.~2--9 2021.

\bibitem[Carreira-Perpi{\~n}{\'a}n and Hada(2023)]{CarreirHada23a}
M.~{\'A}. Carreira-Perpi{\~n}{\'a}n and S.~S. Hada.
\newblock Very fast, approximate counterfactual explanations for decision
  forests.
\newblock In \emph{Proc. of the 37th AAAI Conference on Artificial Intelligence
  (AAAI 2023)}, Washington, DC, Feb.~7--14 2023.

\bibitem[Carreira-Perpi{\~n}{\'a}n and Tavallali(2018)]{CarreirTavall18a}
M.~{\'A}. Carreira-Perpi{\~n}{\'a}n and P.~Tavallali.
\newblock Alternating optimization of decision trees, with application to
  learning sparse oblique trees.
\newblock In S.~Bengio, H.~Wallach, H.~Larochelle, K.~Grauman, N.~Cesa-Bianchi,
  and R.~Garnett, editors, \emph{Advances in Neural Information Processing
  Systems (NEURIPS)}, volume~31, pages 1211--1221. MIT Press, Cambridge, MA,
  2018.

\bibitem[Carreira-Perpi{\~n}{\'a}n and Zharmagambetov(2020)]{CarreirZharmag20a}
M.~{\'A}. Carreira-Perpi{\~n}{\'a}n and A.~Zharmagambetov.
\newblock Ensembles of bagged {TAO} trees consistently improve over random
  forests, {AdaBoost} and gradient boosting.
\newblock In \emph{Proc. of the 2020 ACM-IMS Foundations of Data Science
  Conference (FODS 2020)}, pages 35--46, Seattle, WA, Oct.~19--20 2020.

\bibitem[Carreira-Perpi{\~n}{\'a}n et~al.(2023)Carreira-Perpi{\~n}{\'a}n,
  Gabidolla, and Zharmagambetov]{Carreir_23a}
M.~{\'A}. Carreira-Perpi{\~n}{\'a}n, M.~Gabidolla, and A.~Zharmagambetov.
\newblock Towards better decision forests: {Forest} {Alternating}
  {Optimization}.
\newblock In \emph{Proc. of the 2023 IEEE Computer Society Conf. Computer
  Vision and Pattern Recognition (CVPR'23)}, Vancouver, Canada, June~18--22
  2023.

\bibitem[Cui et~al.(2015)Cui, Chen, He, and Chen]{Cui_15a}
Z.~Cui, W.~Chen, Y.~He, and Y.~Chen.
\newblock Optimal action extraction for random forests and boosted trees.
\newblock In \emph{Proc. of the 21st ACM SIGKDD Int. Conf. Knowledge Discovery
  and Data Mining (SIGKDD 2015)}, pages 179--188, Sydney, Australia,
  Aug.~10--13 2015.

\bibitem[Freund and Schapire(1997)]{FreundSchapir97a}
Y.~Freund and R.~Schapire.
\newblock A decision-theoretic generalization of on-line learning and an
  application to boosting.
\newblock \emph{J. Computer and System Sciences}, 55\penalty0 (1):\penalty0
  119--139, 1997.

\bibitem[Friedman(2001)]{Friedm01a}
J.~H. Friedman.
\newblock Greedy function approximation: A gradient boosting machine.
\newblock \emph{Annals of Statistics}, 29\penalty0 (5):\penalty0 1189--1232,
  2001.

\bibitem[Gabidolla and Carreira-Perpi{\~n}{\'a}n(2022)]{GabidolCarreir22a}
M.~Gabidolla and M.~{\'A}. Carreira-Perpi{\~n}{\'a}n.
\newblock Pushing the envelope of gradient boosting forests via
  globally-optimized oblique trees.
\newblock In \emph{Proc. of the 2022 IEEE Computer Society Conf. Computer
  Vision and Pattern Recognition (CVPR'22)}, pages 285--294, New Orleans, LA,
  June~19--24 2022.

\bibitem[Gabidolla et~al.(2022)Gabidolla, Zharmagambetov, and
  Carreira-Perpi{\~n}{\'a}n]{Gabidol_22a}
M.~Gabidolla, A.~Zharmagambetov, and M.~{\'A}. Carreira-Perpi{\~n}{\'a}n.
\newblock Improved multiclass {AdaBoost} using sparse oblique decision trees.
\newblock In \emph{Int. J. Conf. Neural Networks (IJCNN'22)}, Padua, Italy,
  July~18--22 2022.

\bibitem[Goodfellow et~al.(2015)Goodfellow, Shlens, and Szegedy]{Goodfel_15a}
I.~J. Goodfellow, J.~Shlens, and C.~Szegedy.
\newblock Explaining and harnessing adversarial examples.
\newblock In \emph{Proc. of the 3rd Int. Conf. Learning Representations (ICLR
  2015)}, San Diego, CA, May~7--9 2015.

\bibitem[Guidotti et~al.(2019)Guidotti, Monreale, Giannotti, Pedreschi,
  Ruggieri, and Turini]{Guidot_19a}
R.~Guidotti, A.~Monreale, F.~Giannotti, D.~Pedreschi, S.~Ruggieri, and
  F.~Turini.
\newblock Factual and counterfactual explanations for black box decision
  making.
\newblock \emph{IEEE Access}, 34\penalty0 (6):\penalty0 14--23, Nov. -- Dec.
  2019.

\bibitem[Hada and Carreira-Perpi{\~n}{\'a}n(2021)]{HadaCarreir21b}
S.~S. Hada and M.~{\'A}. Carreira-Perpi{\~n}{\'a}n.
\newblock Exploring counterfactual explanations for classification and
  regression trees.
\newblock In \emph{ECML PKDD 3rd Int. Workshop and Tutorial on eXplainable
  Knowledge Discovery in Data Mining (XKDD 2021)}, pages 489--504, 2021.

\bibitem[Hastie et~al.(2009)Hastie, Tibshirani, and Friedman]{Hastie_09a}
T.~J. Hastie, R.~J. Tibshirani, and J.~H. Friedman.
\newblock \emph{The Elements of Statistical Learning---Data Mining, Inference
  and Prediction}.
\newblock Springer Series in Statistics. Springer-Verlag, second edition, 2009.

\bibitem[Kanamori et~al.(2007)Kanamori, Takagi, Kobayashi, and
  Arimura]{Kanamor_20a}
K.~Kanamori, T.~Takagi, K.~Kobayashi, and H.~Arimura.
\newblock {DACE}: Distribution-aware counterfactual explanation by
  mixed-integer linear optimization.
\newblock In \emph{Proc. of the 20th Int. Joint Conf. Artificial Intelligence
  (IJCAI'07)}, pages 2855--2862, Hyderabad, India, Jan.~6--12 2007.

\bibitem[Karimi et~al.(2020)Karimi, Barthe, Balle, and Valera]{Karimi_20a}
A.-H. Karimi, G.~Barthe, B.~Balle, and I.~Valera.
\newblock Model-agnostic counterfactual explanations for consequential
  decisions.
\newblock In \emph{Proc. of the 23rd Int. Conf. Artificial Intelligence and
  Statistics (AISTATS 2020)}, pages 895--905, Online, Aug.~26--28 2020.

\bibitem[{LeCun} et~al.(1998){LeCun}, Bottou, Bengio, and Haffner]{Lecun_98a}
Y.~{LeCun}, L.~Bottou, Y.~Bengio, and P.~Haffner.
\newblock Gradient-based learning applied to document recognition.
\newblock \emph{Proc. IEEE}, 86\penalty0 (11):\penalty0 2278--2324, Nov. 1998.

\bibitem[Lichman(2013)]{Lichman13a}
M.~Lichman.
\newblock {UCI} machine learning repository.
\newblock \url{http://archive.ics.uci.edu/ml}, 2013.

\bibitem[Lodi and Tramontani(2013)]{LodiTramon13a}
A.~Lodi and A.~Tramontani.
\newblock Performance variability in mixed-integer programming.
\newblock \emph{INFORMS TutORials in Operations Research}, pages 1--12, Sept.
  2013.

\bibitem[Lucic et~al.(2022)Lucic, Oosterhuis, Haned, and de~Rijke]{Lucic_22a}
A.~Lucic, H.~Oosterhuis, H.~Haned, and M.~de~Rijke.
\newblock {FOCUS}: Flexible optimizable counterfactual explanations for tree
  ensembles.
\newblock In \emph{Proc. of the 36th AAAI Conference on Artificial Intelligence
  (AAAI 2022)}, pages 5313--5322, Online, Feb.~22 -- Mar.~1 2022.

\bibitem[Mothilal et~al.(2020)Mothilal, Sharma, and Tan]{Mothil_20a}
R.~K. Mothilal, A.~Sharma, and C.~Tan.
\newblock Explaining machine learning classifiers through diverse
  counterfactual explanations.
\newblock In \emph{Proc. ACM Conf. Fairness, Accountability, and Transparency
  (FAT 2020)}, pages 607--617, 2020.

\bibitem[Parmentier and Vidal(2021)]{ParmenVidal21a}
A.~Parmentier and T.~Vidal.
\newblock Optimal counterfactual explanations in tree ensembles.
\newblock In M.~Meila and T.~Zhang, editors, \emph{Proc. of the 38th Int. Conf.
  Machine Learning (ICML 2021)}, pages 8422--8431, Online, July~18--24 2021.

\bibitem[Russell(2019)]{Russel19a}
C.~Russell.
\newblock Efficient search for diverse coherent explanations.
\newblock In \emph{Proc. ACM Conf. Fairness, Accountability, and Transparency
  (FAT 2019)}, pages 20--28, Atlanta, GA, Jan.~29--31 2019.

\bibitem[Szegedy et~al.(2014)Szegedy, Zaremba, Sutskever, Bruna, Erhan,
  Goodfellow, and Fergus]{Szeged_14a}
C.~Szegedy, W.~Zaremba, I.~Sutskever, J.~Bruna, D.~Erhan, I.~Goodfellow, and
  R.~Fergus.
\newblock Intriguing properties of neural networks.
\newblock In \emph{Proc. of the 2nd Int. Conf. Learning Representations (ICLR
  2014)}, Banff, Canada, Apr.~14--16 2014.

\bibitem[Tolomei et~al.(2017)Tolomei, Silvestri, Haines, and
  Lalmas]{Tolomei_17a}
G.~Tolomei, F.~Silvestri, A.~Haines, and M.~Lalmas.
\newblock Interpretable predictions of tree-based ensembles via actionable
  feature tweaking.
\newblock In \emph{Proc. of the 23rd ACM SIGKDD Int. Conf. Knowledge Discovery
  and Data Mining (SIGKDD 2017)}, pages 465--474, Halifax, Nova Scotia,
  Aug.~13--17 2017.

\bibitem[Ustun et~al.(2019)Ustun, Spangher, and Liu]{Ustun_19a}
B.~Ustun, A.~Spangher, and Y.~Liu.
\newblock Actionable recourse in linear classification.
\newblock In \emph{Proc. ACM Conf. Fairness, Accountability, and Transparency
  (FAT 2019)}, pages 10--19, Atlanta, GA, Jan.~29--31 2019.

\bibitem[Van~Looveren and Klaise(2021)]{VanloovKlaise21a}
A.~Van~Looveren and J.~Klaise.
\newblock Interpretable counterfactual explanations guided by prototypes.
\newblock In N.~Oliver, F.~P{\'e}rez-Cruz, S.~Kramer, J.~Read, and J.~A.
  Lozano, editors, \emph{Proc. of the 32nd European Conf. Machine Learning
  (ECML--21)}, Bilbao, Spain, Sept.~13--17 2021.

\bibitem[White and {d'Avila}~Garcez(2020)]{WhiteDavila20a}
A.~White and A.~{d'Avila}~Garcez.
\newblock Measurable counterfactual local explanations for any classifier.
\newblock In G.~D. Giacomo, A.~Catala, B.~Dilkina, M.~Milano, S.~Barro,
  A.~Bugar{\'{\i}}n, and J.~Lang, editors, \emph{Proc. 24th European Conf.
  Artificial Intelligence (ECAI 2020)}, pages 2529--2535, Aug.~29 -- Sept.~8
  2020.

\bibitem[Yang et~al.(2006)Yang, Yin, Ling, and Pan]{Yang_06c}
Q.~Yang, J.~Yin, C.~X. Ling, and R.~Pan.
\newblock Extracting actionable knowledge from decision trees.
\newblock \emph{IEEE Trans. Knowledge and Data Engineering}, 18\penalty0
  (1):\penalty0 43--56, Jan. 2006.

\bibitem[Zharmagambetov and Carreira-Perpi{\~n}{\'a}n(2020)]{ZharmagCarreir20a}
A.~Zharmagambetov and M.~{\'A}. Carreira-Perpi{\~n}{\'a}n.
\newblock Smaller, more accurate regression forests using tree alternating
  optimization.
\newblock In H.~{Daum{\'e}~III} and A.~Singh, editors, \emph{Proc. of the 37th
  Int. Conf. Machine Learning (ICML 2020)}, pages 11398--11408, Online,
  July~13--18 2020.

\bibitem[Zharmagambetov et~al.(2021)Zharmagambetov, Hada, Gabidolla, and
  Carreira-Perpi{\~n}{\'a}n]{Zharmag_21b}
A.~Zharmagambetov, S.~S. Hada, M.~Gabidolla, and M.~{\'A}.
  Carreira-Perpi{\~n}{\'a}n.
\newblock Non-greedy algorithms for decision tree optimization: An experimental
  comparison.
\newblock In \emph{Int. J. Conf. Neural Networks (IJCNN'21)}, Virtual event,
  July~18--22 2021.

\end{thebibliography}

\end{document}